\documentclass[10pt,journal,compsoc]{IEEEtran}
\ifCLASSOPTIONcompsoc
  \usepackage[nocompress]{cite}
\else
  \usepackage{cite}
\fi
\ifCLASSINFOpdf
\else
\fi
 \usepackage{threeparttable}
 \usepackage{xspace}
 \usepackage{amsmath}
 \usepackage{makecell}
 \usepackage{colortbl}
 \usepackage[dvipsnames]{xcolor}
\usepackage{xcolor}
\usepackage{tabularx} 
\usepackage{arydshln}
\usepackage{booktabs}       % professional-quality tables
\usepackage{amsfonts}       % blackboard math symbols
\usepackage{nicefrac}       % compact symbols for 1/2, etc.
\usepackage{microtype}      % microtypography
\usepackage{diagbox}
\usepackage{slashbox}
\usepackage{array}
\usepackage{tabularx}
\usepackage{array}
\usepackage{multicol}
\usepackage{longtable}
\usepackage{makecell}

\usepackage[pagebackref=false,breaklinks=true,letterpaper=true,colorlinks,bookmarks=false,urlcolor=red,citecolor=cyan]{hyperref}

\newcolumntype{g}{>{\columncolor{Gray}}r}
\newcommand{\transfer}[2]{#1\% [#2]}
\newcommand{\knowledgetransfer}[1]{\mathbb{T}_{\textit{UK}}(#1)}
\newcommand{\algorithm}{\mathcal{A}}
\newcommand{\vilt}{ViLT}
\newcommand{\forgetting}[2]{\mathbb{T}_{F}(#1 \leftarrow #2)}

\newlength\savewidth\newcommand\shline{\noalign{\global\savewidth\arrayrulewidth\global\arrayrulewidth 1pt}\hline\noalign{\global\arrayrulewidth\savewidth}}

\definecolor{shallowblue}{RGB}{227, 240, 249}
\definecolor{deepblue}{RGB}{192, 216, 240}
\definecolor{NA}{rgb}{1, 1, 1}
\definecolor{lightblue}{rgb}{0.93, 0.95, 1.0}
\definecolor{mygray}{gray}{.9}

\usepackage{epsfig}
\usepackage{graphicx}
\usepackage{amsmath}
\usepackage{amssymb}
\usepackage{enumitem}

\usepackage{gensymb,graphics,subfigure,inputenc}
\usepackage[linesnumbered,algo2e,boxed]{algorithm2e}
\graphicspath{{Figure/}}
\usepackage[figuresright]{rotating}
\usepackage{amsmath,amssymb,textcomp,subfigure,multirow,upgreek}
\usepackage{booktabs}
\usepackage{color,mdwlist}

\usepackage{ragged2e}

\usepackage{review}

\usepackage{graphicx,fontawesome}
\graphicspath{{Figures/}}

\usepackage{caption}
\usepackage{wrapfig}

\usepackage{enumitem}
\setitemize{noitemsep,topsep=0pt,parsep=0pt,partopsep=0pt}

\begin{document}
\title{When Continue Learning Meets Multimodal Large Language Model: A Survey}
\author{Yukang Huo \quad Hao Tang$^*$
	\IEEEcompsocitemizethanks{
	   \IEEEcompsocthanksitem Yukang Huo is with the School of College of Information and Electrical Engineering, China Agricultural University, Beijing 100193, China. E-mail: yukanghuo.ai@gmail.com\protect
        \IEEEcompsocthanksitem Hao Tang is with the School of Computer Science, Peking University, Beijing 100871, China. E-mail: haotang@pku.edu.cn \protect
        }% <-this % stops an unwanted space
	\thanks{$^*$Corresponding author: Hao Tang.}
 }

\markboth{IEEE Transactions on Pattern Analysis and Machine Intelligence}%
{Shell \MakeLowercase{\textit{et al.}}: Bare Demo of IEEEtran.cls for Computer Society Journals}



\IEEEtitleabstractindextext{%
%\begin{abstract}
%The abstract goes here.
%\end{abstract}
\justify
\begin{abstract}
In recent years, significant progress has been made in the field of Artificial Intelligence with the development of Multimodal Large Language Models (MLLMs). However, adapting static, pre-trained MLLMs to dynamic data distributions and various tasks in an accurate and efficient manner remains a major challenge. When fine-tuning pre-trained MLLMs for specific tasks, a noticeable performance degradation often occurs in the model's prior knowledge domain — a phenomenon known as ``Catastrophic Forgetting.'' While this issue has been extensively studied within the Continual Learning (CL) community, it presents new challenges in the context of MLLMs. 
As the first review paper in the field of continual learning for multimodal large models, this paper provides a comprehensive overview and detailed analysis of the 440 research papers on MLLM continual learning. Beyond introducing the fundamental concepts, the review is structured into four main sections.
Firstly, it provides an overview of the latest research on MLLMs, including various model innovation strategies, benchmarks, and applications across diverse fields. Secondly, it presents a detailed categorization and overview of the latest research on continual learning, divided into three key areas: non-large language models(LLMs) unimoda continual learning (Non-LLM Unimodal CL), non-large language models multimodal continual learning (Non-LLM Multimoda CL), and continual learning in large language models (CL in LLM). In-depth and extensive research in both the MLLM and CL domains has laid a solid foundation for research on MLLM continual learning. In the fourth section, we conduct an in-depth analysis of the current research status of MLLM continual learning, examining common benchmark evaluations, innovative improvements in model architectures and methods, and systematically summarizing and reviewing existing theoretical and empirical studies. This review aims to connect the basic setup, theoretical foundations, method innovations, and practical applications of continual learning in multimodal large models, shedding light on the research progress and challenges in the field. Finally, this paper offers a forward-looking discussion on the challenges and future development trends of continual learning in multimodal large models, aiming to inspire researchers in the field and promote the advancement of related technologies.
\end{abstract}

}

\maketitle

\IEEEdisplaynontitleabstractindextext

% \fi
\IEEEpeerreviewmaketitle



\begin{figure*}[htbp]
    \centering
    \includegraphics[width=0.9\linewidth,height=0.5\linewidth]{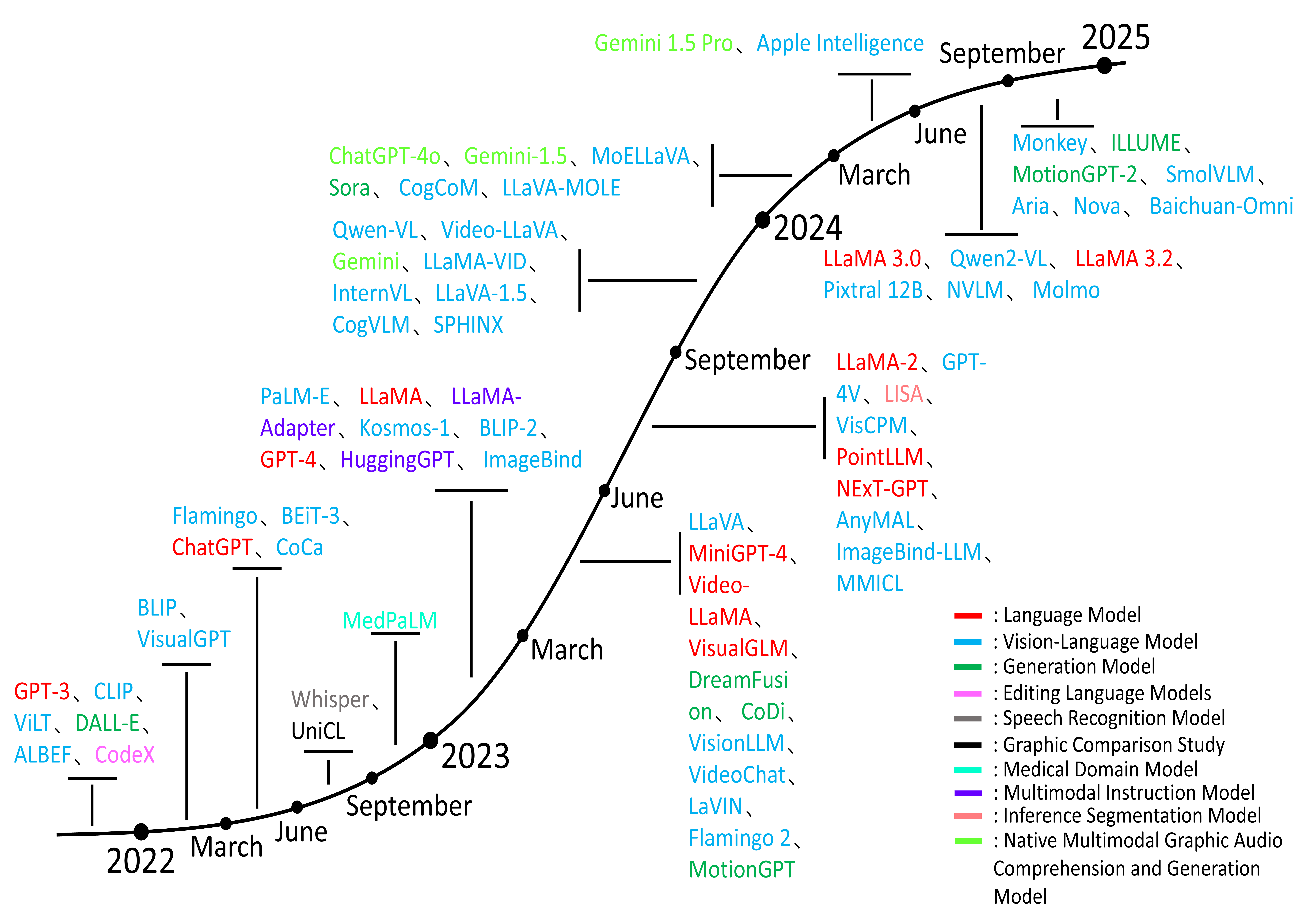}
    \caption{Timeline of Multimodal Large Model Development.}
    \label{develop}
    \vspace{-4mm} 
\end{figure*} 

\section{Introduction}
Research on Multimodal Large Language Models (MLLMs) has rapidly advanced in recent years, becoming a significant direction in the field of artificial intelligence~\cite{zhan2024anygpt,chiang2023vicuna,jiang2023motiongpt,zhang2024motiongpt,liu2024sphinx,sarch2024ical,wu2024visionllm,li2024single,dong2024internlm,mu2024embodiedgpt}. By integrating multimodal information such as language, vision, and audio, these models demonstrate powerful cross-modal understanding and generation capabilities, providing innovative solutions to complex real-world problems~\cite{huang2024multimodal,dong2024multiood,wu2024hypergraph,liu2024multi,wang2024cloud}. To enhance the performance of MLLMs, researchers have proposed various improvement strategies. Firstly, for cross-modal information fusion, more efficient architectural designs have been introduced~\cite{shukor2024implicit,zhang2024caption,xu2024libra}, such as Transformer-based multimodal joint encoders and decoders, as well as lightweight cross-modal attention modules~\cite{zhuang2024towards,forouzandehmehr2024decoding,gao2024clova}. Secondly, pre-training techniques have been further developed, significantly improving the model's generalization ability and robustness through the introduction of multimodal contrastive learning, cross-modal consistency constraints, and self-supervised learning objectives~\cite{szot2024grounding,wu2024controlmllm,li2024membership,sagar2024failures}. In addition, fine-tuning techniques have become increasingly refined~\cite{zhai2024fine}, including efficient parameter adjustment methods (such as LoRA~\cite{hu2021lora}) and task-specific adaptation layer designs. These approaches enable MLLMs to adapt to diverse task scenarios with lower computational costs~\cite{houlsby2019parameter,liu2021p,liu2024gpt,li2021prefix}. As shown in Figure~\ref{develop}, the performance evaluation of MLLMs is based on multimodal benchmarks that cover a wide range of task categories. For example, benchmarks in the vision and language domain include Visual Question Answering (VQA)~\cite{antol2015vqa, mathew2021docvqa, masry2022chartqa,kafle2018dvqa,chen2021geoqa}, Image Captioning~\cite{lin2014microsoft,sharma2018conceptual,sidorov2020textcaps,gurari2020captioning,pont2020connecting,agrawal2019nocaps}, and Visual Grounding~\cite{cui2021s,yu2016modeling,mao2016generation,tanaka2019generating}; in the audio and language domain, benchmarks include Audio-Text Alignment and Audio Generation~\cite{gemmeke2017audio,hernandez2018ted,ardila2019common}; there are also more complex cross-modal reasoning tasks, among others~\cite{wadhawan2024contextual,wu2024see}. Moreover, MLLMs are also showing great potential in real-world applications. They are playing an increasingly important role in fields such as healthcare, education, robotics, and autonomous driving~\cite{guo2024llava,chen2024llm,huang2025making}.

Continual learning aims to address the challenge of how models can effectively learn new tasks while retaining prior knowledge when faced with dynamically changing data streams, thus mitigating the problem of catastrophic forgetting~\cite{cossu2024continual,qin2022elle,sun2020ernie}. In recent years, research in the field of continuous learning has been deepened, particularly with significant developments in its application across models of various scales and multimodal learning scenarios~\cite{wang2022learning,fu2024data,guo2025pilora,pang2024ftf,hu2023dense,abati2020conditional}. In unimodal settings, the focus has mainly been on the design of algorithms to alleviate the problem of catastrophic forgetting, enabling models to maintain performance in previous tasks while incorporating new ones~\cite{yang2024generating,guha2024diminishing,cha2024regularizingpseudonegativescontinualselfsupervised,lee2024becotta,zhao2024statistical,garg2025poet}. Research in multimodal continual learning is more challenging than in unimodal  settings, as models must simultaneously handle the characteristics of different modalities and their cross-modal interactions~\cite{kim2025one,pang2024ftf,yang2024semantic,nori2024task}. Researchers have primarily focused on techniques for cross-modal feature extraction, alignment, and processing, aiming to reduce cross-modal interference, enhance inter-modal consistency, and improve the model's generalization ability~\cite{kim2025open,marczak2025revisiting,li2024learn,chen2023promptfusion}. With the widespread application of large language models (LLMs) in natural language processing, research on their continual learning has become a new hotspot~\cite{guo2023continuous,colombo2024saullm,deng2024k2,gururangan2020don,han2020econet,ma2023ecomgpt}. Due to the massive parameter scale of LLMs and their reliance on vast amounts of pre-trained data, traditional continual learning strategies face challenges such as high computational costs and limited adaptability. To address these challenges, researchers have proposed several optimization directions: Parameter-Efficient Fine-Tuning (PEFT) methods (such as LoRA, Prefix Tuning, etc.)~\cite{hu2021lora,houlsby2019parameter,liu2021p,liu2024gpt,li2021prefix}, prompt learning methods, and so on. These approaches have shown tremendous potential in tasks such as open-domain question answering, continual dialogue systems, and cross-domain text generation~\cite{wang2017rtheta3,yang2019end,li2024reinforcement}.

The rapid development of MLLMs and the in-depth integration of CL research have provided new perspectives for the exploration of the frontier in the field of artificial intelligence~\cite{dong2024internlm,liu2024multi,zhang2024caption,li2024membership,guo2024llava,hadsell2020embracing,guha2024diminishing,garg2025poet,deng2024k2}. A key research challenge in this domain is how to efficiently retain knowledge from previous tasks while learning new ones while maintaining cross-modal collaboration capabilities~\cite{roth2024practitioner,zhang2023citb,panagopoulou2023x}. This has become a central research question in the field. Building on existing research, this paper provides a systematic review and summary of the research on continual learning in multimodal large models. It delves into the innovations in model structure and methods, including the design of various model frameworks, dynamic parameter adjustment mechanisms, and modules that support task adaptation~\cite{le2024mixture,jha2024clap4clip,chen2024dual,vesdapunt2025hvclip}. These techniques not only significantly mitigate the problem of catastrophic forgetting, but also effectively enhance the task adaptability and generalization ability of MLLMs. In addition, this paper also introduces existing benchmarks for evaluating continual learning in multimodal large models, which provide important support for assessing model performance in continual learning tasks~\cite{chen2024coin,srinivasan2022climb,cao2024continual,tang2024vilco}. Research on continual learning in multimodal large models not only provides new technological means for the dynamic adaptation of cross-modal tasks, but also offers innovative solutions for complex tasks in real-world domains such as intelligent education, healthcare, and robotic interaction~\cite{he2024towards,panagopoulou2023x,wang2024freeze,qi2024interactive}.

Finally, this paper offers a forward-looking discussion on the challenges and future development trends of continual learning in multimodal large models, covering aspects such as catastrophic forgetting, the improvement and standardization of evaluation benchmarks, and the enhancement of interpretability and transparency in multimodal large model continual learning. Through these discussions, the paper aims to provide valuable research insights for scholars in the field and promote the further development and application of continual learning technologies in multimodal large models.

\begin{table*}[htbp]
\small
\renewcommand\arraystretch{1.2}
  \centering
  \caption{Innovations in MLLM Frameworks.}
    \begin{tabularx}{\textwidth}{>{\centering\arraybackslash}m{2cm}|p{0.4\textwidth}|p{0.4\textwidth}}
    \hline
   \multicolumn{1}{c|}{MLLMs} & \multicolumn{1}{c|}{Starting point of the problem} & \multicolumn{1}{c}{How to solve} \\
   
   \hline
   \multirow{2}{*}{\textbf{MaVEn}~\cite{jiang2024maven}} & 
Enhancing the image visual understanding of MLLMs.& MaVEn proposes an effective multi-granularity hybrid visual encoding framework.\\
   
   \hline
   \multirow{2}{*}{\textbf{MoVA}~\cite{zong2024mova}} & 
No single visual encoder can dominate the understanding of various image contents.& MoVA incorporates coarse-grained context-aware expert routing and fine-grained expert fusion.\\

   \hline
   \multirow{2}{*}{\textbf{MoME}~\cite{shen2024mome}} & 
The performance of general-purpose MLLMs is typically inferior to that of expert MLLMs. & MoME combines the MoVE and the MoLE to reduce task interference.\\

   \hline
   \multirow{2}{*}{\textbf{Meteor}~\cite{lee2024meteor}} & 
The performance gap of MLLMs in understanding and answering complex questions.& Meteor introduced the new concept of "traversal of rationales."\\

    \hline
   \multirow{2}{*}{\textbf{CORY}~\cite{ma2024coevolving}} & 
The stability and performance issues MLLMs encounter in RL fine-tuning.& CORY leverages the inherent cooperative evolution and emergence capabilities of multi-agent systems.\\

    \hline
   \multirow{2}{*}{\textbf{Lumen}~\cite{jiao2024lumen}} & 
MLMs overlook the intrinsic characteristics of different visual tasks.& Lumen enhances multimodal understanding by separating task-agnostic and task-specific learning.\\

    \hline
   \multirow{2}{*}{\textbf{Octopus}~\cite{zhaooctopus}} & 
MLLMs combine visual recognition and understanding sequentially at the LLM, which is suboptimal.& Octopus proposed the "Parallel Recognition → Sequential Understanding" MLLM framework.\\

    \hline
   \multirow{2}{*}{\textbf{Wings}~\cite{zhang2024wings}} & 
MLLMs tend to forget knowledge acquired from text-only instructions during training.& Wings introduces additional modules and mechanisms to compensate for attention shifts.\\

    \hline
   \multirow{2}{*}{\textbf{Cantor}~\cite{gao2024cantor}} & 
The "hallucination" problem in decision-making is caused by insufficient visual information.& Cantor inspires a multimodal chain-of-thought of MLLM.\\

    \hline
   \multirow{2}{*}{\textbf{AutoM3L}~\cite{luo2024autom3l}} & 
The limitations of automation in multimodal machine learning.& AutoM3L proposes an automated multimodal machine learning framework with MLLMs.\\

    \hline
   \multirow{2}{*}{\textbf{DI-MML}~\cite{fan2024detached}} & 
The modality competition issue in multimodal learning.& DI-MML proposes detached and interactive multimodal learning.\\

    \hline
   \multirow{2}{*}{\textbf{MEM}~\cite{liu2024multimodal}} & 
Data scraped from networks may leak personal privacy.& MEM optimizes by combining image noise and text triggers to mislead the model into learning shortcuts.\\

    \hline
   \multirow{2}{*}{\textbf{CREAM}~\cite{zhang2024cream}} & 
The lack of cross-page interaction support in document visual question answering.& CREAM proposes Coarse-to-Fine retrieval and multi-modal efficient tuning for document VQA.\\

    \hline
   \multirow{3}{*}{\textbf{SLUDA}~\cite{zheng2024self}} & 
Insufficient labeled data and the underutilization of unlabeled data.& SLUDA generates fine-grained data, optimizes unlabeled data usage, and employs adaptive selection and dynamic threshold strategies. \\

    \hline
   \multirow{2}{*}{\textbf{SAM}~\cite{wu2024semantic}} & 
The semantic alignment issue in MLLMs when processing multi-image instructions.& SAM enhances image-semantic associations through a bidirectional semantic guidance mechanism. \\

    \hline
   \multirow{3}{*}{\textbf{CTVLMs}~\cite{lu2024collaborative}} & 
Improving performance and reducing computational resource demands in MLLMs for multimodal tasks.& CTVLMs use knowledge distillation and multimodal alignment to transfer knowledge from large models to smaller ones.\\

    \hline
   \multirow{2}{*}{\textbf{Bloom}~\cite{kim2024efficient}} & 
Reducing the high computational cost of large-scale multilingual visual data modeling.& Bloom proposes pre-training with discretized visual speech representation.\\

    \hline
   \multirow{4}{*}{\shortstack{\textbf{MA-AGIQA} \\ \cite{wang2024large}}}& 
The quality evaluation issue of AI-generated images (AGIs).& MA-AGIQA combines multimodal models and traditional DNNs, utilizing semantic information extraction and the mixture of experts (MoE) structure to dynamically integrate quality-aware features.\\

    \hline
   \multirow{2}{*}{\makecell{\textbf{WorldGPT}\\~\cite{ge2024worldgpt}}} & Enhancing the applicability and generalization ability of MLLMs.& WorldGPT includes memory offloading, knowledge retrieval, and a Context Reflector.\\

    \hline
   \multirow{2}{*}{\makecell{\textbf{Q-ALIGN}\\~\cite{wu2023q}}} & 
Enhancing the applicability and generalization ability of MLLMs.& Q-ALIGN unifies IQA, IAA, and VQA tasks to enhance the model's cross-task generalization ability.\\

    \hline
   \multirow{2}{*}{\textbf{Flextron}~\cite{cai2024flextron}} & 
The deployment challenges of MLLMs in resource-constrained environments.& Flextron selects different sub-models or sub-networks by using routers.\\

    \hline
   \multirow{2}{*}{\makecell{\textbf{NExT-GPT}\\~\cite{wu2023next}}} & 
Existing MLLMs can only understand the input modality.& NExT-GPT proposes lightweight alignment techniques and modality-switching instruction tuning.\\

    \hline
    \end{tabularx}
  \label{MLLM_frame}%
  \vspace{-5mm} 
\end{table*}%

\section{Multimodal Large Language Model}

\subsection{Preliminary}
In this section, we provide an overview of the latest research on MLLMs, including various model innovation strategies, a range of benchmarks, and the application of MLLMs in diverse domains.

\begin{table*}[htbp]
\small
\renewcommand\arraystretch{1.2}
  \centering
  \caption{Innovations in MLLM Methods.}
    \begin{tabularx}{\textwidth}{>{\centering\arraybackslash}m{2cm}|p{0.4\textwidth}|p{0.4\textwidth}}
    \hline
   \multicolumn{1}{c|}{Method} & \multicolumn{1}{c|}{Starting point of the problem} & \multicolumn{1}{c}{How to solve} \\
   
   \hline
   \multirow{2}{*}{\makecell{\textbf{DenseFusion}\\~\cite{li2024densefusion}}} & 
Enhancing the visual perception ability of MLLMs.& DenseFusion proposes a multimodal perception fusion method that integrates visual experts.\\
      
   \hline
   \multirow{2}{*}{\textbf{E2E-MFD}~\cite{zhang2024e2e}} & 
The complex training process hinders the broader application of MLLMs.& E2E-MFD proposes a novel end-to-end algorithm for multimodal fusion detection.\\
      
   \hline
   \multirow{2}{*}{\textbf{NAM}~\cite{fangtowards}} & 
Neuron attribution in MLLMs has not been fully explored yet.& NAM proposes a neuron attribution method tailored for MLLMs.\\
      
   \hline
   \multirow{2}{*}{\textbf{CODE}~\cite{kim2024code}} & 
Addressing the hallucination problem in MLLMs when generating visual content.& CODE utilizes self-generated descriptions as contrastive references to adjust the information flow.\\
      
   \hline
   \multirow{2}{*}{\makecell{\textbf{MULTEDIT}\\~\cite{basu2024understanding}}} & 
To correct errors and insert new information. & MULTEDIT introduces a multimodal causal tracking method.\\
      
   \hline
   \multirow{2}{*}{\textbf{QSLAW}~\cite{xie2024advancing}} & 
Tackling the resource consumption issue faced by MLLMs in visual-language instruction tuning. & QSLAW learns group scale factors of quantized weights and adopts multimodal pretraining method.\\
      
   \hline
   \multirow{2}{*}{\textbf{LECCR}~\cite{wang2024multimodal}} & 
To improve the quality of cross-modal alignment.& LECCR proposes the MLLM-enhanced cross-lingual, cross-modal retrieval method.\\
      
   \hline
   \multirow{2}{*}{\textbf{ERL-MR}~\cite{han2024erl}} & 
To address the modality imbalance problem in MLLMs.& ERL-MR uses Euler transformations and multimodal constraint loss.\\
      
   \hline
   \multirow{2}{*}{\textbf{AMMPL}~\cite{wu2024adaptive}} & 
Enhancing the model's performance and reasoning ability. & AMMPL proposes an adaptive multimodal prompt learning method.\\
      
   \hline
   \multirow{2}{*}{\textbf{PaRe}~\cite{cai2024enhancing}} & 
Enhancing the model's performance and reasoning ability. & PaRe progressively generates intermediate modalities and replaces modality-agnostic fragments.\\
      
   \hline
   \multirow{2}{*}{\textbf{MCL}~\cite{liimproving}} & 
Addressing the insufficient interaction problem when handling complex multimodal scenarios. & MCL proposes the multimodal combination learning (MCL) method.\\
      
   \hline
   \multirow{2}{*}{\textbf{FARE}~\cite{schlarmann2024robust}} & 
MLLMs are vulnerable to adversarial attacks in the visual modality. & FARE proposes the unsupervised adversarial fine-tuning scheme.\\
      
   \hline
   \multirow{2}{*}{\textbf{DICL}~\cite{huang2023machine}} & 
Reducing the reliance on manual annotations. & DICL leverages MLLMs knowledge to enhance the robustness of visual models.\\
      
   \hline
   \multirow{2}{*}{\textbf{API}~\cite{yu2025attention}} & 
Addressing the limitations of traditional visual prompting techniques. & API enhances model perception through attention heatmaps guided by text queries.\\
      
   \hline
   \multirow{2}{*}{\textbf{IVTP}~\cite{huang2025ivtp}} & 
Addressing the high computational cost problem in MLLMs. & IVTP proposeS the instruction-guided visual token pruning method.\\

   \hline
   \multirow{2}{*}{\makecell{\textbf{ChatTracker}\\~\cite{sun2024chattracker}}} & 
Enhancing the tracking performance of MLLM trackers.& ChatTracker proposes a novel reflection-based prompt optimization module.\\
   
   \hline
   \multirow{2}{*}{\makecell{\textbf{Optimus-1}\\~\cite{li2024optimus}}} & 
Current general agents lack the necessary world knowledge and multimodal experience.& Optimus-1 proposes a hybrid multimodal memory module.\\
   
   \hline
   \multirow{2}{*}{\textbf{CuMo}~\cite{li2024cumo}} & 
Improving the performance of MLLMs on multimodal tasks.& CuMo integrates sparse gated Top-K MoE blocks in the visual encoder and MLP connectors.\\
   
   \hline
   \multirow{2}{*}{\makecell{\textbf{AcFormer}\\~\cite{liu2024visual}}} & 
The connection between visual encoders and LLMs has limitations.& AcFormer identified visual anchors and proposed a novel vision-language connector\\
   
   \hline
   \multirow{2}{*}{\makecell{\textbf{Chain-of-Sight}\\~\cite{huang2024accelerating}}} & 
Accelerating the pretraining process and improving model performance.& Chain-of-Sight captures visual details at different spatial scales through a multi-scale visual resampler.\\
   
   \hline
   \multirow{2}{*}{\makecell{\textbf{Dense Con-}\\ \textbf{nector}~\cite{yao2024dense}}} & 
Existing MLLMs underutilise the visual encoder while overly emphasising the language modality.& Dense Connector enhances the visual perception ability by integrating multi-layer visual features.\\
   
   \hline
   \multirow{2}{*}{\textbf{GCG}~\cite{wang2024weakly}} & 
In video question answering, MLLMs overlook visually relevant cues related to the question.& GCG learns to represent the temporal structure of videos and selects key frames.\\
   
   \hline
   \multirow{2}{*}{\textbf{Q-MoE}~\cite{wang2024q}} & 
Connection structure struggles with filtering visual information according to task requirements.& Q-MoE proposes a query-based hybrid expert connector.\\

    \hline
    \end{tabularx}
  \label{MLLM_method}%
  \vspace{-5mm} 
\end{table*}%

\subsection{Model Innovation}
With the continuous development of MLLMs, researchers have made various innovations in their structure, methods, and functional modules to enhance model performance, generalization ability, and adaptability. This section reviews the main innovations, which focus on three core directions: framework design, method optimization, and functional module improvements. These innovations collectively drive the performance of MLLMs in complex multimodal tasks. This section will explore the latest research advancements in these areas.

\subsubsection{Framework Innovation}
Framework innovation is the foundation of MLLM development, aiming to achieve efficient fusion and processing of cross-modal information by improving the overall architectural design. In recent years, researchers have proposed many efficient framework designs. As shown in Table \ref{MLLM_frame}, researchers have proposed several efficient framework designs, such as MaVEn, MoVA, AutoM3L, DI-MML and et. These framework innovations provide more efficient tools and methods for MLLMs to handle multimodal tasks involving language, vision, and hearing. They enable MLLMs to achieve more precise reasoning and decision-making in the interaction of multimodal data, thereby offering strong support for solving complex problems in practical applications.
More details of the innovation of MLLMs frameworks are provided in Section \ref{appendix_MLLM_MI} of the Appendix.

\subsubsection{Method Innovation}

Method innovation is the core driving force behind the performance improvement of MLLMs. By designing more efficient training methods and optimization objectives, it helps models better adapt to dynamic task environments. As shown in Table~\ref{MLLM_method}, in recent years, researchers have proposed numerous novel and efficient methods to enhance the accuracy and robustness of MLLMs. These method research has explored cutting-edge techniques such as multimodal contrastive learning, self-supervised learning objectives, and multimodal alignment mechanisms. These methods not only enhance the model's generalization ability but also significantly improve the accuracy and robustness of cross-modal tasks.
More details of the innovation of MLLMs methods are provided in Section \ref{appendix_MLLM_MI} of the Appendix.

% \subsubsection{Module Innovation}

% In terms of module innovation, researchers have focused on refining the internal design of models by improving specific modules to enhance cross-modal interaction and representational capabilities. These modules have increased the flexibility and efficiency of the models. As shown in Table~\ref{MLLM_module}, in recent years, many efficient and streamlined modules have been proposed, further boosting the overall performance of the models.These modules, through independent optimization and joint modeling of different modalities, enable the model to perform exceptionally well in tasks such as multimodal reasoning, generation, and understanding. They allow for more accurate cross-modal knowledge fusion and semantic understanding, providing crucial support for the functional expansion and practical application of MLLMs.

% More details of the innovation of MLLMs Modules are provided in Section \ref{appendix_MLLM_MI} of the Appendix.

% \begin{table*}[htbp]
% \small
% \renewcommand\arraystretch{1.2}
%   \centering
%   \caption{Innovations in Multimodal Large Model Modules.}
%     \begin{tabularx}{\textwidth}{>{\centering\arraybackslash}m{2cm}|p{0.4\textwidth}|p{0.4\textwidth}}
%     \hline
%    \multicolumn{1}{c|}{Module} & \multicolumn{1}{c|}{Starting point of the problem} & \multicolumn{1}{c}{How to solve} \\

%     \hline
%     \end{tabularx}
%   \label{MLLM_module}%
%   \vspace{-5mm} 
% \end{table*}%

\begin{table*}[htbp]
\small
\renewcommand\arraystretch{1.2}
  \centering
  \caption{Innovations in Non-LLM Unimodal CL Frameworks.}
    \begin{tabularx}{\textwidth}{>{\centering\arraybackslash}m{2cm}|p{0.4\textwidth}|p{0.4\textwidth}}
    \hline
   \multicolumn{1}{c|}{Framework} & \multicolumn{1}{c|}{Starting point of the problem} & \multicolumn{1}{c}{How to solve} \\
   
   \hline
   \multirow{2}{*}{\textbf{NTE}~\cite{benjamin2024continual}} & 
Addressing the catastrophic forgetting problem in graph neural networks.& NTE views a neural network as an ensemble of fixed experts.\\
   
   \hline
   \multirow{2}{*}{\textbf{IsCiL}~\cite{lee2024incremental}} & 
To address the issue of new data lacking labels due to annotation delays in continual learning.&IsCiL improves sample efficiency and task adaptability by incrementally learning shared skills.\\
   
   \hline
   \multirow{3}{*}{\textbf{CKP}~\cite{xu2024mitigate}} & 
To address the performance degradation caused by incorrect labels in the Lifelong Person Re-Identification task.& CKP purifies data through the CDP and ILR modules, and filters out erroneous knowledge using the EKF algorithm.\\
   
   \hline
   \multirow{2}{*}{\textbf{PBR}~\cite{liu2024prior}} & 
To reduce forgetting and enhances long-tail continual learning performance.& PBR proposes an uncertainty-guided sampling strategy and two prior-free constraints.\\
   
   \hline
   \multirow{2}{*}{\textbf{OSN}~\cite{hutask}} & 
Reducing the interference of new tasks on old tasks.& OSN explores shared knowledge between old and new tasks through parameter sharing.\\
   
   \hline
   \multirow{2}{*}{\textbf{MoDE}~\cite{lee2024becotta}} & 
Improving adaptation to new domains while preserving old knowledge.& MoDE includes domain-adaptive routing and domain-expert collaborative loss.\\
   
   \hline
   \multirow{2}{*}{\textbf{SB-MCL}~\cite{lee2024learning}} & 
To address the catastrophic forgetting problem in continual learning.& SB-MCL achieves continual learning through sequential Bayesian updates.\\
   
   \hline
   \multirow{2}{*}{\textbf{PNR}~\cite{charegularizing}} & 
Addressing the knowledge transfer and catastrophic forgetting issues.& PNR Generates pseudo-negative samples and optimizing knowledge transfer.\\
   
   \hline
   \multirow{2}{*}{\makecell{\textbf{CompoNet}\\~\cite{malagonself}}} & 
Addressing the issue of old task forgetting caused in continual reinforcement learning.& CompoNet proposes a modular neural network with linearly growing parameters.\\
   
   \hline
   \multirow{2}{*}{\makecell{\textbf{Vector-HaSH}\\~\cite{wangrapid}}} & 
To enable fast learning and continual memory.& Vector-HaSH combines hetero-associative memory and spatially invariant CNNs.\\
   
   \hline
   \multirow{2}{*}{\textbf{DDDR}~\cite{liang2025diffusion}} & 
Addressing the issue of catastrophic forgetting in federated continual learning.& DDDR uses diffusion models to generate historical data and employs contrastive learning.\\
   
   \hline
   \multirow{2}{*}{\makecell{\textbf{PromptCCD}\\~\cite{cendra2025promptccd}}} & 
Mitigating catastrophic forgetting.& PromptCCD introduces the GMP, which dynamically generates prompts to adapt to new classes.\\
   
   \hline
   \multirow{2}{*}{\textbf{Mecoin}~\cite{li2024efficient}} & 
To reduce parameter fine-tuning, lower the forgetting rate.& Mecoin employs SMU and a MeCo for efficient storage and updating of class prototypes.\\
   
   \hline
   \multirow{2}{*}{\textbf{RP2F}~\cite{sun2024incremental}} & 
Enabling effective knowledge sharing and backward knowledge transfer.& RP2F uses perturbation methods to approximate the Hessian matrix and introduces a prior.\\
   
   \hline
   \multirow{2}{*}{\makecell{\textbf{HAMMER}\\~\cite{liu2024hierarchical}}} & 
To address the catastrophic forgetting issue in multilingual text recognition.& HAMMER proposes online knowledge analysis and a hierarchical language evaluation mechanism.\\
   
   \hline
   \multirow{2}{*}{\textbf{FedCBC}~\cite{yu2024overcoming}} & 
Mitigating catastrophic forgetting.& FedCBC proposes category-specific binary classifiers and selective knowledge fusion.\\
   
   \hline
   \multirow{2}{*}{\textbf{TS-ILM}~\cite{xiaochen2024ts}} & 
Reducing information redundancy and enhancing memory retention.& TS-ILM proposes a task-level temporal pattern extractor and a time-sensitive example selector.\\
   
   \hline
   \multirow{2}{*}{\makecell{\textbf{AutoActivator}\\~\cite{li2024harnessing}}} & 
To address the issue of model forgetting old classes when continuously learning new classes.& AutoActivator dynamically adapts neural units to new tasks, enabling on-demand network expansion.\\
   
   \hline
   \multirow{2}{*}{\textbf{iNeMo}~\cite{fischer2024inemo}} & 
To achieve efficient class-incremental learning.& iNeMo proposes latent space initialization and position regularization.\\
   
   \hline
   \multirow{3}{*}{\textbf{TACO}~\cite{han2024topology}} & 
Offering a novel perspective for understanding and mitigating catastrophic forgetting.& TACO combines graph coarsening and continual learning to dynamically store information from previous tasks.\\

    \hline
    \end{tabularx}
  \label{CL_NonL_Framework}%
  \vspace{-5mm} 
\end{table*}%

\subsection{Benchmarks}
As MLLMs continue to achieve breakthroughs in multimodal tasks such as vision, language, and speech, comprehensive benchmarks have become crucial for systematically evaluating and comparing model performance. These benchmarks not only provide standardized datasets and tasks, but also define metrics for assessing models' abilities in cross-modal reasoning, generation, classification, and other areas. They play a key role in guiding research directions, identifying model limitations, and advancing technological progress. More details of the overview of MLLM benchmarks are provided in Section~\ref{appendix_benchmarks} of the Appendix. Section~\ref{appendix_benchmarks} in the Appendix introduces some of the recent representative benchmarks, covering a wide range of scenarios from academic research to practical applications, reflecting the diverse needs and challenges in the multimodal field.

\begin{table*}[htbp]
\small
\renewcommand\arraystretch{1.2}
  \centering
  \caption{Innovations in Non-LLM Unimodal CL Methods.}
    \begin{tabularx}{\textwidth}{>{\centering\arraybackslash}m{2cm}|p{0.4\textwidth}|p{0.4\textwidth}}
    \hline
   \multicolumn{1}{c|}{Method} & \multicolumn{1}{c|}{Starting point of the problem} & \multicolumn{1}{c}{How to solve} \\
   
   \hline
   \multirow{2}{*}{\textbf{GACL}~\cite{zhuang2024gacl}} & 
Addressing the catastrophic forgetting problem of models in class-incremental learning.& GACL establishes the equivalence between incremental learning and joint training.\\
   
   \hline
   \multirow{2}{*}{\textbf{C-Flat}~\cite{zhuang2024gacl}} & 
Addressing the balance between new task training sensitivity and memory retention.& C-Flat optimizes the flatness of the loss landscape.\\
   
   \hline
   \multirow{2}{*}{\textbf{DSGD}~\cite{fan2024dynamic}} & 
Addressing the practical deployment challenge.& DSGD uses structural and semantic information for stable knowledge distillation.\\
   
   \hline
   \multirow{2}{*}{\makecell{\textbf{VQ-Prompt}\\~\cite{jiao2024vector}}} & 
To improve continual learning performance.& VQ-Prompt utilizes vector quantization to achieve end-to-end optimization of discrete prompt selection.\\
   
   \hline
   \multirow{2}{*}{\makecell{\textbf{RanDumb}\\~\cite{prabhurandom}}} & 
Exploring whether the representations generated by continual learning algorithms are truly effective.& RanDumb uses random transformations and linear classifiers to address.\\
   
   \hline
   \multirow{2}{*}{\textbf{IWMS}~\cite{csabalabel}} & 
The label delay issue in online continual learning.& IWMS prioritizes the memory of samples similar to new data.\\
   
   \hline
   \multirow{2}{*}{\textbf{PPE}~\cite{li2024progressive}} & 
To address the catastrophic forgetting problem in non-sample online continual learning.& PPE learns class prototypes during the online learning phase.\\
   
   \hline
   \multirow{2}{*}{\textbf{GPCNS}~\cite{yang2024introducing}} & 
Improving the performance of continual learning.& GPCNS enhances plasticity by utilizing gradient information from old tasks.\\
   
%    \hline
%    \multirow{2}{*}{\makecell{\textbf{Bayesian}\\ \textbf{Adaptation}~\cite{thapabayesian}}} & 
% Improving the performance of continual learning.& Non-parametric Bayesian adapts the width through a conjugate Bernoulli process.\\
   
   \hline
   \multirow{2}{*}{\textbf{CILA}~\cite{wen2024provable}} & 
Improving the performance of continual learning.& CILA proposes an adaptive distillation coefficient and theoretical performance guarantees.\\
   
   \hline
   \multirow{2}{*}{\textbf{POCL}~\cite{wumitigating}} & 
Existing methods fail to fully leverage the inter-task dependencies.& POCL models task relationships through Pareto optimization and dynamically adjusts weights.\\
   
   \hline
   \multirow{2}{*}{\textbf{Powder}~\cite{piaofederated}} & 
Addressing the cross-task and cross-client knowledge transfer in federated continual learning.& Powder enables prompt-based dual knowledge transfer.\\
   
   \hline
   \multirow{2}{*}{\makecell{\textbf{AdaPromptCL}\\~\cite{kim2023one}}} & 
Addressing the challenge of task-specific semantic variations.& AdaPromptCL proposes dynamic semantic grouping and prompt adjustment.\\
   
   \hline
   \multirow{2}{*}{\textbf{LPR}~\cite{kim2023one}} & 
To reduce catastrophic forgetting and underfitting.& LPR adjusts the optimization geometry to balance the learning of new and old data.\\
   
   \hline
   \multirow{2}{*}{\textbf{InfLoRA}~\cite{liang2024inflora}} & 
To address the issue of forgetting old tasks when adapting to new tasks.& InfLoRA injects parameter reparameterization into pre-trained weights.\\
   
   \hline
   \multirow{2}{*}{\textbf{F-OAL}~\cite{zhuangf}} & 
To alleviate the issue of catastrophic forgetting in online class-incremental learning.& F-OAL proposes a forward online analytical learning method.\\
   
   \hline
   \multirow{2}{*}{\textbf{PRL}~\cite{shiprospective}} & 
Improving performance in non-sample class-incremental learning.& PRL aligns reserved space and latent space to adapt new class features to the reserved space.\\
   
   \hline
   \multirow{2}{*}{\textbf{CIL}~\cite{hao2024addressing}} & 
To address the issue of catastrophic forgetting.& CIL proposes the CIL-balanced classification loss and distribution margin loss.\\
   
   \hline
   \multirow{2}{*}{\textbf{DSSP}~\cite{yang2024domain}} & 
To eliminate the need for sample replay.& DSSP leverages domain sharing and task-specific prompt learning.\\
   
   \hline
   \multirow{2}{*}{\textbf{MRFA}~\cite{zhengmulti}} & 
To reduce catastrophic forgetting.& MRFA optimizes the entire layer margin by enhancing the features of review samples.\\
   
   \hline
   \multirow{2}{*}{\textbf{DARE}~\cite{jeeveswaran2024gradual}} & 
Improving the model's performance on old tasks.& DARE reduces representation drift through a three-stage training process.\\
   
   \hline
   \multirow{2}{*}{\textbf{EASE}~\cite{zhou2024expandable}} & 
To reduce catastrophic forgetting.& EASE constructs task-specific subspaces using lightweight adapters.\\

    \hline
    \end{tabularx}
  \label{CL_NonL_Method}%
  \vspace{-5mm} 
\end{table*}%

\subsection{Applications of MLLMs}

Multimodal large models (MLLMs) have emerged as a significant direction in artificial intelligence research in recent years~\cite{zhan2024anygpt,chiang2023vicuna,jiang2023motiongpt,zhang2024motiongpt,huang2023visual,li2023large,liu2024improved,mu2024embodiedgpt}. With the rapid development of technologies such as natural language processing, computer vision, and speech recognition, single-modal intelligent systems can no longer meet the increasingly complex requirements of real-world applications~\cite{park2023generative,radford2021learning,rocamonde2023vision,sun2023aligning}. Multimodal learning, by integrating different types of data inputs, simulates the diversity and complexity of human information processing, offering more comprehensive and flexible intelligent services. At the same time, with the deepening of interdisciplinary research, MLLMs will not only play a role in traditional AI tasks but will also expand into more edge domains, driving artificial intelligence from closed systems to a more open and intelligent ecosystem.
More details of the applications of MLLMs are provided in Section \ref{applications_MLLM} of the Appendix.

In summary, the application prospects of multimodal large models are vast. However, to fully unleash their potential, this requires the combined advancement of technological innovation and theoretical breakthroughs. In the future, with ongoing progress in algorithms, hardware, and cross-domain collaboration, it is expected that MLLMs will achieve more efficient and intelligent performance in a wider range of practical applications, further advancing the development of artificial intelligence.

\begin{table*}[htbp]
\small
\renewcommand\arraystretch{1.2}
  \centering
  \caption{Innovations in Non-LLM Multimodal CL Methods.}
    \begin{tabularx}{\textwidth}{>{\centering\arraybackslash}m{2cm}|p{0.4\textwidth}|p{0.4\textwidth}}
        \hline
   \multicolumn{1}{c|}{Method} & \multicolumn{1}{c|}{Starting point of the problem} & \multicolumn{1}{c}{How to solve} \\
   
   \hline
   \multirow{2}{*}{\textbf{CPP}~\cite{yuan2024continual}} & 
Improving the performance of continual learning.& CPP incorporates the CCE, TKD, and TPL mechanisms to achieve multimodal vision perception.\\
      
   \hline
   \multirow{2}{*}{\makecell{\textbf{CP-Prompt}\\~\cite{feng2024cp}}} & 
To reduce catastrophic forgetting.& CP-Prompt utilizes a dual-prompt strategy and parameter-efficient adjustments.\\
      
   \hline
   \multirow{2}{*}{\textbf{MMAL}~\cite{yue2024mmal}} & 
Reducing forgetting and enhancing incremental learning performance.& MMAL proposes the modality fusion module and MSKC module.\\
      
   \hline
   \multirow{2}{*}{\textbf{MSPT}~\cite{chen2023continual}} & 
To reduce catastrophic forgetting.& MSPT optimizes multimodal learning through gradient modulation and attention distillation.\\
      
   \hline
   \multirow{2}{*}{\makecell{\textbf{MedCoSS}\\~\cite{ye2024continual}}} & 
To reduce catastrophic forgetting.& MSPT propose a staged multimodal self-supervised learning framework that avoids modality conflicts.\\

   \hline
   \multirow{2}{*}{\textbf{ZiRa}~\cite{deng2024zero}} & 
Retaining zero-shot generalization ability.& ZiRa proposes zero-interference loss and a reparameterized dual-branch structure.\\
   
   \hline
   \multirow{2}{*}{\textbf{STELLA}~\cite{leestella}} & 
To reduce forgetting of previously learned knowledge.& STELLA proposes a localized patch importance scoring method.\\
   
   \hline
   \multirow{2}{*}{\makecell{\textbf{RCS-Prompt}\\~\cite{yang2024rcs}}} & 
To address the issue of overlap between old and new category spaces.& RCS-Prompt proposes bidirectional prompt optimization and prompt magnitude normalization.\\
   
   \hline
   \multirow{2}{*}{\textbf{ZSCL}~\cite{zheng2023preventing}} & 
To reduce catastrophic forgetting.& ZSCL proposes feature space distillation and parameter space weight integration.\\
   
   \hline
   \multirow{3}{*}{\textbf{CoCoOp}~\cite{zhou2022conditional}} & 
To address the issue of pretrained models lacking generalization ability to unseen classes when adapting to new tasks.& CoCoOp generates dynamic prompts using a lightweight neural network.\\
   
   \hline
   \multirow{2}{*}{\textbf{RAIL}~\cite{xu2024advancing}} & 
Improving cross-domain classification capabilities during continual learning.& RAIL uses recursive ridge regression and a no-training fusion module.\\

    \hline
    \end{tabularx}
  \label{CL_NonLM_Method}%
  \vspace{-5mm} 
\end{table*}%

\section{Continue Learning}

\subsection{Preliminary}

Continual Learning (CL) has become a central focus in AI research due to the rapid growth of deep learning and LLMs~\cite{li2017learning,loo2020generalized,pellegrini2021continual,sarfraz2023sparse,abbasi2022sparsity,huang2024etag,huang2024kfc,ke2020continual,yu2020semantic}. The challenge is to enable models to retain and enhance learning capabilities when faced with continuously changing data and tasks. Traditional methods assume that models can learn all tasks at once and maintain a fixed knowledge base, but in reality, data and tasks evolve, often leading to ``Catastrophic Forgetting''~\cite{chaudhry2018efficient,chen2020mitigating,de2021continual,miao2021continual,pham2021dualnet,konishi2023parameter,li2023memory,li2023variational}. Therefore, CL, as a learning paradigm that better aligns with real-world application needs, aims to enable models to effectively accumulate and update knowledge across multiple stages, thereby better adapting to dynamic and evolving environments.

This section will provide a detailed classification and overview of the latest innovative research in continual learning. The specific content is divided into three parts: 1) Exploring non-LLMs unimodal continual learning and focusing on traditional models' continual learning research in unimodal data; 2) Analyzing non-LLMs multimodal continual learning and discussing the challenges and research in continual learning across multi-modal data; 3) Analyzing and summarizing the latest advancements in continual learning for LLMs and examining the unique challenges and solutions they face when handling large-scale textual data.

% \subsection{Non-LLM Unimodal Continual Learning}
\subsection{Non-LLM Unimodal CL}
In traditional unimodal learning, research on continual learning primarily focuses on how to prevent models from forgetting previously learned knowledge when learning new tasks. Many researchers have proposed solutions to this problem, including strategies based on knowledge retention, incremental learning methods, and improvements to neural network architectures~\cite{shin2017continual,tao2020topology,wang2021training,sun2023decoupling,sun2021ilcoc,rypesc2024divide,sarfraz2023sparse,shi2023prototype}. For non-large models, the challenges of continual learning are particularly pronounced due to limitations in computational resources. Furthermore, the unimodal continual learning for non-large models primarily focuses on individual modalities such as vision, speech, and text. As show in Tables~\ref{CL_NonL_Framework} and~\ref{CL_NonL_Method}, to address the specific characteristics of these tasks, researchers have proposed a variety of innovative frameworks and methods. Overall, unimodal continual learning with non-large models has made significant progress in scenarios with limited computational resources. Many innovative frameworks and methods have been developed to effectively mitigate catastrophic forgetting. However, how to scale these approaches to multimodal and large-scale data remains an important direction for future research.
More details of the non-LLM unimodal continual learning are provided in Section \ref{appendix_cl_nlu} of the Appendix.

% \subsection{Non-LLM Multimodal Continual Learning}
\subsection{Non-LLM Multimodal CL}
Compared to unimodal continual learning, multimodal continual learning presents more complex challenges. Data from different modalities often exhibit heterogeneity, and the key difficulty in multimodal continual learning for non-large models lies in how to effectively fuse information across modalities while retaining previously acquired knowledge during the process of learning new modalities. In recent years, researchers have proposed various methods to address these challenges, including inter-modal collaborative learning, shared and independent representations for each modality, and others~\cite{kirkpatrick2017overcoming,rebuffi2017icarl,ahn2019uncertainty,zenke2017continual,yoon2017lifelong,lee2020neural,madaan2021representational,cossu2024continual,fini2022self,yoon2023continual,yan2022generative,pian2023audio,mo2023class}. As shown in Table~\ref{CL_NonLM_Method}, these innovative methods enable non-large models to perform continual learning in multimodal environments, while minimizing knowledge conflicts between different modalities.
More details of the non-LLM multimodal continual learning are provided in Section \ref{appendix_cl_nlm} of the Appendix.

\begin{table*}[htbp]
\small
\renewcommand\arraystretch{1.2}
  \centering
  \caption{Innovations in LLM Instruction Fine-tuning Methods.}
    \begin{tabularx}{\textwidth}{>{\centering\arraybackslash}m{2cm}|p{0.4\textwidth}|p{0.4\textwidth}}
    \hline
   \multicolumn{1}{c|}{Method} & \multicolumn{1}{c|}{Starting point of the problem} & \multicolumn{1}{c}{How to solve} \\
   
   \hline
   \multirow{2}{*}{\makecell{\textbf{ConTinTin}\\~\cite{yin2022contintin}}} & 
To reduce catastrophic forgetting.& InstructionSpeak learns from negative outputs and revisites the instructions of previous tasks.\\
      
   \hline
   \multirow{2}{*}{\textbf{OLoRA}~\cite{wang2023orthogonal}} & 
Improving the performance of continual learning.& OLoRA introduces orthogonal low-rank adaptation for CIT.\\
      
   \hline
   \multirow{2}{*}{\textbf{DAPT}~\cite{zhao2024dapt}} & 
To reduce catastrophic forgetting.&DAPT proposes a dual-attention learning and selection module.\\
      
   \hline
   \multirow{2}{*}{\textbf{ELM}~\cite{jang2023exploring}} & 
To reduce catastrophic forgetting.& ELM trains a small expert adapter for each task on top of the LLM.\\
      
   \hline
   \multirow{2}{*}{\makecell{\textbf{LLaMA PRO}\\~\cite{wu2024llama}}} & 
Retaining the initial functionality through post-training. & LLaMA PRO introduces an innovative block expansion technique.\\
      
   \hline
   \multirow{3}{*}{\makecell{\textbf{AdaptLLM}\\~\cite{cheng2023adapting}}} & 
To help the model leverage domain-specific knowledge while enhancing prompt performance. & AdaptLLM adapts the LLM to different domains by enriching the original training corpus with a series of content-related reading comprehension tasks.\\
      
   \hline
   \multirow{2}{*}{\textbf{DynaInst}~\cite{mok2023large}} & 
To enhance the generalization of the LLM.& DynaInst combines dynamic instruction replay with a local minima-inducing regularizer.\\

   \hline
   \multirow{2}{*}{\textbf{TAALM}~\cite{seo2024train}} & 
Enabling targeted knowledge updates and reducing forgetting.& TAALM uses meta-learning to dynamically predict token importance.\\
   
   \hline
   \multirow{2}{*}{\makecell{\textbf{D-CPT Law}\\~\cite{que2024d}}} & 
To reduce GPU resource consumption and improve domain adaptability.& D-CPT Law predicts the optimal training ratio.\\
   
   \hline
   \multirow{2}{*}{\textbf{COPAL}~\cite{malla2024copal}} & 
High computational demands and model adaptability limitations. & COPAL enables continual pruning without the need for retraining.\\
   
   \hline
   \multirow{2}{*}{\textbf{MagMax}~\cite{marczak2025magmax}} & 
To reduce catastrophic forgetting. & MagMax proposes sequential fine-tuning and maximum magnitude weight selection.\\
   
   \hline
   \multirow{2}{*}{\textbf{SAPT}~\cite{zhao2024sapt}} & 
Enabling effective knowledge retention and transfer.& SAPT aligns the learning and selection of PET blocks through a shared attention mechanism.\\
   
   \hline
   \multirow{2}{*}{\textbf{SSR}~\cite{huang2024mitigating}} & 
To reduce catastrophic forgetting.& SSR utilizes LLM-generated synthetic instances for rehearsal.\\
   
   \hline
   \multirow{2}{*}{\makecell{\textbf{LoRAMoE}\\~\cite{dou2024loramoe}}} & 
Enhancing multi-task handling capabilities.& LoRAMoE integrates LoRA and router networks, and introduces local balance constraints.\\
   
   \hline
   \multirow{2}{*}{\makecell{\textbf{F-Learning}\\ \textbf{paradigm}~\cite{dou2024loramoe}}} & 
Improving the performance of continual learning.& F-Learning paradigm first forgets old knowledge before learning new knowledge.\\

    \hline
    \end{tabularx}
  \label{CL_LLM_instr}%
  \vspace{-5mm} 
\end{table*}%

% \subsection{Continual Learning in LLM}
\subsection{CL in LLM}

LLMs such as GPT and BERT, with their powerful language understanding and generation capabilities, have achieved remarkable results on various natural language processing tasks~\cite{devlin2018bert,du2024chinese,eloundou2023gpts,kukreja2024literature,kasneci2023chatgpt,zhao2023survey,naveed2023comprehensive,chang2024survey,chen2021evaluating,unlu2023interpretutor,wu2024survey,zhang2023instruction}. However, LLMs still face unique challenges in continual learning. Particularly in the context of increasing data volume and task diversity, how to effectively update models, avoid catastrophic forgetting, and maintain efficient computational capabilities are key focuses in the research of LLMs for continual learning. As shown in Table~\ref{CL_LLM_instr}, researchers have proposed a variety of instruction fine-tuning methods. Through model improvements and methods such as instruction fine-tuning, LLMs are able to expand their knowledge while effectively addressing the issue of catastrophic forgetting. However, as model sizes continue to grow, core challenges in the field of continual learning for LLMs remain, such as how to handle updates and learning with large-scale data, and how to maintain good adaptability in multi-task and cross-modal environments. These remain critical issues that need to be addressed.
More details of the LLM continual learning are provided in Section \ref{appendix_cl_llm} of the Appendix.

Continual learning is a multidimensional and complex research field, characterized by both challenges and opportunities. From unimodal to multimodal, and then to continual learning in LLMs, each category of methods and strategies presents its own unique challenges and innovations. Future research will not only need to deepen the understanding of existing methods, but also explore how to achieve more efficient and robust continual learning in environments with large-scale, multimodal data and tasks. As computational power and data scale continue to expand, research in continual learning will provide a more solid theoretical and technological foundation for the adaptability, robustness, and sustainability of intelligent systems.

\begin{table*}[htbp]
\small
\renewcommand\arraystretch{1.2}
  \centering
  \caption{Innovations in MLModel CL Frameworks.}
    \begin{tabularx}{\textwidth}{>{\centering\arraybackslash}m{2cm}|p{0.4\textwidth}|p{0.4\textwidth}}
    \hline
   \multicolumn{1}{c|}{Framework} & \multicolumn{1}{c|}{Starting point of the problem} & \multicolumn{1}{c}{How to solve} \\
         
   \hline
   \multirow{2}{*}{\makecell{\textbf{PathWeave}\\~\cite{yu2024llms}}} & 
To reduce the dependency on large-scale joint pre-training.& PathWeave enhances modality alignment and collaboration.\\
   
   \hline
   \multirow{2}{*}{\textbf{CLAP}~\cite{jha2024clap4clip}} & 
To enhance the model's uncertainty estimation capabilities.& CLAP is compatible with various prompt methods.\\
   
   \hline
   \multirow{2}{*}{\textbf{DIKI}~\cite{tang2025mind}} & 
To reduce catastrophic forgetting.& DIKI proposes a residual mechanism and distribution-aware calibration.\\
         
   \hline
   \multirow{2}{*}{\textbf{GMM}~\cite{cao2024generative}} & 
To reduce catastrophic forgetting.& GMM implements incremental learning through generated label text and feature matching.\\
         
   \hline
   \multirow{2}{*}{\makecell{\textbf{PriViLege}\\~\cite{park2024pre}}} & 
To address catastrophic forgetting and overfitting in MLLMs.& PriViLege proposes prompt functionality and knowledge distillation.\\
         
   \hline
   \multirow{2}{*}{\makecell{\textbf{ModalPrompt}\\~\cite{zeng2024modalprompt}}} & 
To address catastrophic forgetting and overfitting in MLLMs.& ModalPrompt proposes bi-modal guided prototype prompts and knowledge transfer.\\
         
   \hline
   \multirow{2}{*}{\textbf{CGIL}~\cite{frascaroli2024clip}} & 
To reduce catastrophic forgetting.& CGIL uses VAEs to learn class-conditioned distributions and generate synthetic samples.\\
         
   \hline
   \multirow{2}{*}{\makecell{\textbf{CoLeCLIP}\\~\cite{li2024coleclip}}} & 
To  reduce interference between tasks.& CoLeCLIP proposes joint learning of task prompts and cross-domain vocabularies.\\
         
   \hline
   \multirow{2}{*}{\textbf{ICL}~\cite{qi2024interactive}} & 
To enhance the efficiency of continual learning in MLLMs.& ICL enables interaction between a fast intuition model and a slow deep thinking model.\\
         
   \hline
   \multirow{2}{*}{\textbf{EMT}~\cite{zhai2023investigating}} & 
To evaluate catastrophic forgetting in MLLMs.& EMT offers a new perspective for improving fine-tuning strategies in MLLMs.\\
         
   \hline
   \multirow{2}{*}{\makecell{\textbf{Freeze-Omni}\\~\cite{wang2024freeze}}} & 
To reduce catastrophic forgetting.& Freeze-Omni implements a three-stage training strategy.\\
         
   \hline
   \multirow{2}{*}{\textbf{Adapt-$\infty$}~\cite{maharana2024adapt}} & 
To reduce catastrophic forgetting.& Adapt-$\infty$ proposes dynamic data selection and a clustering-based permanent pruning strategy.\\
         
   \hline
   \multirow{3}{*}{\makecell{\textbf{Mono-}\\ \textbf{InternVL}~\cite{luo2024mono}}} & 
To address the performance degradation and catastrophic forgetting issues that arise when expanding the visual and language capabilities of MLLMs.& Mono-InternVL integrates visual experts using a MOE structure and introduces endogenous visual pretraining.\\
         
   \hline
   \multirow{2}{*}{\makecell{\textbf{MoExtend}\\~\cite{zhong2024moextend}}} & 
To  address the issues of catastrophic forgetting and high training costs.& MoExtend designes a three-stage training process, including alignment, extension, and fine-tuning.\\

    \hline
    \end{tabularx}
  \label{CL_MLLM_Framework}%
  \vspace{-5mm} 
\end{table*}%

\section{Continual Learning in MLLMs}

\subsection{Preliminary}

Recent advancements in MLLMs have shown remarkable capabilities across various domains. However, as their scale grows, maintaining long-term effectiveness in dynamic environments is a critical challenge~\cite{young2014image,achiam2023gpt,anil2023palm,bai2023qwen,chen2015microsoft,chen2024internvl,panagopoulou2023x,dong2024internlm,fu2024video,goyal2017making,gurari2018vizwiz,liu2024llava}. CL addresses this by enabling models to learn new tasks without forgetting previously acquired knowledge in evolving data and task contexts. For MLLMs, continual learning is more complex due to the vast data and complex computations involved, requiring significant computational resources and storage. Although existing research provides valuable theoretical and experimental insights~\cite{liu2025mmbench,luo2024cheap,yang2023dawn,team2023gemini,team2023internlm,touvron2023llama,wang2023cogvlm,wu2024parameter,yue2024mmmu}, applying MLLMs to continual learning still faces many challenges. This section explores innovations in multimodal large model continual learning and the related evaluation benchmarks.

\subsection{Model Innovation}

As shown in Tables~\ref{CL_MLLM_Framework} and~\ref{CL_MLLM_Method}, to achieve multi-task CL in multimodal large models and avoid catastrophic forgetting, researchers have proposed numerous innovative frameworks and methods~\cite{li2024coleclip,qi2024interactive,maharana2024adapt,luo2024mono,lester2021power,yan2022generative,villa2023pivot,he2024towards}. These innovations not only facilitate knowledge sharing and transfer between multiple tasks but also effectively address challenges such as catastrophic forgetting, modality conflicts, and computational resource constraints. These efforts collectively advance the continual learning capabilities of multimodal large models in dynamic environments.
More details of the model innovation in the continual learning of MLLMs are provided in Section \ref{Appendix_MLLMCL} of the Appendix.

\begin{table*}[htbp]
\small
\renewcommand\arraystretch{1.2}
  \centering
  \caption{Innovations in MLLModel CL Methods.}
    \begin{tabularx}{\textwidth}{>{\centering\arraybackslash}m{2cm}|p{0.4\textwidth}|p{0.4\textwidth}}
    \hline
   \multicolumn{1}{c|}{Method} & \multicolumn{1}{c|}{Starting point of the problem} & \multicolumn{1}{c}{How to solve} \\
         
   \hline
   \multirow{2}{*}{\textbf{NoRGa}~\cite{yu2024llms}} & 
To enhance the continual learning performance of multimodal large language models. & NoRGa proposes the non-linear residual gate.\\
         
   \hline
   \multirow{2}{*}{\textbf{ZAF}~\cite{gaostabilizing}} & 
To reduce catastrophic forgetting.& ZAF preserves knowledge through zero-shot stability regularization. \\
         
   \hline
   \multirow{3}{*}{\makecell{\textbf{DualLoRA}\\~\cite{chen2024dual}}} & 
Improving the efficiency and effectiveness of continual learning in multimodal large language models.& DualLoRA utilizes orthogonal and residual low-rank adapters along with a dynamic memory mechanism to balance model stability and plasticity. \\
         
   \hline
   \multirow{3}{*}{\textbf{LPI}~\cite{yan2024low}} & 
To address the insufficient interaction between modalities and tasks.& LPI enhances inter-modal and inter-task interactions through low-rank decomposition and contrastive learning. \\
         
   \hline
   \multirow{2}{*}{\makecell{\textbf{Model Tailor}\\~\cite{zhu2024model}}} & 
To reduce catastrophic forgetting.& Retaining most of the pre-trained parameters and  replacing a small number of fine-tuned parameters. \\
         
   \hline
   \multirow{2}{*}{\textbf{HVCLIP}~\cite{vesdapunt2025hvclip}} & 
Enhancing the model's ability to retain critical information while adapting to new tasks or domains.& HVCLIP uses strategies such as forgetting reduction, discrepancy reduction, and feature enhancement. \\
         
   \hline
   \multirow{2}{*}{\makecell{\textbf{Continual}\\ \textbf{LLaVA}~\cite{cao2024continual}}} & 
Enhancing the ability to preserve knowledge from previous tasks while accommodating new ones..& Continual LLaVA proposes a parameter-efficient tuning method that does not require rehearsal. \\
         
   \hline
   \multirow{2}{*}{\textbf{LLaCA}~\cite{qiao2024llaca}} & 
To reduce forgetting and lower computational costs.& LLaCA dynamically adjusts the EMA weights and introduces an approximation mechanism. \\
         
   \hline
   \multirow{2}{*}{\textbf{CVM}~\cite{rebillard2024continually}} & 
To reduce forgetting and improve generalization.&CVM maps the representations of small visual models to the knowledge space of a fixed LLM. \\
         
   \hline
   \multirow{2}{*}{\textbf{RE-tune}~\cite{mistretta2024re}} & 
Addressing challenges related to computational resources, data privacy, and catastrophic forgetting.& RE-tune freezes the backbone of the model and trains adapters, using text prompts to guide training. \\
         
   \hline
   \multirow{2}{*}{\textbf{CluMo}~\cite{cai2024clumo}} & 
Enhancing the performance of MLLMs in CL and improving their ability to retain old knowledge.& CluMo employs a two-stage training and modality fusion prompt strategy. \\
         
   \hline
   \multirow{2}{*}{\makecell{\textbf{Fwd-Prompt}\\~\cite{zheng2024beyond}}} & 
To achieve anti-forgetting and positive transfer.& Fwd-Prompt utilizes gradient projection techniques and proposes a multimodal prompt pool. \\
         
   \hline
   \multirow{2}{*}{\makecell{\textbf{CPE-CLIP}\\~\cite{d2023multimodal}}} & 
Enhancing the performance of few-shot class incremental learning in MLLMs.& CPE-CLIP using learnable prompts and regularization strategies. \\
         
   \hline
   \multirow{2}{*}{\textbf{TG}~\cite{zhang2024preserving}} & 
To reduce catastrophic forgetting.& TG proposes the model-agnostic self-uncompression method. \\
         
   \hline
   \multirow{2}{*}{\textbf{LiNeS}~\cite{wang2024lines}} & 
Preserving the generalization ability of pretraining while improving fine-tuning task performance. & LiNeS proposes parameter updates with differentiated layer depth. \\
         
   \hline
   \multirow{2}{*}{\textbf{AttriCLIP}~\cite{wang2023attriclip}} & 
Enhancing the generalization and continual learning capabilities of MLLMs in multimodal tasks. & AttriCLIP adapts to new tasks using an attribute lexicon and textual prompts. \\
         
   \hline
   \multirow{2}{*}{\textbf{AttriCLIP}~\cite{wang2023attriclip}} & 
Enhancing the generalization and continual learning capabilities of MLLMs in multimodal tasks. & AttriCLIP adapts to new tasks using an attribute lexicon and textual prompts. \\
         
   \hline
   \multirow{2}{*}{\textbf{C-LoRA}~\cite{smith2023continual}} & 
To reduce catastrophic forgetting.& C-LoRA performs continual adaptive low-rank adjustments in the cross-attention layers of MLLMs. \\

    \hline
    \end{tabularx}
  \label{CL_MLLM_Method}%
  \vspace{-5mm} 
\end{table*}%

\subsection{Benchmarks}

As the application of multimodal large models in continual learning increases, evaluating their CL capability has become a key issue. To comprehensively assess the continual learning performance of multimodal large models, benchmarks and evaluation frameworks have emerged. However, benchmarks specifically designed for continual learning in multimodal large models are still relatively scarce, and the relevant evaluation standards are still in the process of development. Section~\ref{MLLMCLBenchmark} in the Appendix analyzes and lists the few existing benchmarks to evaluate the continual learning capability of multimodal large models, exploring their design concepts, evaluation metrics, and applicability in different application scenarios.

Existing benchmarks for multimodal large model continual learning provide some reference value for assessing a model's learning ability. However, due to the scarcity of such benchmarks, with only a few available for use, many issues and limitations remain to be addressed. In the future, there is a need to design more comprehensive, flexible, and scalable evaluation benchmarks to meet the evolving demands of multimodal large model continual learning technologies.

% todo
% 1. 图标风格统一（格式统一）
% 2. 表格重新绘画一下
% 3. pipline

\section{Challenges and Future Trends in Multimodal Large Model Continual Learning}

\subsection{Catastrophic Forgetting}

\subsubsection{Challenges Encountered}

Catastrophic forgetting has long been a classic problem in continual learning tasks, and its presence significantly limits the adaptability and generalization ability of models in real-world dynamic environments. For multimodal large models, this issue becomes even more complex due to the need for training on large-scale data, as well as the immense computational resources and storage space required.

\subsubsection{Future Trends}

Balancing forgetting management with learning efficiency, especially as tasks increase, is a complex optimization challenge. The goal is to prevent catastrophic forgetting while maintaining learning efficiency. Future research should focus on strategies to mitigate forgetting, such as frameworks or algorithms that preserve old knowledge while learning new information, or mechanisms for periodic knowledge consolidation. In addition, techniques such as self-supervised learning and transfer learning can be utilized. By sharing latent features or representations across different modalities, these methods can reduce interference between tasks, thereby alleviating the impact of catastrophic forgetting.

\subsection{Improvement and Standardization of Evaluation Benchmarks}

\subsubsection{Challenges Encountered}

Evaluation benchmarks should not only consider a model's performance in learning new tasks but also assess its ability to retain knowledge across different modalities, the effectiveness of cross-task transfer, and its stability over long-term learning. Currently, benchmarks for evaluating continual learning in multimodal large models are still relatively scarce. As multimodal large models become increasingly complex in real-world applications, developing comprehensive and systematic evaluation benchmarks for their continual learning capabilities is an urgent problem that needs to be addressed.

\subsubsection{Future Trends}

Future research should focus on designing more comprehensive and flexible evaluation benchmarks that support the assessment of continual learning in multimodal large models within multi-task environments. Researchers need to develop evaluation metrics capable of measuring a model's performance in multi-task learning, knowledge transfer, catastrophic forgetting, and cross-modal consistency. Furthermore, the standardization of evaluation benchmarks will be a key direction for future development. By establishing unified evaluation frameworks, it will be possible to more effectively compare the strengths and weaknesses of different models, thereby advancing research in this field.

\subsection{Improving the Interpretability and Transparency of Continual Learning in Multimodal Large Models}

\subsubsection{Challenges Encountered}

In multimodal learning tasks, models need to integrate information from different modalities (such as images, text, audio, etc.), which makes their decision-making process more complex and harder to trace. In particular in continual learning environments, the model must continuously learn new tasks while retaining knowledge from previous tasks. The integration and transfer of information across different modalities during this learning process make the model's decision mechanism even more challenging to interpret. Enhancing the interpretability of multimodal large models in continual learning not only helps increase the model's trustworthiness but also provides effective debugging and error diagnosis mechanisms during the learning process.

\subsubsection{Future Trends}

In future research on continual learning for multimodal large models, to enhance model interpretability, researchers can design more transparent and traceable architectures that allow for clear tracking and analysis of the model's decision-making rationale when handling different tasks. At the model design level, researchers can integrate the latest advances in explainable AI (XAI) to incorporate highly interpretable model structures, thus improving transparency in the decision-making process. Furthermore, by combining techniques such as cross-modal learning and transfer learning, researchers can effectively facilitate the transfer and retention of cross-task knowledge during continual learning, while also enhancing the understanding and explainability of the knowledge transfer mechanisms.

\section{Conclusion}

In this review, we systematically discuss the latest advancements and challenges in the continual learning of multimodal large models (MLLM). First, we review the innovative strategies of multimodal large models and their applications across different fields, highlighting their advantages in handling diverse data sources. We also introduce the most commonly used benchmark testing methods and provide application examples in various domains such as natural language processing and computer vision. 

Next, we provide a detailed overview of the latest research in continual learning, offering a classification of unimodal and multimodal continual learning in non-large models, and delving into the current state of research on large language models (LLMs) in continual learning. By comparing research across these different areas, we further clarify their approaches and limitations in dealing with data distribution changes. 

The extensive and in-depth research in both the multimodal large model and continual learning domains has laid a solid foundation for research in multimodal large model continual learning. We conduct a thorough analysis of the current state of research in this area, discussing aspects such as benchmark evaluation, model structures, and innovations in methods, revealing both the potential and the challenges faced by MLLM in continual learning. 

Finally, we provide a forward-looking discussion on the challenges and future development trends in the continual learning of multimodal large models. Our goal is to inspire researchers in the field and provide valuable insights for future research directions, aiming to promote the advancement and innovation of technologies related to the continual learning of multimodal large models.

\small
\bibliographystyle{IEEEtran}
\bibliography{ref}

\clearpage

\label{appendix}
% \section{Appendix}
\section{Multimodal Large Language Model}
\label{Appendix_MLLM}
\subsection{Model Innovation}

\label{appendix_MLLM_MI}

\subsubsection{Framework Innovation}

Chaoya Jiang et al.~\cite{jiang2024maven} introduced the multi-granularity hybrid visual encoding framework MaVEn, which combines discrete visual symbol sequences representing abstract, coarse-grained semantic concepts with traditional continuous representation sequences that simulate fine-grained features. This combination enhances the model's ability to understand visual information in images.

Zhuofan Zong et al.~\cite{zong2024mova} proposed the MoVA framework, which incorporates coarse-grained context-aware expert routing and fine-grained expert fusion. This framework adaptively routes and fuses visual experts for specific tasks through a coarse-to-fine mechanism, thereby mitigating the bias of the CLIP visual encoder and enhancing the model's ability to understand and process diverse image content.

Leyang Shen et al.~\cite{shen2024mome} proposed a multimodal expert mixing framework, MoME, which combines the visual expert mixture model (MoVE) and the language expert mixture model (MoLE) to reduce task interference.

Byung-Kwan Lee et al.~\cite{lee2024meteor} proposed the Meteor model, based on the Mamba architecture, which enhances the comprehension and response capabilities of large language and vision models through multifaceted reasoning.

Hao Ma et al.~\cite{ma2024coevolving} proposed the sequential cooperative multi-agent reinforcement learning framework, CORY, which enhances the stability and performance of multimodal large models in reinforcement learning fine-tuning by leveraging the inherent collaborative evolution and emergent capabilities of multi-agent systems.

Yang Jiao et al.~\cite{jiao2024lumen} proposed a vision-centric multimodal large model framework, Lumen, which strengthens multimodal content understanding by decoupling task-agnostic and task-specific learning. This framework enables flexible adaptation to various vision tasks, enhancing the LMM's capabilities in visual perception and instruction following.

Chuyang Zhao et al.~\cite{zhaooctopus} proposed the "Parallel Recognition → Sequential Understanding" MLLM framework, Octopus. This framework achieves parallel recognition of object queries at the lower LLM layers and passes the results to the top LLM layers for sequential understanding, thereby improving the efficiency and accuracy of MLLMs.

Yikai Zhang et al.~\cite{zhang2024wings} proposed the Wings framework, which introduces additional modules and mechanisms to compensate for attention shifts. This allows the model to effectively process visual information while maintaining focus on textual information.

Timin Gao et al.~\cite{gao2024cantor} proposed the Cantor framework, which integrates visual inputs with logical reasoning and leverages the advanced cognitive functions of MLLMs. By acting as a multifaceted expert, it directly acquires higher-level information, thereby improving decision-making quality.

Daqin Luo et al.~\cite{luo2024autom3l} proposed the AutoM3L framework, based on the AutoML architecture, which automates the construction of multimodal training pipelines, feature engineering, and model selection using LLMs, thereby reducing manual intervention.

Yunfeng Fan et al.~\cite{fan2024detached} proposed the DI-MML framework, which addresses modality competition in multimodal learning by independently training modality encoders. They introduced a shared classifier and DUC loss to facilitate cross-modal interaction and knowledge transfer, thereby mitigating the modality competition issue in multimodal learning.

Xinwei Liu et al.~\cite{liu2024multimodal} proposed the multi-step error minimization framework, MEM, which optimizes by combining image noise and text triggers. This approach misleads the model into learning shortcuts, thereby protecting data privacy.

Jinxu Zhang et al.~\cite{zhang2024cream} proposed the CREAM framework, which integrates high-performance retrieval enhancement, multi-image and multimodal processing, and efficient instruction tuning. This effectively addresses the challenges in document-based VQA tasks.

Li Zheng et al.~\cite{zheng2024self} proposed the Adaptive Multimodal Data Augmentation framework, SLUDA, which generates fine-grained data, optimizes the utilization of unlabeled data, and employs adaptive selection strategies and dynamic threshold adjustments. This approach addresses the issues of insufficient labeled data and the underutilization of unlabeled data.

Tao Wu et al.~\cite{wu2024semantic} proposed the SAM model, which enhances semantic associations between images by introducing a bidirectional semantic guidance mechanism. This improves the semantic alignment ability of multimodal instructions.

Shichen Lu et al.~\cite{lu2024collaborative} proposed the Tiny-Large collaborative training framework, CTVLMs, which leverages knowledge distillation and multimodal alignment to enable large models to transfer knowledge to smaller models. This approach achieves a dual improvement in both performance and efficiency.

Minsu Kim et al.~\cite{kim2024efficient} proposed the Bloom framework, which uses bidirectional modality transformation and adaptive cross-modal fusion. It pretrains a VSR (Visual Speech Recognition) model with visual and speech units and introduces a curriculum learning strategy to enhance training efficiency and multilingual recognition performance.

Yunshan Ma et al.~\cite{ma2024cirp} proposed the CIRP framework, which uses a multimodal encoder and cross-item contrastive loss to learn individual item semantics and relationships. By introducing a relationship pruning module, this framework enhances the ability to align cross-modal information and capture cross-item relationships in cold-start items.

Puyi Wang et al.~\cite{wang2024large} proposed the multimodal large model-assisted artificial intelligence-generated image quality assessment framework, MA-AGIQA. By combining multimodal models with traditional DNNs, and utilizing semantic information extraction and a mixture of experts (MoE) structure, the framework dynamically integrates quality perception features. This significantly improves the quality assessment performance of AGIs, particularly excelling in reducing the false-negative rate.

Zhiqi Ge et al.~\cite{ge2024worldgpt} proposed a novel cognitive framework, WorldGPT, which includes memory offloading, knowledge retrieval, and a Context Reflector to enhance the model's performance in specific scenarios and long-term tasks.

Haoning Wu et al.~\cite{wu2023q} proposed the ONEALIGN model, which unifies IQA, IAA, and VQA tasks, thereby enhancing the model's cross-task generalization ability.

Zixin Zhang et al.~\cite{zhangenhancing} proposed the M2FEDSA framework, which combines segmentation learning and multimodal federated learning. By introducing dual-adaptive fine-tuning and dual knowledge transfer strategies, the framework improves both computational and storage efficiency, as well as performance, when deploying large-scale multimodal models in federated learning settings.

Ruisi Cai et al.~\cite{cai2024flextron} proposed an elastic architecture called Flextron, which supports adaptive subnetwork selection. By using routers to choose different sub-models or subnetworks, Flextron addresses the deployment challenges of multimodal large models in resource-constrained environments.

Shengqiong Wu et al.~\cite{wu2023next} proposed an end-to-end Any-to-Any multimodal large model framework, which achieves efficient cross-modal understanding and generation through lightweight alignment techniques and modality-switching instruction tuning.

\subsubsection{Method Innovation}

Xiaotong Li et al.~\cite{li2024densefusion} proposed a comprehensive multimodal perception fusion method that integrates visual experts, thereby enhancing the visual perception capability of MLLMs.

Jiaqing Zhang et al.~\cite{zhang2024e2e} proposed a novel end-to-end algorithm for multimodal fusion detection, achieving high performance through a single training phase and simplifying the overall process.

Junfeng Fang et al.~\cite{fangtowards} proposed a neuron attribution method tailored for MLLMs, called NAM. NAM reveals the modality-specific semantic knowledge learned by neurons in MLLMs and highlights certain neuron characteristics that collectively elucidate the internal workings of MLLMs.

Jayneel Parekh et al.~\cite{parekh2024concept} proposed a concept extraction method based on dictionary learning to interpret the internal representations of large multimodal models. They innovatively defined multimodal concepts and validated their effectiveness in interpreting models and understanding test sample representations.

Junho Kim et al.~\cite{kim2024code} proposed CODE, which utilizes self-generated descriptions as contrastive references to dynamically adjust the information flow, enhancing the coherence and informativeness of responses. This approach addresses the hallucination problem in MLLMs when generating visual content.

Samyadeep Basu et al.~\cite{basu2024understanding} proposed the model editing algorithm MULTEDIT, which can correct errors and insert new information. They also introduced a multimodal causal tracking method, extending the research on information storage to other domains.

Jingjing Xie et al.~\cite{xie2024advancing} proposed the Quantized Scale Learning Method (QSLAW), which effectively reduces quantization errors, prevents overfitting, and improves model adaptability and efficiency by learning the group scale factors of quantized weights and employing a multimodal pretraining strategy.

Yabing Wang et al.~\cite{wang2024multimodal} proposed the MLLM-enhanced cross-lingual, cross-modal retrieval method LECCR. This approach leverages MLLMs to generate visual descriptions, which are then aggregated into multi-view semantic slots to enhance the semantic richness of visual features. By incorporating English feature guidance, it improves the quality of cross-modal alignment.

Zihao Liu et al.~\cite{liu2024adaptively} proposed a visual perception adapter and fine-grained tri-modal contrastive learning method. By aligning tokens across modalities, they reduce semantic gaps, thereby improving the performance of multimodal video tasks.

Weixiang Han et al.~\cite{han2024erl} proposed the ERL-MR strategy, which uses Euler transformations and multimodal constraint loss to transform inter-modal competition into cooperation, thereby achieving performance improvement.

Qiang Wang et al.~\cite{wang2024bilateral} proposed a bilateral adaptive cross-modal fusion prompt learning paradigm, Bloom, which achieves more flexible cross-modal interaction and alignment through bidirectional modal transformation and adaptive fusion functions. This significantly enhances the performance of CLIP on a variety of generalization tasks.

Zongqian Wu et al.~\cite{wu2024adaptive} proposed an adaptive multimodal prompt learning method, AMMPL, which effectively handles meaningless image patches and enhances the model's generalization ability through image prompts and cross-modal interaction learning.

Minghe Gao et al.~\cite{gao2024fact} proposed the Fact paradigm, which teaches MLLMs by generating Faithful, Concise, and Transferable multimodal rationales, enhancing the model's performance and reasoning ability across various visual tasks.

Lincan Cai et al.~\cite{cai2024enhancing} proposed the PaRe method, which enhances the stability and transferability of cross-modal fine-tuning by progressively generating intermediate modalities and replacing modality-agnostic fragments.

Wei Li et al.~\cite{liimproving} proposed the Multimodal Combination Learning (MCL) method, which strengthens the mapping between visual and language modalities. By leveraging LLMs to automatically generate multimodal learning samples, they introduced a stacked retrieval mechanism to extract diverse multimodal information.

Christian Schlarmann et al.~\cite{schlarmann2024robust} proposed the FARE unsupervised adversarial fine-tuning scheme, which enhances the robustness of the CLIP model while preserving its original performance, without the need for retraining on downstream tasks.

Zhuo Huang et al.~\cite{huang2023machine} proposed the DICL strategy, which leverages MLLM knowledge to enhance the robustness of visual models and align MLLMs with visual tasks. This approach enables unsupervised fine-tuning, improving performance in out-of-distribution (OOD) scenarios.

Runpeng Yu et al.~\cite{yu2025attention} proposed the API technique, which enhances model perception through attention heatmaps guided by text queries. This approach enables model self-reflection and integration, improving performance on visual-linguistic tasks and addressing the limitations of traditional visual prompting techniques.

Kai Huang et al.~\cite{huang2025ivtp} proposed the Instruction-guided Visual Token Pruning method (IVTP), which includes an intra-group Token Pruning (GTP) module and cross-modal instruction-guided pruning. This approach effectively reduces the number of visual tokens and lowers computational complexity, while maintaining model performance.

\subsubsection{Module Innovation}

Wenfang Yao et al.~\cite{sun2024chattracker} proposed a novel reflection-based prompt optimization module, leveraging multimodal large language models to generate high-quality language descriptions to improve tracking performance. By iteratively refining the vague and inaccurate descriptions of targets through tracking feedback, this approach addresses the issue of frequent ambiguous language descriptions in annotations.

Zaijing Li et al.~\cite{li2024optimus} proposed a hybrid multimodal memory module that transforms knowledge into a hierarchical directed knowledge graph, enabling agents to explicitly represent and learn world knowledge. Additionally, historical information is summarized into an abstract multimodal experience pool, providing agents with rich contextual learning references. This approach addresses the challenge of general agents struggling to complete long-term tasks in open-world environments.

Jiachen Li et al.~\cite{li2024cumo} enhanced model capabilities by integrating sparse gated Top-K MoE (Mixture-of-Experts) blocks in the visual encoder and MLP connectors, and by introducing MoE blocks during the visual instruction fine-tuning phase. This approach improves the performance of MLLMs on multimodal tasks.

Haogeng Liu et al.~\cite{liu2024visual} innovatively identified visual anchors and proposed a novel vision-language connector, AcFormer. By utilizing visual anchors to aggregate information, this approach significantly enhances the accuracy and computational efficiency of MLLMs.

Ziyuan Huang et al.~\cite{huang2024accelerating} proposed the Chain-of-Sight module, which captures visual details at different spatial scales through a multi-scale visual resampler. This module enables flexible expansion of the number of visual tokens after pretraining, accelerating the pretraining process while maintaining or improving model performance.

Huanjin Yao et al.~\cite{yao2024dense} proposed a new connector, the Dense Connector, which enhances the visual perception ability of MLLMs by integrating multi-layer visual features. It is characterized by high computational efficiency and ease of integration, addressing the issue of existing MLLMs underutilizing the visual encoder while overly emphasizing the language modality.

Haibo Wang et al.~\cite{wang2024weakly} designed the Gaussian Contrastive Localization (GCG) module, which learns to represent the temporal structure of videos and selects key frames relevant to the question. This approach addresses the issue in video question answering where large multimodal models neglect question-related visual cues and lack key timestamp annotations.

Hanzi Wang et al.~\cite{wang2024q} proposed a query-based hybrid expert connector, Q-MoE, which utilizes text-driven routing and an optimal expert path training strategy to achieve precise extraction and processing of task-specific visual information. This approach addresses the issue in MLLMs where the connection structure struggles to filter visual information according to task requirements in vision-language tasks.

\subsection{Benchmarks}

\label{appendix_benchmarks}

\subsubsection{ROPE: Recognition-based Object Probing Evaluation Benchmark}

\label{appendix_rope}

Despite the impressive performance of MLLMs in various downstream applications, they often encounter the issue of object hallucination~\cite{rohrbach2018object,dai2022plausible,li2023evaluating,zhang2024groundhog,zhai2023halle,liu2023mitigating,you2023ferret,zhou2023analyzing,wang2023llm}, where the model erroneously generates objects that do not exist in the image. Current benchmarks for evaluating object hallucination mainly focus on the presence of a single object category, rather than individual entities. 

Xuweiyi Chen et al.~\cite{chen2024multi} conducted a systematic study of the multi-object hallucination problem, examining how models misidentify objects when attending to multiple objects simultaneously (e.g., inventing non-existent objects or being distracted). They introduced an automated evaluation protocol called Recognition-based Object Probing Evaluation (ROPE), which considers the distribution of object categories within a single image during testing. By using visual reference to disambiguate, the protocol systematically analyzes multi-object hallucination, revealing the hallucination behaviors and influencing factors when models process multiple objects. In addition, ROPE designs multiple task prompts, including Default Multi-Object, Student-Forcing, Teacher-Forcing, and Single-Object. The dataset is divided into four subsets, each considering different object category distributions: 1) Homogeneous: All test objects belong to the same category. 2) Heterogeneous: All test objects belong to different categories. 3) In-the-Wild: A mixed object category distribution, with test objects randomly selected and ordered. 4) Adversarial: After multiple repetitions of the same category, a different category object is introduced. The dataset is further divided into Seen and Unseen based on whether the model has encountered these images during instruction tuning.

More details of the overview of MLLM performance on the  ROPE are provided in table~\ref{rope}.

\begin{table*}[t]
\small
\centering
\caption{Averaged accuracy of baselines on the \textit{In-the-Wild, \textit{Homogeneous}, and \textit{Heterogeneous} splits.} \vspace{3mm}}
\begin{threeparttable}
\begin{tabular}{l|rrr|rrr|rrr|rrr}
\hline
\multicolumn{1}{c|}{\multirow{2}[4]{*}{Model}}
& \multicolumn{3}{c|}{Default Multi-Object}
& \multicolumn{3}{c|}{Student-Forcing}            & \multicolumn{3}{c|}{Teacher-Forcing}            & \multicolumn{3}{c}{Single-Object}             \\
\cmidrule(r){2-4}\cmidrule(r){5-7}\cmidrule(r){8-10}\cmidrule(r){11-13}
% \multirow{-2}{*}{Models}
& Wild
& Hom.
& Het.
& Wild
& Hom.
& Het.
& Wild
& Hom.
& Het.
& Wild 
& Hom.
& Het. \\ 
\hline
% \vspace{0.5mm}
\multicolumn{13}{c}{\textit{Seen}} \\ 
 \hline
Yi-VL-6B~\cite{young2024yi} & 2.95         & 5.65         & 1.99         & 3.44 & 6.80 & 3.78 & 5.45 & 26.25 & 4.36 & 0.19 & 0.30 & 0.13 \\
 
Yi-VL-34B~\cite{young2024yi}                  & 8.50         & 15.35        & 3.33         & 8.97   & 16.30                  & 4.23   & 10.09                  & 19.75   & 4.94   & 0.22   & 2.60   & 0.13   \\
 
LLaVA-7B~\cite{liu2024visual}   & 31.29  & 67.50 & 8.00 & 31.28  & 67.25 & 11.22  & 31.49  & 92.15   & {\color[HTML]{434343} 12.37} & 35.32  & 62.35  & 17.37  \\
 
LLaVA-13B~\cite{liu2024visual}                  & 31.54        & 67.63        & 12.64        & 31.49                  & 73.25                  & 11.54                  & 34.97                  & 94.25   & 16.03                  & 43.13                  & 80.60                  & 23.91                  \\
 
LLaVA-34B~\cite{liu2024visual}                  & 39.95        & 85.75        & 18.85        & \textbf{52.75}             & \textbf{85.20}             & \textbf{33.91}             & \textbf{56.41}             & \textbf{95.81}            & \textbf{25.31}             & 55.05                  & \textbf{86.50}             & 18.97                  \\
\hline
 
Qwen VL~\cite{bai2023qwen}    & 2.73         & 6.60         & 1.03         & 6.25   & 16.00                  & 3.65   & 18.74                  & 71.50   & 5.45   & 8.73   & 16.05                  & 5.58   \\
 
Qwen VL-C~\cite{bai2023qwen}                  & 8.72         & 16.90        & 6.67         & 5.26   & 8.60   & 4.10   & 12.11                  & 47.75   & 8.08   & 25.99                  & 43.40                  & 13.21                  \\
 
CogVLM~\cite{wang2023cogvlm}     & 0.04         & 0.00         & 0.00         & 0.00   & 0.00   & 0.00   & 0.10   & 0.95    & 0.00   & 0.00   & 0.00   & 0.00   \\
 
CogVLM-G~\cite{wang2023cogvlm}   & 0.00         & 0.00         & 0.00         & 9.86   & 13.50                  & 6.79   & 22.64                  & 75.45   & 0.45   & 11.25                  & 22.65                  & 7.12   \\
 
CogVLM-C~\cite{wang2023cogvlm}   & 12.89        & 22.75        & 7.18         & 25.37                  & 43.63                  & 12.03                  & 28.25                  & 72.80   & 17.50                  & 30.16                  & 56.00                  & 16.35                  \\
\hline

LLaVA-7B~\cite{liu2024visual}      &- &- &-   & 9.16               & 16.40              & 5.51        &- &- &-                & 11.68              & 23.55              & 9.36                  \\
GLaMM~\cite{rasheed2024glamm}      &- &- &-   & 27.11        & 53.35        & 13.01        &- &- &-                 & \textbf{63.81}                  & 81.75                  & 53.40                  \\
GroundHOG~\cite{zhang2024groundhog}     &- &- &-   & 23.57              & 30.80              & 24.23        &- &- &-               & 44.80              & 43.10              & 38.97                  \\
\hline
IDEFICS~\cite{laurenccon2024obelics}    & 0.00         & 1.45         & 0.13         & 6.25   & 18.70                  & 0.64   & 17.37                  & 76.15   & 10.06                  & 4.62   & 0.00   & 0.32   \\
IDEFICS~\cite{laurenccon2024obelics}    & 0.00         & 1.45         & 0.13         & 6.25   & 18.70                  & 0.64   & 17.37                  & 76.15   & 10.06                  & 4.62   & 0.00   & 0.32   \\
CogVLM-2~\cite{wang2023cogvlm}   & 21.51        & 37.55        & 17.31        & 37.02                  & 70.85                  & 12.69                  & 37.10                  & 73.50   & 17.44                  & 21.16                  & 38.75                  & 13.65                  \\
MiniCPM-V~\cite{hu2024minicpm}    & 34.75        & 59.91        & 17.37        & 31.62                  & 62.80                  & 13.65                  & 32.16                  & 68.05   & 16.79                  & 27.42                  & 55.35                  & 16.92                  \\
GPT-4V~\cite{achiam2023gpt}     & 53.80        & 77.55        & 40.83       &- &- &-   &- &- &-                 & 55.89                  & 78.25                  & 41.03                  \\
 
GPT-4O~\cite{hurst2024gpt}      & \textbf{71.27} & \textbf{89.25} & \textbf{66.03} &- &- &-   &- &- &-               & 60.77             & 73.92                  & \textbf{54.31}             \\
\hline
LLaVA-7B~\cite{liu2024visual}      & 21.26 & 52.40 & 7.69 &- &- &-   &- &- &-               & 30.59             & 60.85                  & 12.69            \\
+OPERA~\cite{huang2024opera}      & 24.07 & 58.65 & 7.35 &- &- &-   &- &- &-               & 30.44             & 60.85                  & 13.27            \\
\cmidrule(r){1-13}

\multicolumn{13}{c}{\textit{Unseen}} \\    \hline
Yi-VL-6B~\cite{young2024yi} & 2.74         & 3.88         & 1.14         & 3.18   & 4.24   & 5.20   & 4.04   & 10.90   & 10.57                  & 0.14   & 0.45   & 0.08   \\
 
Yi-VL-34B~\cite{young2024yi}                  & 7.77         & 15.63        & 4.23         & 10.28                  & 18.04                  & 7.97   & 11.24                  & 22.49   & 12.03                  & 0.46   & 2.37   & 0.41   \\
 
LLaVA-7B~\cite{liu2024visual}   & 30.56        & 68.12        & 10.33        & 30.55                  & 68.16                  & 10.24                  & 31.89                  & 90.33   & {\color[HTML]{434343} 13.25} & 34.88                  & 64.41                  & 16.18                  \\
 
LLaVA-13B~\cite{liu2024visual}                  & 27.56        & 63.10        & 8.37         & 27.41                  & 63.10                  & 8.37   & 35.65                  & 91.09   & 14.80                  & 42.66                  & 71.92                  & 23.41                  \\
 
LLaVA-34B~\cite{liu2024visual}                  & 29.30        &79.43        & 17.72        & 29.45                  & \textbf{91.18}             & 14.39             & \textbf{37.40}             & \textbf{95.51}            & 17.92                  & 51.71                  & 77.88                  & 30.81                  \\
\hline
 
Qwen VL~\cite{bai2023qwen}    & 2.80         & 1.95         & 7.06         & 7.17   & 16.41                  & 4.15   & 10.34                  & 58.00   & 4.07   & 17.73                  & 31.22                  & 9.51   \\
 
Qwen VL-C~\cite{bai2023qwen}                  & 18.86        & 30.73        & 8.78         & 16.16                  & 27.80                  & 7.72   & 21.81                  & 58.00   & 11.14                  & 34.20                  & 57.31                  & 15.37                  \\
 
CogVLM~\cite{wang2023cogvlm}     & 0.03         & 0.00         & 0.00         & 0.00   & 0.00   & 0.00   & 0.00   & 0.15    & 0.00   & 0.00   & 0.00   & 0.00   \\
 
CogVLM-G~\cite{wang2023cogvlm}   & 0.00         & 0.00         & 0.00         & 8.20   & 1.47   & 5.77   & 23.82                  & 81.20   & 1.81   & 10.32                  & 10.74                  & 9.11   \\
 
CogVLM-C~\cite{wang2023cogvlm}   & 15.56        & 26.57        & 5.53         & 17.18                  & 41.27                  & 6.02   & 22.81                  & 56.04   & 6.67   & 30.56                  & 52.00                  & 13.50                  \\
\hline
LLaVA-7B~\cite{liu2024visual}      &- &- &-  & 7.59               & 12.12              & 4.88       &- &- &-                 & 12.71              & 22.49              & 8.46                  \\
GLaMM~\cite{rasheed2024glamm}      &- &- &-   & 29.11        & 54.53        & 14.23        &- &- &-                & \textbf{68.65}             & 77.06                  & 52.28                  \\
GroundHOG~\cite{zhang2024groundhog}      &- &- &-   & 23.11              & 24.69              & \textbf{26.26}       &- &- &-                & 40.73              & 30.37              & 38.13                  \\
\hline
IDEFICS~\cite{laurenccon2024obelics}    & 0.39         & 0.37         & 0.33         & 9.03   & 24.45                  & 2.68   & 24.80                  & 83.02   & 7.64   & 4.62   & 3.67   & 6.50   \\
CogVLM-2~\cite{wang2023cogvlm}   & 20.99        & 35.06        & 15.93        & 24.64                  & 38.04                  & 23.17                 & 26.74                  & 46.04   & \textbf{26.59}             & 11.13                  & 30.94                  & 5.77   \\
MiniCPM-V~\cite{hu2024minicpm}    & 32.96        & 59.92        & 16.60        & \textbf{31.77}             & 58.98                  & 14.15                  & 31.87                  & 60.98   & 16.34                  & 25.56                  & 47.76                  & 14.39                  \\
GPT-4V~\cite{achiam2023gpt}      & 45.46        & 63.12        & 34.17        &- &- &-   &- &- &-                 & 47.34                  & 64.94                  & 35.45                  \\
 
GPT-4O~\cite{hurst2024gpt}      & \textbf{63.27} & \textbf{80.29} & \textbf{54.47} &- &- &-   &- &- &-                & 63.45                  & \textbf{79.84}             & \textbf{53.74}           \\  
\hline
LLaVA-7B~\cite{liu2024visual}      & 13.96 & 31.88 & 3.98 &- &- &-  &- &- &-                & 26.95                  & 54.41             & 11.06         \\  
+OPERA~\cite{huang2024opera}      & 13.20 & 37.14 & 3.82 &- &- &-   &- &- &-                 & 27.90                  & 56.69             & 11.22         \\  
\hline
\end{tabular}
\end{threeparttable}
% \vspace*{2pt}
\label{rope}
\end{table*}

\subsubsection{CVQA: Culturally-diverse Multilingual Visual Question Answering Benchmark}
\label{appendix_cvqa}

Visual Question Answering (VQA) is a crucial task in MLLMs, designed to test their understanding and reasoning capabilities across visual and textual data~\cite{antol2015vqa, mathew2021docvqa,masry2022chartqa,kafle2018dvqa,chen2021geoqa}. However, most existing VQA datasets primarily focus on English and a few major world languages, with images often being Western-centric. While recent efforts have expanded the linguistic coverage of VQA datasets, they still lack diversity in low-resource languages. Moreover, these datasets typically extend their language range through translation or other methods while keeping the images unchanged, leading to limited cultural representation. To address these limitations, David Romero et al.~\cite{wang2023llm} developed a new benchmark, CVQA, which aims to encompass rich linguistic and cultural diversity. This benchmark involves native speakers and cultural experts in the data collection process to ensure authenticity and inclusivity.

\begin{figure*}[htbp]
    \centering
    \includegraphics[width=0.96\linewidth,height=0.48\linewidth]{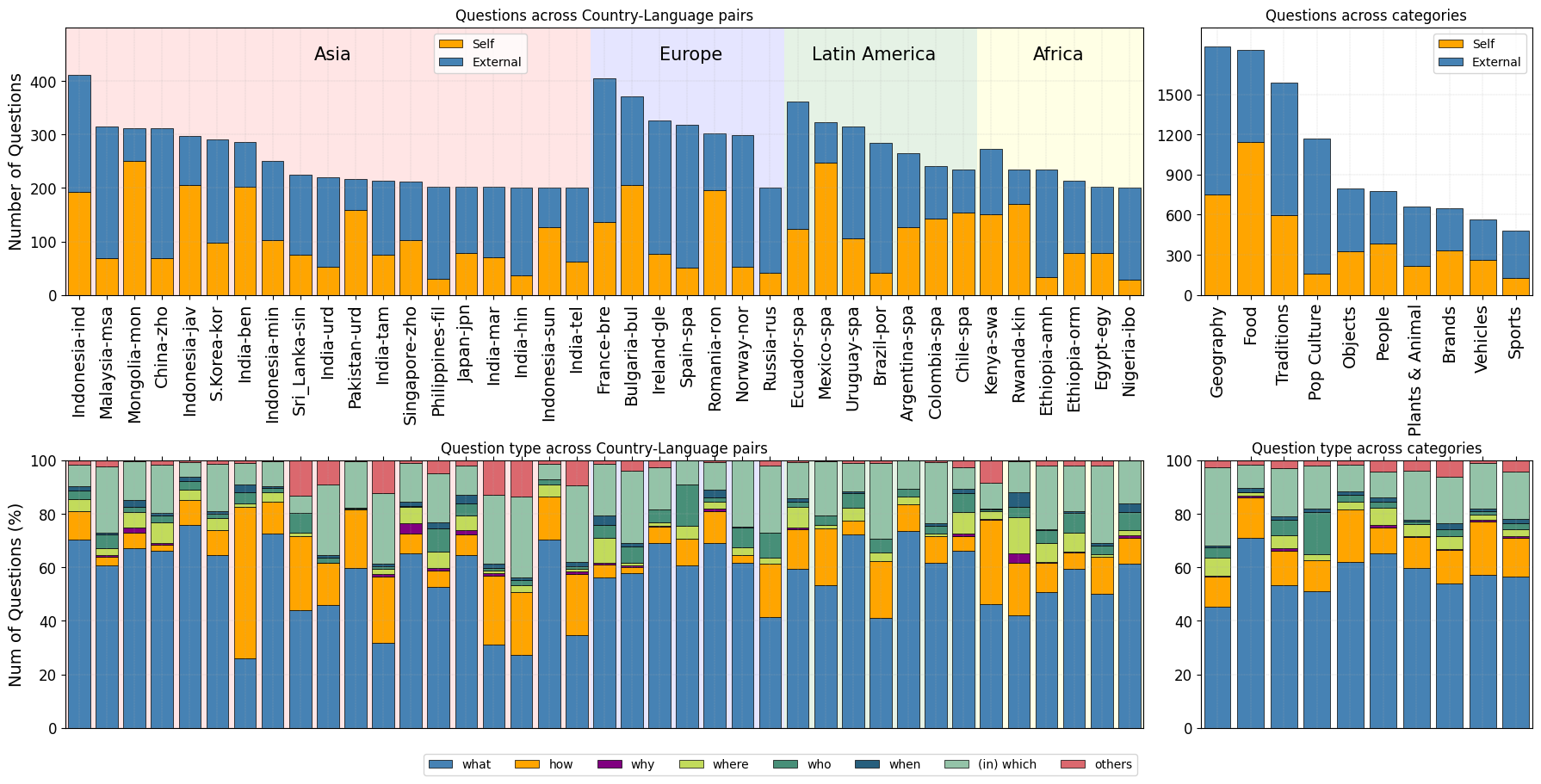}
    \caption{Statistics of the CVQA Benchmark.~\cite{mathew2021docvqa}}
    \label{cvqa_fig1}
\end{figure*}

Figure~\ref{cvqa_fig1} illustrates the scale and diversity of the CVQA benchmark, which includes 10,374 questions and languages from 30 different countries. This demonstrates how it covers a wide range of languages and cultures.

\begin{figure*}[t]
\centering  %图片全局居中
\subfigure[image 1]{
\label{cvqa_fig2}
\includegraphics[width=0.4\linewidth,height = 0.4\linewidth]{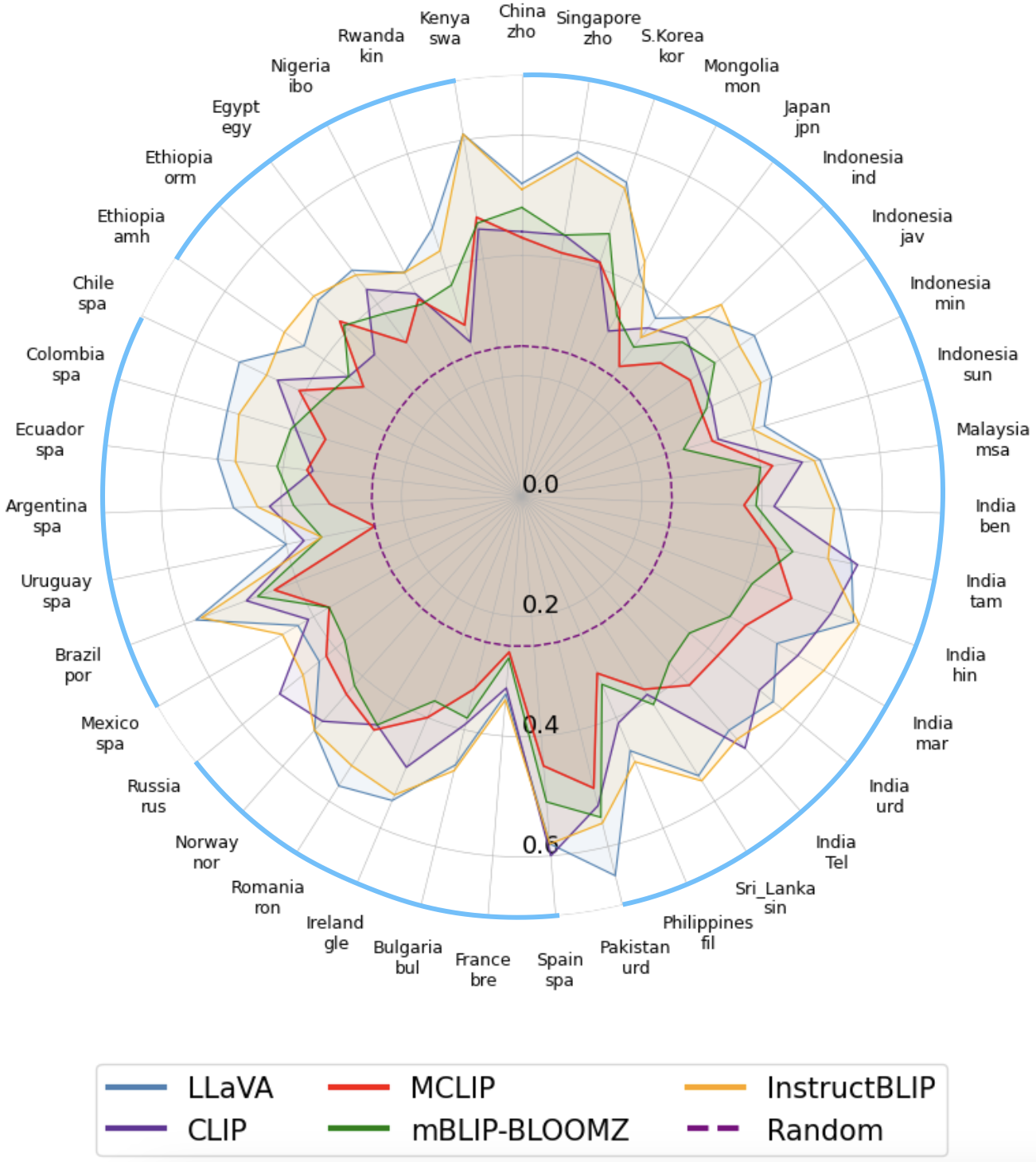}}
\subfigure[image 2]{
\label{cvqa_fid3}
\includegraphics[width=0.4\linewidth,height = 0.4\linewidth]{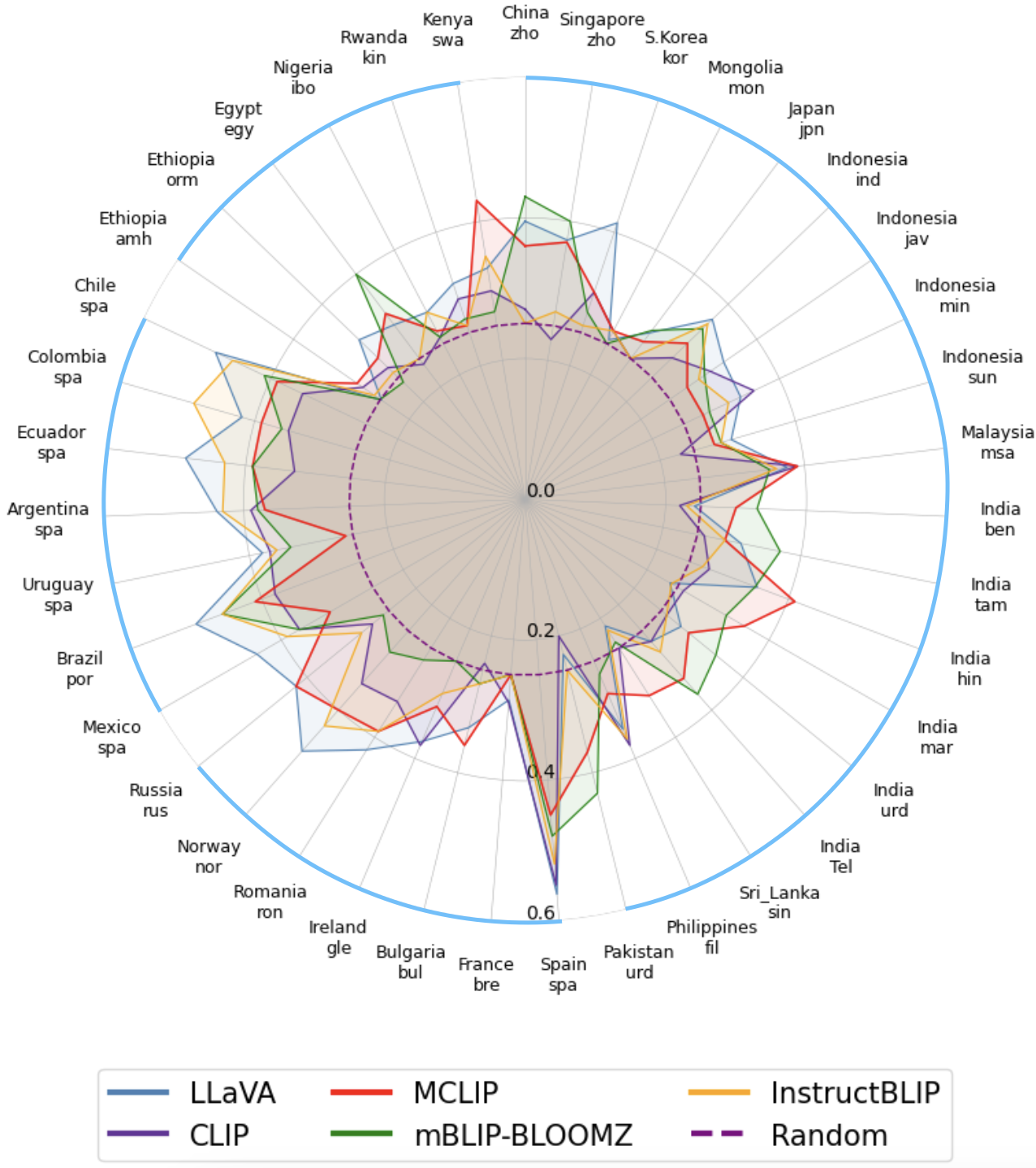}}
\caption{Model performance per Country-Language pair. The blue lines indicate separation by continent. All models show similar behaviour in the majority of cases, despite having different sizes.~\cite{mathew2021docvqa}}
\label{cvqa_fig23}
\end{figure*}

Figure~\ref{cvqa_fig23} shows the performance of different models across various country-language pairs, including question-option pairs in both English and local languages. The blue line in the figure represents performance separated by continents. Despite differences in scale, it highlights the similar behavior of all models in most cases. This figure reveals the challenges models face when handling questions in local languages, as well as the performance variations across different regions and languages.

% 插入两张并排的图片

Table~\ref{cvqa_tab1} shows the average performance of different MLLMs on the CVQA dataset using English prompts (EN) and local language prompts (LOC). These results indicate that even the best-performing open models, such as LLaVA-1.5-7B, significantly lag behind closed models on CVQA. Furthermore, their performance is poorer with local language prompts, highlighting the challenges models face when processing non-English prompts.

\begin{table*}[htbp]
 \setlength{\tabcolsep}{0pt}
 \renewcommand\arraystretch{1.4}
\centering
\belowrulesep=0pt
\aboverulesep=0pt
\caption{Average performance of MLLMs on our CVQA dataset with English prompts~(EN) and local language prompts~(LOC).~\cite{mathew2021docvqa}}
% \vspace{0.05in}
\resizebox{\textwidth}{!}{
\begin{tabular}{cc|cc|cc|cc|cc|cc|cc|ccccc}
\hline
\multicolumn{2}{c|}{\textbf{LLaVA-1.5-7B}~\cite{liu2024visual}} &\multicolumn{2}{c|}{\textbf{M-CLIP}~\cite{chen2023mclip}} & \multicolumn{2}{c|}{\textbf{CLIP}~\cite{radford2021learning}} & 
\multicolumn{2}{c|}{\textbf{mBLIP-mT0}~\cite{geigle2023mblip}} & \multicolumn{2}{c|}{\textbf{mBLIP-BLOOMZ}~\cite{geigle2023mblip}} & \multicolumn{2}{c|}{\textbf{InstructBLIP}~\cite{dai2023instructblipgeneralpurposevisionlanguagemodels}} &  \multicolumn{2}{c|}{\textbf{Gemini-1.5-Flash}~\cite{team2024gemini}} & \multicolumn{2}{c}{\textbf{GPT-4o}~\cite{hurst2024gpt}} \\
\cmidrule(r){0-2}\cmidrule(r){2-4}\cmidrule(r){4-6}\cmidrule(r){6-8}\cmidrule(r){8-10}\cmidrule(r){10-12}\cmidrule(r){12-14}\cmidrule(r){14-16}
\textbf{EN} & \textbf{LOC} &
\textbf{EN} & \textbf{LOC} &
 \textbf{EN} & \textbf{LOC} & \textbf{EN} & \textbf{LOC} & \textbf{EN} & \textbf{LOC} & \textbf{EN} & \textbf{LOC} & \textbf{EN} & \textbf{LOC} & \textbf{EN} & \textbf{LOC} \\
\hline
49.6 & 35.5 & 38.0 & 33.7 & 42.7 & 30.6 & 31.3 & 30.9 & 39.3 & 32.7 & 49.0 & 31.9 & 66.9 & 68.5 & 75.4 & 74.3 \\
\hline
\end{tabular}
}
\label{cvqa_tab1}
\end{table*}

Table~\ref{cvqa_tab2} compares the performance of LLaVA-1.5-7B and InstructBLIP on CVQA and other established English VQA benchmarks. The results show that while LLaVA-1.5-7B performs better on other English VQA benchmarks, it still faces challenges on CVQA, highlighting the difficulty of culturally specific questions in CVQA.

\begin{table*}[htbp]
\small
\centering
\belowrulesep=0pt
\aboverulesep=0pt
\caption{LLaVA-1.5-7B and InstructBLIP results on various VQA datasets.~\cite{mathew2021docvqa}}
\resizebox{\textwidth}{!}{
\begin{tabular}{l|c|c|c|c|c|c|cccc}
\hline
\textbf{Model} & \textbf{VQAv2} & \textbf{GQA} & \textbf{VizWiz} & \textbf{SciQA-IMG} & \textbf{TextVQA} & \textbf{CVQA~(EN)} & \textbf{CVQA~(LOC)}\\
\hline
LLaVA-1.5-7B~\cite{liu2024visual}& 78.5 & 62.0 & 50.0 & 66.8 & 58.2 & 48.9 & 36.5\\
InstructBLIP~\cite{dai2023instructblipgeneralpurposevisionlanguagemodels} & - & 49.2 & 34.5 & 60.5 & 50.1& 47.8 & 32.7\\
\hline
\end{tabular}}
\label{cvqa_tab2}
\end{table*}

Table~\ref{cvqa_tab3} presents the performance of models across 10 categories in CVQA. It shows that models achieve the highest accuracy in the "People" category, while the accuracy in the "Food" and "Pop Culture" categories is lower with local language prompts. This indicates that the diversity of food and pop culture across different cultures presents a challenge for the generalization of MLLMs.

\begin{table*}[htbp]
\centering
\caption{Accuracy of models across categories.~\cite{mathew2021docvqa}}
\resizebox{1\textwidth}{!}{
\begin{tabular}{@{}l|cc|cc|cc|cc|cc|cc@{}}
\hline
\multirow{2}{*}{\textbf{Categories}} & \multicolumn{2}{c|}{\textbf{LLaVA-1.5-7B}~\cite{liu2024visual}} & \multicolumn{2}{c|}{\textbf{M-CLIP}~\cite{chen2023mclip}} & \multicolumn{2}{c|}{\textbf{CLIP}~\cite{radford2021learning}} & \multicolumn{2}{c|}{\textbf{mBLIP-mT0}~\cite{geigle2023mblip}} & \multicolumn{2}{c|}{\textbf{mBLIP-BLOOMZ}~\cite{geigle2023mblip}} & \multicolumn{2}{c}{\textbf{InstructBLIP}~\cite{dai2023instructblipgeneralpurposevisionlanguagemodels}} \\
\cmidrule(r){2-3}\cmidrule(r){3-5}\cmidrule(r){5-7}\cmidrule(r){7-9}\cmidrule(r){9-11}\cmidrule(r){11-13}
 & \textbf{EN} & \textbf{LOC} & \textbf{EN} & \textbf{LOC} & \textbf{EN} & \textbf{LOC} & \textbf{EN} & \textbf{LOC} & \textbf{EN} & \textbf{LOC} & \textbf{EN} & \textbf{LOC} \\
\hline
Brands & \textbf{49.9} & 36.5 & 37.2& 35.7& 36.6& 29.7& 33.7 & 30.8 & 40.5 & 35.1 & 48.4 & 32.6 \\
Food & \textbf{45.4} & 31.9 & 34.5& 29.1& 39.2& 30.4& 28.1 & 27.6 & 37.7 & 29.8 & 44.4 & 30.6 \\
Geography & \textbf{47.1} & 38.2 & 37.1& 34.2& 41.8& 31.9& 30.6 & 31.6 & 35.0 & 32.3 & 45.3 & 33.2 \\
Objects & 51.8& 33.0& 39.4& 34.5 & 39.7& 25.4& 34.3 & 33.0 & 43.1 & 34.0 & \textbf{52.3}& 29.1 \\
People & 58.9& 38.1 & 45.0& 37.8& 46.8& 30.9& 35.3 & 34.7 & 46.3 & 36.7 & \textbf{59.8}& 34.0 \\
Plants \& Animals & \textbf{55.7}& 37.5 & 43.7& 32.0& 48.0& 27.2& 35.2 & 35.5 & 46.0 & 36.0 & 55.4 & 35.1 \\
Pop Culture & 44.5& 36.3 & 33.7& 31.5& \textbf{46.1}& 36.3& 28.8 & 29.9 & 35.7 & 30.7 & 45.1 & 34.6 \\
Sports & \textbf{50.7} & 39.1& 39.3& 33.3& 43.5& 32.4& 32.6 & 31.4 & 40.1 & 34.9 & 50.5 & 34.7 \\
Tradition & \textbf{50.4} & 35.8 & 37.0& 35.2& 41.9& 32.2& 31.6 & 31.5 & 39.0 & 32.2 & 47.9 & 30.8 \\
Vehicles & 50.6& 41.4& 39.5& 41.1& 44.6& 30.5& 35.6 & 33.9 & 42.0 & 34.0 & \textbf{55.0}& 33.0 \\
\hline
\end{tabular}
}
\label{cvqa_tab3}
\end{table*}

\subsubsection{II-Bench: Image Implication Understanding Benchmark}

\label{appendix_II-Bench}

Images often contain rich emotional and cultural narratives, and understanding their meaning and exploring the human emotions and cultural context they reflect requires attention to detail~\cite{bubeck2023sparks,achiam2023gpt,wachowiak2023does}. While MLLMs have made significant progress in understanding and generating cross-modal content, achieving new breakthroughs in benchmarks like image captioning~\cite{lin2014microsoft,sharma2018conceptual,sidorov2020textcaps,gurari2020captioning,pont2020connecting,agrawal2019nocaps} and visual question answering~\cite{antol2015vqa, mathew2021docvqa, masry2022chartqa,kafle2018dvqa,chen2021geoqa}, there has been insufficient exploration of their higher-order perceptual abilities. Ziqiang Liu et al.~\cite{liu2024ii} introduced a new benchmark, II-Bench, designed to evaluate MLLMs' ability to understand and reason about the complex implicit meanings in images, addressing the gap in existing benchmarks for assessing higher-order perceptual abilities in MLLMs.

II-Bench includes 1,222 images across six different domains: life, art, society, psychology, environment, and others. The images consist of various types, including illustrations, memes, posters, comics, logos, and paintings. Each image is accompanied by one to three multiple-choice questions, totaling 1,434 questions. Of these, 1,399 questions are used to construct the test set, and 35 questions are used for the development and validation sets.

Table~\ref{iibench_tab} presents the overall results of different MLLMs and human participants on the II-Bench benchmark. It shows model performance across various domains, such as life, art, society, psychology, and environment, as well as across different emotional categories (positive, neutral, and negative). The table lists the average and best accuracies for multiple open-source and closed-source MLLMs, alongside the performance of human participants.

\begin{table*}[!thp]
\renewcommand\arraystretch{1.2}
\centering
\caption{Overall results of different MLLMs and humans on different domains and emotions.~\cite{liu2024ii}}
\begin{tabular}{l|c|cccccc|ccc}
\hline
\textbf{Models} & \textbf{Overall} & \textbf{Life} & 
\textbf{Art} & \textbf{Society} & \textbf{Psy.} & \textbf{Env.} & \textbf{Others} & \textbf{Positive} & \textbf{Neutral} & \textbf{Negative} \\
 {} & (1,399) & (585) & (85) & (461) & (152) & (51) & (65) & (196) & (789) & (414) \\
\hline
\multicolumn{11}{c}{\textit{Open-source Models}} \\ 
\hline
InstructBLIP-T5-XL~\cite{dai2023instructblipgeneralpurposevisionlanguagemodels} & 47.3 & 45.6 & 48.2 & 48.8 & 44.7 & 52.9 & 50.8 & 46.9 & 48.3 & 45.4 \\
BLIP-2 FLAN-T5-XL~\cite{li2023blip2} & 52.8 & 53.0 & 58.8 & 52.5 & 42.8 & 64.7 & 58.5 & 56.1 & 52.9 & 51.0 \\
mPLUGw-OWL2~\cite{ye2023mplugowl2} & 53.2 & 54.0 & 56.5 & 50.5 & 52.0 & 60.8 & 56.9 & 55.6 & 52.6 & 53.1 \\
Qwen-VL-Chat~\cite{bai2023qwen}  & 53.4 & 53.2 & 49.4 & 52.1 & 50.0 & 60.8 & 72.3 & 56.1 & 52.6 & 53.6 \\
InstructBLIP-T5-XXL~\cite{dai2023instructblipgeneralpurposevisionlanguagemodels} & 56.7 & 56.2 & 58.8 & 58.6 & 45.4 & 64.7 & 64.6 & 63.3 & 56.1 & 54.6 \\
Mantis-8B-siglip-Llama3 & 57.5 & 56.8 & 61.2 & 57.5 & 53.9 & 64.7 & 61.5 & 59.2 & 58.0 & 55.6 \\
BLIP-2 FLAN-T5-XXL~\cite{li2023blip2} & 57.8 & 57.1 & 63.5 & 57.0 & 53.3 & 66.7 & 66.2 & 67.9 & 57.2 & 54.3 \\
DeepSeek-VL-Chat-7B~\cite{lu2024deepseekvl} & 60.3 & 59.0 & 58.8 & 58.4 & 61.8 & 68.6 & 76.9 & 65.8 & 60.1 & 58.0 \\
Yi-VL-6B-Chat~\cite{young2024yi} & 61.3 & 60.9 & 63.5 & 60.7 & 56.6 & 66.7 & 72.3 & 61.7 & 61.7 & 60.1 \\
InternLM-XComposer2-VL~\cite{dong2024internlmxcomposer2} & 62.1 & 61.7 & 62.4 & 62.3 & 58.6 & 70.6 & 66.2 & 65.8 & 63.0 & 58.7 \\
InternVL-Chat-1.5~\cite{chen2024far} & 66.3 & 63.6 & 65.9 & 68.5 & 65.8 & 64.7 & 76.9 & 73.5 & 65.4 & 64.5 \\
Idefics2-8B~\cite{laurenccon2024obelics} & 67.7 & 67.2 & \textbf{74.1} & 67.7 & 62.5 & 74.5 & 70.8 & 68.9 & 67.0 & 68.4 \\
Yi-VL-34B-Chat~\cite{young2024yi} & 67.9 & 67.5 & 70.6 & 67.7 & 63.8 & 70.6 & 76.9 & 74.0 & 68.2 & 64.5 \\
MiniCPM-Llama3-2.5~\cite{hu2024minicpm} & 69.4 & 68.4 & 71.8 & 69.4 & 64.5 & \textbf{80.4} & 78.5 & 75.0 & 69.3 & 66.9\\
CogVLM2-Llama3-Chat~\cite{hong2024cogvlm2} & 70.3 & 68.9 & 68.2 & 70.9 & 67.8 & 72.5 & \textbf{86.2} & 69.9 & 71.1 & 69.1 \\
LLaVA-1.6-34B~\cite{liu2024visual} & \textbf{73.8} & \textbf{73.8} & 71.8 & \textbf{73.3} & \textbf{71.1} & 78.4 & 81.5 & \textbf{79.1} & \textbf{72.9} & \textbf{72.9} \\
\hline
\multicolumn{11}{c}{\textit{Closed-source Models}} \\ 
\hline
GPT-4V~\cite{achiam2023gpt} & 65.9 & 65.0 & 69.4 & 65.3 & 59.9 & 76.5 & 80.0 & 69.4 & 66.0 & 64.0 \\
GPT-4o~\cite{hurst2024gpt} & 72.6 & 72.5 & 72.9 & 73.3 & 68.4 & 76.5 & 75.4 & 78.6 & 71.2 & 72.5 \\
Gemini-1.5 Pro~\cite{geminiteam2024} & 73.9 & 73.7 & \textbf{74.1} & 74.4 & 63.2 & \textbf{80.4} & 83.1 & \textbf{80.1} & 70.8 & \textbf{75.4} \\
Qwen-VL-MAX~\cite{bai2023qwen} & \textbf{74.8} & \textbf{74.7} & 71.8 & \textbf{74.6} & \textbf{73.0} & 76.5 & \textbf{84.6} & \textbf{80.1} & \textbf{74.5} & 72.9 \\ 
\hline
\multicolumn{11}{c}{\textit{Humans}} \\ 
\hline
Human\_avg~\cite{liu2024ii} & 90.3 & 90.0 & 88.2 & 91.4 & 86.6 & 96.1 & 92.3 & 84.7 & 89.1 & 92.2  \\ 
Human\_best~\cite{liu2024ii} & \textbf{98.2} & \textbf{97.9} & \textbf{98.8} & \textbf{98.3} & \textbf{97.4} & \textbf{100.0} & \textbf{100.0} & \textbf{98.0} & \textbf{98.0} & \textbf{98.8} \\ 
\hline
\end{tabular}%
\label{iibench_tab}
\end{table*}

\subsubsection{ConBench: MLLMs Answer Consistency Evaluation Benchmark}

\label{appendix_ConBench}

MLLMs have made rapid progress in visual information perception and reasoning. Although MLLMs are capable of generating high-quality task prompt responses, simply modifying the prompt can lead to contradictory answers, even when the correct answer is provided. Specifically, under different prompt space sizes, these models lack consistency in answers to the same knowledge point, which significantly undermines trust in these models~\cite{li2023benchmarking,lin2023generating}. To ensure that MLLMs can predict correct and consistent answers when faced with various query formats, Yuan Zhang et al.~\cite{zhang2024unveiling} proposed a multimodal benchmark tool, ConBench, designed to comprehensively assess the consistency of MLLMs—specifically, their ability to provide the same answer to the same knowledge point across different query formats.

ConBench evaluates MLLMs by offering a diverse set of question formats, including true/false questions, multiple-choice questions, and limited visual question answering (VQA) problems. It also introduces two multidimensional evaluation metrics: 1)Discriminative Domain Evaluation Metric (ConScore[D]): Assesses consistency based on the accuracy of the model's answers to discriminative questions. 2)Generative Domain Evaluation Metric (ConScore[C]): Evaluates consistency by comparing the coherence between the model-generated captions and the discriminative answers.

\begin{figure}[htbp]
    \centering
    \includegraphics[width=0.9\linewidth,height=0.65\linewidth]{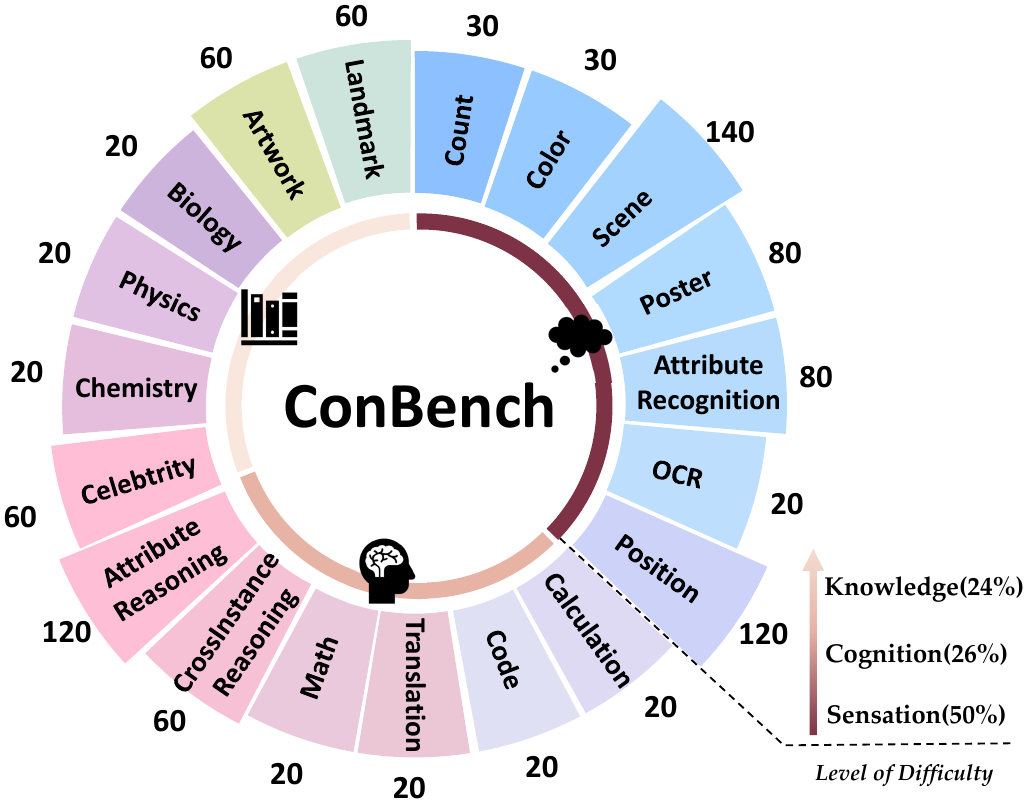}
    \caption{Overview of 19 evaluation detailed categories in ConBench.~\cite{zhang2024unveiling}}
    \label{conbench_fig}
\end{figure}

The specific structure of ConBench is shown in figure ~\ref{conbench_fig}, providing an overview of the 19 evaluation categories in ConBench. These categories are distributed across three core capabilities: Sensation, Cognition, and Knowledge. The benchmark comprehensively covers tasks of varying difficulty levels, thereby assessing the performance of MLLMs across different aspects.

Table~\ref{conbench_tab1} presents the performance evaluation results of different MLLMs on ConBench. These results are based on ConScore[D], which evaluates the correctness of the model's answers to discriminative questions. The table includes three types of questions: True/False (T), Multiple-Choice (C), and Limited Visual Question Answering (VQA) (V). It also shows the models' performance across the three core capabilities: Sensation, Cognition, and Knowledge.

\begin{table*}[t]
  \renewcommand\arraystretch{1.2}
    \centering
	\caption{\textbf{Evaluation[D] of mainstreams series of MLLMs on ConBench.} The detailed results of the Sensation, Cognition, and Knowledge core capabilities are listed below. T, C, and V represent true-false, multiple-choice, and limited VQA questions, respectively. The ranking can be found below the respective numbers. $\dagger$: \scriptsize{Due to safety considerations, GPT-4V declined to answer the celebrity category.}~\cite{zhang2024unveiling}}
	\label{conbench_tab1}
	\scriptsize
        \begin{center}
        \scalebox{1.05}{
	\begin{tabular}{l|c|cccc|cccc|cccc}
	    \shline
	    \multirow{2}{*}{\footnotesize{Method}} & \multirow{2}{*}{\footnotesize{ConScore[D]}} & \multicolumn{4}{c|}{\footnotesize{Sensation}} & \multicolumn{4}{c|}{\footnotesize{Cognition}} & \multicolumn{4}{c}{\footnotesize{Knowledge}}  \\
        \cmidrule(r){3-14}
          & & \multicolumn{1}{c}{\footnotesize{T}} & \multicolumn{1}{c}{\footnotesize{C}} & \multicolumn{1}{c}{\footnotesize{V}} & Con & \multicolumn{1}{c}{\footnotesize{T}} & \multicolumn{1}{c}{\footnotesize{C}} & \multicolumn{1}{c}{\footnotesize{V}} & Con & \multicolumn{1}{c}{\footnotesize{T}} & \multicolumn{1}{c}{\footnotesize{C}} & \multicolumn{1}{c}{\footnotesize{V}} & Con \\
	    \shline
	    \multicolumn{14}{c}{\textit{Closed-source Vision Language Models}}\\
        \hline
            GPT-4V$^\dagger$~\cite{achiam2023gpt}   & $29.20$ & $80.4$ & $79.0$ & $61.7$ & $48.3$ & $68.8$ & $53.2$ & $39.9$ & $20.4$ & $63.1$ & $57.2$ & $30.0$ & $14.2$ \\
            GPT-4-Omni~\cite{hurst2024gpt}  & $35.70$ & $89.2$ & $79.4$ & $64.4$ & $55.0$ & $71.8$ & $62.8$ & $44.9$ & $27.8$ & $64.7$ & $61.7$ & $39.7$ & $23.3$ \\
    	Gemini-Pro-Vision~\cite{geminiteam2024geminifamilyhighlycapable}  & $25.00$ & $85.2$ & $60.7$ & $63.4$ & $39.3$ & $71.8$ & $45.0$ & $44.2$ & $15.1$ & $65.0$ & $51.4$ & $39.7$ & $15.8$ \\
            Gemini-Ultra-Vision~\cite{geminiteam2024geminifamilyhighlycapable}  & $33.10$ & $78.9$ & $78.6$ & $66.3$ & $50.3$ & $68.1$ & $58.5$ & $47.9$ & $28.5$ & $62.9$ & $62.2$ & $44.7$ & $19.7$ \\
            Qwen-VL-Plus~\cite{bai2023qwen}  & $28.10$ & $82.7$ & $74.9$ & $60.4$ & $45.0$ & $64.2$ & $41.7$ & $30.8$ & $16.3$ & $63.6$ & $54.2$ & $33.3$ & $15.8$ \\
            Qwen-VL-Max~\cite{bai2023qwen} & $\mathbf{37.00}$ & $86.4$ & $80.7$ & $65.4$ & $\mathbf{56.3}$ & $72.9$ & $51.4$ & $51.3$ & $28.1$ & $68.3$ & $58.6$ & $38.9$ & $\mathbf{24.2}$ \\
	    \hline
	    \multicolumn{14}{c}{\textit{7B Vision Language Models}}\\
        \hline
            LLaVA-v1.5-7B~\cite{liu2024visual}  & $16.60$ & $79.3$ & $56.8$ & $44.3$ & $28.3$ & $51.4$ & $33.5$ & $15.8$ & $4.7$ & $61.7$ & $44.4$ & $16.9$ & $7.8$ \\
            Qwen-VL-Chat~\cite{bai2023qwen}  & $26.40$ & $81.0$ & $79.6$ & $54.2$ & $39.0$ & $55.0$ & $46.3$ & $33.2$ & $13.5$ & $60.3$ & $54.2$ & $28.9$ & $14.7$ \\
	    \hline
	    \multicolumn{14}{c}{$\sim$ \textit{13B Vision Language Models}}\\
        \hline
            LLaVA-v1.5-13B~\cite{liu2024visual}  & $24.00$ & $82.9$ & $77.1$ & $49.6$ & $39.5$ & $53.6$ & $37.8$ & $20.1$ & $10.4$ & $65.6$ & $50.3$ & $17.2$ & $9.7$ \\
            MiniGemini-13B~\cite{li2024minigemini}  & $21.80$ & $81.9$ & $69.7$ & $52.8$ & $39.3$ & $51.9$ & $38.2$ & $21.1$ & $6.9$ & $52.8$ & $36.7$ & $17.5$ & $9.2$ \\
            InternVL-v1.5-26B~\cite{chen2024far} & $31.40$ & $85.6$ & $84.8$ & $65.0$ & $54.3$ & $59.7$ & $58.6$ & $44.4$ & $19.4$ & $58.1$ & $55.8$ & $25.3$ & $12.2$ \\
	    \hline
	    \multicolumn{14}{c}{$\sim$ \textit{34B Vision Language Models}}\\
        \hline
            LLaVA-NeXT-34B~\cite{li2024llavanextinter} & $27.70$ & $82.4$ & $81.7$ & $55.6$ & $43.6$ & $50.7$ & $47.5$ & $25.6$ & $9.9$ & $60.4$ & $56.1$ & $27.8$ & $12.8$ \\
            MiniGemini-34B~\cite{li2024minigemini} & $23.00$ & $80.8$ & $76.8$ & $48.2$ & $39.7$ & $36.9$ & $30.7$ & $18.9$ & $6.0$ & $58.1$ & $42.3$ & $20.8$ & $8.2$ \\
            InternVL-v1.2P-40B~\cite{chen2024internvl} & $34.70$ & $83.7$ & $83.2$ & $66.6$ & $53.4$ & $74.2$ & $67.6$ & $57.1$ & $\mathbf{34.9}$ & $72.2$ & $58.3$ & $28.6$ & $13.6$ \\

	    \shline
	\end{tabular}
    }
         \end{center}
\end{table*}

Table~\ref{conbench_tab2} further evaluates the consistency between the captions generated by MLLMs and the discriminative answers (ConScore[C]). This includes the overall ConScore[C], as well as consistency scores for the three question types: True/False (T), Multiple-Choice (C), and Limited Visual Question Answering (VQA) (V).

\newcommand{\fg}[1]{\mathbf{\mathcolor{ForestGreen}{#1}}}
\newcommand{\fr}[1]{\mathbf{\mathcolor{Forestred}{#1}}}
\begin{table*}[t]
\renewcommand\arraystretch{1.2}
    \centering
	\caption{\textbf{Evaluation of Consistency between caption and three discriminative types of answer on ConBench.} The Con[$X$] is the Consistency ratio between discriminative answer type $X$ and caption. The "ordered" represents whether Con[T] $<$ Con[C] $<$ Con[V] is in its line.~\cite{zhang2024unveiling}}
	\label{conbench_tab2}
	% \footnotesize
	\scriptsize
	%\normalsize
    \begin{center}
    \scalebox{1.2}{
	\begin{tabular}{l|c|c|cccc}
	    \shline
        \footnotesize{Method} & \footnotesize{ConScore[C]} & \footnotesize{Con[T]} & \footnotesize{Con[C]} & \footnotesize{Con[V]} & \footnotesize{Ordered}\\
	    \shline
	    \multicolumn{6}{c}{\textit{Closed-source Vision Language Models}}\\ \hline
            GPT-4V~\cite{achiam2023gpt}  & 55.6 & $51.20$ & $56.50$ & $59.10$ & Y  \\
            GPT-4-Omni~\cite{hurst2024gpt}  & $\mathbf{62.2}$ & $58.00$ & $62.50$ & $66.10$ & Y \\
            Gemini-Pro-Vision~\cite{geminiteam2024geminifamilyhighlycapable}  & $48.4$ & $43.30$ & $45.20$ & $56.80$ & Y \\
            Gemini-Ultra-Vision~\cite{geminiteam2024geminifamilyhighlycapable} & $54.6$ & $47.80$ & $55.20$ & $60.70$ & Y \\
            Qwen-VL-Plus~\cite{bai2023qwen}  & $50.2$ & $47.10$ & $49.10$ & $54.30$ & Y \\
            Qwen-VL-Max~\cite{bai2023qwen}  & $ 58.4$ & $54.30$ & $58.00$ & $62.90$ & Y \\
	    \hline
    
	    \multicolumn{6}{c}{\textit{7B Vision Language Models}}\\     \hline
            LLaVA-v1.5-7B~\cite{liu2024visual}  & $38.4$ & $39.20$ & $36.60$ & $39.50$ & N\\
            Qwen-VL-Chat~\cite{bai2023qwen}    & $48.0$ & $42.00$ & $50.80$ & $51.30$ & Y\\
	    \hline
	    \multicolumn{6}{c}{$\sim$ \textit{13B Vision Language Models}}\\     \hline
            LLaVA-v1.5-13B~\cite{liu2024visual}   & $44.4$ & $41.50$ & $45.80$& $46.00$ & Y\\
            MiniGemini-13B~\cite{li2024minigemini}  & $41.7$ & $38.80$ & $42.90$ & $43.30$ & Y\\
            InternVL-v1.5-26B~\cite{chen2024far} & $50.9$ & $44.50$ & $53.90$& $54.20$ & Y\\
	    \hline
	    \multicolumn{6}{c}{$\sim$ \textit{34B Vision Language Models}}\\    \hline
            LLaVA-NeXT-34B & $48.3$ & $46.00$ & $52.20$ & $46.80$ & N\\
            MiniGemini-34B~\cite{li2024minigemini}  & $49.6$ & $56.80$ & $48.00$ & $44.10$ & N\\
            InternVL-v1.2P-40B~\cite{chen2024internvl} & $53.7$ & $49.80$ & $55.50$ & $55.80$ & Y\\

	    \shline
	\end{tabular}
    }
         \end{center}
\end{table*}

\subsubsection{COMPBENCH: Comparative Reasoning Benchmark}

\label{appendix_COMPBENCH}

The ability to compare objects, scenes, or situations is crucial for decision-making and problem-solving in everyday life~\cite{masry2022chartqa,hudson2019gqa,lu2022learn}. Although this ability is widespread in human cognition, it has not been fully explored in the field of Artificial General Intelligence (AGI). Jihyung Kil et al.~\cite{kil2024compbench} proposed a benchmark, COMPBENCH, designed to evaluate the comparative reasoning ability of MLLMs.

As show in table~\ref{COMPBENCH_tab1}. COMPBENCH questions are carefully crafted to distinguish relative features between two images, testing the models' performance across eight different comparative dimensions by providing image pairs and related questions. Table~\ref{COMPBENCH_tab2} presents the performance of recent MLLMs on the COMPBENCH benchmark.

\begin{table}[t]
\renewcommand\arraystretch{1.2}
\caption{Overall statistics of COMPBENCH.~\cite{kil2024compbench}}
\setlength{\tabcolsep}{3mm}{
\begin{tabular}{cccc}
\toprule
\multirow{2}{*}{Relativity} & \multirow{2}{*}{Dataset} &\multirow{2}{*}{Domain}  &\multirow{2}{*}{samples}\\ \\
\midrule
\multirow{5}{*}{Attribute} & MIT-States & Open & 0.2K \\
& Fashionpedia & Fashion & 2.4K \\
& VAW & Open & 0.9K \\
& CUB-200-2011 & Bird & 0.9K \\
& Wildfish++ & Fish & 0.9K \\
\cmidrule{1-4}
\multirow{2}{*}{Existence} & MagicBrush & Open & 0.9K \\
& Spot-the-diff & Outdoor Scene & 1.2K \\
\cmidrule{1-4}
\multirow{2}{*}{State} & MIT-States & Open & 0.6K \\
& VAW & Open & 0.5K \\
\cmidrule{1-4}
\multirow{2}{*}{Emotion} & CelebA & Face & 1.5K \\
& FER-2013 & Face & 3.8K \\
\cmidrule{1-4}
\multirow{2}{*}{Temporality} & SoccerNet & Sport & 8.3K \\
& CompCars & Car & 5K \\
\cmidrule{1-4}
Spatiality & NYU-Depth V2 & Indoor Scene & 1.9K \\
\cmidrule{1-4}
Quantity & VQAv2 & Open & 9.8K \\
\cmidrule{1-4}
Quality & Q-Bench2 & Open & 1K \\
\cmidrule{1-4}
Total & - & - & 39.8K \\
\bottomrule
\end{tabular}}
\label{COMPBENCH_tab1}
\end{table}

\begin{table*}[t]
\renewcommand\arraystretch{1.2}
\centering
\caption{Overall results on COMPBENCH test split. Evaluating four leading MLLMs across eight relative comparisons spanning sixteen tasks.~\cite{kil2024compbench}
}
\scalebox{0.9}{
\begin{tabular}{l|ccccc|cc|cc|cc|cc|c|c|c|c}
\hline
\multirow{2}{*}{Model} &
\multicolumn{5}{c|}{Attribute} &
\multicolumn{2}{c|}{Exist.} &
\multicolumn{2}{c|}{State} &
\multicolumn{2}{c|}{Emot.} &
\multicolumn{2}{c|}{Temp.} &
\multicolumn{1}{c|}{Spat.} &
Quan. &
Qual. & 
\multirow{2}{*}{Avg} \\

 \cmidrule(r){2-17}
%\cmidrule(r){2-6} \cmidrule(r){7-8} \cmidrule(r){9-10} \cmidrule(r){11-12} \cmidrule(r){13-14} \cmidrule(r){15-15} \cmidrule(r){16-16} \cmidrule(r){17-17}
& ST & FA & VA & CU & WF & MB & SD & ST & VA & CE & FE & SN & CC & ND & VQ & QB \\
%& MST & FAS & VAW & CUB & WFS & MBR & SPD & MST & VAW & CEA & FER & SCN & COC & NYD & VQA & QBE \\
\hline
GPT-4V~\cite{achiam2023gpt}                  & \textbf{91.8} & \textbf{89.0}    & 76.9 & 71.4 & \textbf{72.1}     & \textbf{58.3} & 41.9      & \textbf{92.2} & \textbf{87.8} & 91.8   & 83.4     & \textbf{71.4} & \textbf{73.7} & 56.1  & \textbf{63.8} & \textbf{73.0}     & \textbf{74.7}                 \\ 
Gemini1.0-Pro~\cite{geminiteam2024geminifamilyhighlycapable}              & 71.9 & 76.3    & 69.3 & 59.9 & 54.9     & 53.7 & \textbf{53.0}      & 81.8 & 70.7 & 60.6   & 71.2     & 55.1 & 58.2 & 56.6  & 54.6 & 59.5     & 63.0                 \\
LLaVA-1.6~\cite{liu2024visual}                & 84.9 & 72.1    & \textbf{77.7} & \textbf{72.6} & 68.7     & 26.5 & 20.7      & 89.7 & 79.3 & \textbf{96.2}   & \textbf{83.5}     & 51.0 & 50.2 & \textbf{67.2}  & 50.1 & 64.8     & 66.0                 \\
VILA-1.5~\cite{lin2024vila}                 & 69.9 & 66.2    & 70.9 & 55.9 & 52.0     & 49.5 & 36.8      & 71.9 & 74.5 & 57.1   & 55.6     & 51.1 & 52.9 & 51.8  & 47.7 & 64.8     & 58.0                 \\
Chance level~\cite{kil2024compbench} & 50.0 & 50.0    & 50.0 & 50.0 & 50.0     & 8.6 & 9.7      & 50.0 & 50.0 & 50.0   & 50.0     & 50.0 & 50.0 & 50.0  & 33.3 & 37.4     & 43.1 \\

\hline
\end{tabular}
}
%\vspace{-12pt}
\label{COMPBENCH_tab2}
\end{table*}

\subsubsection{Hallu-PI: Evaluating Hallucination in Multi-modal Large Language Models within Perturbed Inputs}

\label{appendix_Hallu-PI}

Similarly, in the context of the hallucination problem faced by MLLMs in visual-language understanding and generation tasks~\cite{rohrbach2018object,dai2022plausible,li2023evaluating,zhang2024groundhog,zhai2023halle,liu2023mitigating,you2023ferret,zhou2023analyzing,wang2023llm}, Peng Ding et al.~\cite{ding2024hallu} pointed out that previous studies have mainly focused on evaluating hallucinations on standard, undisturbed benchmarks, neglecting the prevalent interference inputs in the real world. This is crucial for a comprehensive evaluation of hallucinations in MLLMs. They proposed the first benchmark designed to evaluate hallucinations in MLLMs under disturbed inputs, called Hallu-PI, which includes seven types of disturbed scenarios: noise, blur, weather, digits, image stitching, image cropping, and prompt misdirection.

Table~\ref{hallupi_tab1} presents the performance of MLLMs under four basic disturbance types (noise, blur, weather, and digits). The "Before/After" columns compare the performance before and after the perturbation, using the ACC+ (Accuracy+) and CHAIR (Hallucinated Object Occurrence Rate) metrics to measure the level of hallucinations in the models.

\begin{table*}[htbp]
\renewcommand\arraystretch{1.2}
  \centering
  \caption{The results under noise, blur, weather, and digital perturbations. Before/After means before/after perturbation.~\cite{ding2024hallu}}
    \begin{tabular}{l|cc|cc|cc|cc|cc}
   \hline
    \multicolumn{1}{c|}{\multirow{3}[4]{*}{Model}} & \multicolumn{2}{c|}{\multirow{2}[1]{*}{Before}} & \multicolumn{8}{c}{After} \\

     \cmidrule(r){4-11}
          & \multicolumn{2}{c|}{} & \multicolumn{2}{c|}{Noise} & \multicolumn{2}{c|}{Blur} & \multicolumn{2}{c|}{Weather } & \multicolumn{2}{c}{Digital } \\
\cmidrule{2-11}          & \multicolumn{1}{l}{ACC+} & \multicolumn{1}{l|}{CHAIR} & \multicolumn{1}{l}{ACC+} & \multicolumn{1}{l|}{CHAIR} & \multicolumn{1}{l}{ACC+} & \multicolumn{1}{l|}{CHAIR} & \multicolumn{1}{l}{ACC+} & \multicolumn{1}{l|}{CHAIR} & \multicolumn{1}{l}{ACC+} & \multicolumn{1}{l}{CHAIR} \\
    \hline
    CogVLM~\cite{wang2023cogvlm} & \textbf{49.0}  & 62.0    & \textbf{48.5}  & 68.2 & \textbf{47.4} & 68.6 & 42.8 & 67.9 & \textbf{48.4} & 69.8 \\
    Multi-GPT~\cite{achiam2023gpt} & 13.3  & \textbf{73.5}  & 9.6  & 73.6 & 12.8 & \textbf{76.1} & 11.2 & \textbf{73.4} & 9.2  & \textbf{77.8} \\
    LLaVA~\cite{liu2024visual} & 6.3   & 68.5  & 4.33  & 67.7 & 5.0     & 70.6 & 4.17  & 69.8 & 3.6  & 74.2 \\
    LLaVA1.5~\cite{liu2024visual} & 43.0    & 68.9  & 42.6 & 70.1 & 42.4 & 68.7 & 43.3 & 68.0  & 36.8 & 74.5 \\
    MiniGPT-4~\cite{zhu2023minigpt} & 16.0    & 72.4  & 15.8 & 70.2 & 15.9 & 72.1 & 14.5  & 72.6 & 13.8 & 73.9 \\
    MiniGPT4-v2~\cite{chen2023minigpt} & 28.3  & 72.1  & 26.7 & \textbf{74.7} & 28.8 & 74.0  & 28.2 & 72.8 & 27.1 & 74.9 \\
    mPLUG2~\cite{xu2023mplug} & 38.0    & 65.0    & 33.3 & 67.6 & 33.1 & 69.1 & 35.3 & 66.9 & 32.3 & 73.6 \\
    Gemini~\cite{team2023gemini} & 46.0    & 57.3  & 44.2 & 60.0   & 45.1 & 59.7   & 44.8 & 58.5 & 37.5  & 61.3 \\
    GPT-4V~\cite{achiam2023gpt} & 47.3  & 66.1  & 42.3 & 66.9 & 41.8 & 68.4 & \textbf{47.8} & 60.9 & 34.0    & 65.4 \\
    \hline
    \end{tabular}%
  \label{hallupi_tab1}%
\end{table*}%

Table~\ref{hallupi_tab2} focuses on the performance of MLLMs under three additional disturbance types in Hallu-PI: Concat, Cropping, and Prompt Mislead. The PI-Score (a comprehensive evaluation metric) is used to assess the overall performance of the models under these specific disturbance scenarios.

\begin{table}[htbp]
\renewcommand\arraystretch{1.2}
\small
  \centering
  \caption{The results under image concatenation, image cropping, and prompt misleading perturbations.~\cite{ding2024hallu}}
  \setlength{\tabcolsep}{1mm}{
    \begin{tabular}{l|cc|cc|cc}
    \hline
    \multicolumn{7}{c}{PI-Score} \\
    \hline
    \multicolumn{1}{c|}{\multirow{2}[2]{*}{MLLMs}} & \multicolumn{2}{c|}{Concat} & \multicolumn{2}{c|}{Cropping} & \multicolumn{2}{c}{Prompt Mislead} \\
     \cmidrule(r){2-7}
          & \multicolumn{1}{l}{Before} & \multicolumn{1}{l|}{After} & \multicolumn{1}{l}{Before} & \multicolumn{1}{l|}{After} & \multicolumn{1}{l}{Before} & \multicolumn{1}{l}{After} \\
    \hline
    CogVLM~\cite{wang2023cogvlm} & \textbf{45.4}  & 22.5 & 10.0    & 5.0     & 39.6  & 11.4 \\
    Multi-GPT~\cite{achiam2023gpt} & 8.3   & 15.0    & 11.7  & 0.0     & 18.9  & 7.2 \\
    LLaVA~\cite{liu2024visual} & 6.5   & 2.2   & 3.4   & 6.7   & 14.4  & 5.2 \\
    LLaVA1.5~\cite{liu2024visual} & 32.4  & 5.9   & 10.0    & 8.4   & 26.4  & 8.1 \\
    MiniGPT-4~\cite{zhu2023minigpt} & 8.9   & 5.9   & 10.0    & 8.4   & 18.5  & 7.0 \\
    MiniGPT-v2~\cite{chen2023minigpt} & 15.8  & 12.3  & 16.7  & 15.0    & 26.4  & 11.3 \\
    mPLUG2~\cite{xu2023mplug} & 25.7  & 18.9  & 10.0    & 8.3   & 29.7  & 15.7 \\
    InternLM~\cite{cai2024internlm2} & 38.3  & \textbf{37.3}  & 8.3   & 10.0    & 34.4  & 28.0 \\
    Qwen-VL~\cite{bai2023qwen} & 46.3  & 19.6  & 20.0    & 11.7  & 53.2  & 38.2 \\
    VisualGLM~\cite{du2022glm} & 6.8   & 0.6   & 34.0    & 0.0     & 21.2  & 11.3 \\
    Gemini~\cite{team2023gemini} & 44.6  & 21.4  & \textbf{45.0}    & 26.7  & 59.2  & 39.4 \\
    GPT-4V~\cite{achiam2023gpt} & 42.0    & 18.0    & 43.4  & \textbf{30.0}    & \textbf{61.4}  & \textbf{48.2} \\
    \hline
    \end{tabular}
    }
  \label{hallupi_tab2}%
\end{table}%

Table~\ref{hallupi_tab3} provides the performance details of MLLMs in generation tasks under the Concat, Cropping, and Prompt Mislead disturbances. The metrics CHAIR, Cover, Hal, and Cog are used to evaluate the models' performance in generation tasks. These metrics help us understand the models' accuracy and hallucination tendencies when generating descriptions that are consistent with the image content.

\begin{table*}[htbp]
\renewcommand\arraystretch{1.2}
\small
  \centering
  \caption{The results of generative task on image concatenation, cropping, and prompt misleading.~\cite{ding2024hallu}
 }
 \scalebox{0.95}{
    \begin{tabular}{l|cc|cc|cc|cc|cc|cc}
    \hline
    \multicolumn{1}{c|}{\multirow{3}[6]{*}{MLLMs}} & \multicolumn{8}{c|}{Image Concatenation}                      & \multicolumn{2}{c|}{Image Cropping} & \multicolumn{2}{c}{Prompt Misleading} \\
\cmidrule{2-13}          & \multicolumn{2}{c|}{ CHAIR } & \multicolumn{2}{c|}{ Cover } & \multicolumn{2}{c|}{Hal } & \multicolumn{2}{c|}{Cog} & \multicolumn{2}{c|}{Hal} & \multicolumn{2}{c}{Hal} \\
\cmidrule{2-13}          & \multicolumn{1}{l}{Before} & \multicolumn{1}{l|}{After} & \multicolumn{1}{l}{Before} & \multicolumn{1}{l|}{After} & \multicolumn{1}{l}{Before} & \multicolumn{1}{l|}{After} & \multicolumn{1}{l}{Before} & \multicolumn{1}{l|}{After} & \multicolumn{1}{l}{Before} & \multicolumn{1}{l|}{After} & \multicolumn{1}{l}{Before} & \multicolumn{1}{l}{After} \\
    \hline
    CogVLM~\cite{wang2023cogvlm} & 62.0  & 69.0  & 55.3  & 48.3  & 58.3  & 97.1  & 4.3   & 5.9   & 80.0  & 90.0  & 36.7  & \textbf{93.3}  \\
    Multi-GPT~\cite{achiam2023gpt} & 73.5  & \textbf{97.5}  & 22.5  & 2.0   & 96.7  & 86.3  & \textbf{30.8}  & \textbf{77.1}  & 76.7  & \textbf{100.0}  & 63.3  & \textbf{93.3}  \\
    LLaVA~\cite{liu2024visual} & 68.5  & 92.3  & 38.8  & 7.4   & 93.3  & 96.7  & 4.3   & 14.9  & \textbf{93.3}  & 86.7  & \textbf{66.7}  & \textbf{93.3}  \\
    LLaVA1.5~\cite{liu2024visual} & 68.9  & 76.1  & 43.8  & 25.0  & 78.3  & 96.3  & 3.4 & 5.7   & 86.7  & 90.0  & 63.3  & 90.0  \\
    MiniGPT-4~\cite{zhu2023minigpt} & 72.4  & 89.3  & 46.5  & 24.8  & 98.3  & 95.8  & 5.1   & 8.2   & 80.0  & 83.3  & 63.3  & \textbf{93.3}  \\
    MiniGPT-v2~\cite{chen2023minigpt} & 72.1  & 88.9  & 49.6  & 32.5  & \textbf{100.0}  & 96.7  & 4.0   & 7.1   & \textbf{93.3}  & 93.3  & 53.3  & \textbf{93.3}  \\
    mPLUG2~\cite{xu2023mplug} & 65.0  & 82.3  & 44.6  & 14.3  & 86.7  & 89.6  & 6.2   & 6.4   & \textbf{93.3}  & 96.7  & 46.7  & 80.0  \\
    InternLM~\cite{cai2024internlm2} & 58.4  & 79.2  & 16.3  & 9.5   & 71.7  & 62.5  & 18.8  & 16.7  & 86.7  & 86.7  & 43.3  & 63.3  \\
    Qwen-VL~\cite{bai2023qwen}  & 58.2  & 56.3  & 35.8  & 32.3  & 46.7  & 79.2  & 9.8   & 11.1  & 83.3  & 93.3  & 6.7   & 16.7 \\
    VisualGLM~\cite{du2022glm} & \textbf{76.9}  & 89.1  & 45.0  & 29.6  & \textbf{100.0}  & \textbf{99.2}  & 4.4   & 9.2   & \textbf{93.3}  & \textbf{100.0}  & 46.7  & 66.7  \\
    Gemini~\cite{team2023gemini} & 57.3  & 63.4  & 50.2  & 43.7  & 56.7  & 90.8  & 3.6   & 4.5   & 26.7  & 56.7  & 12.1  & 30.0  \\
    GPT-4V~\cite{achiam2023gpt} & 66.1  & 63.6  & \textbf{66.6}  & \textbf{53.6}  & 63.3  & 98.3  & 1.6   & 1.9   & 33.3  & 73.3  & 1.1   & 3.3  \\
    \hline
    \end{tabular}%
    }
  \label{hallupi_tab3}%
\end{table*}%

Table~\ref{hallupi_tab4} presents the performance of MLLMs in discriminative tasks under image stitching, cropping, and prompt misdirection disturbances. The metrics ACC, ACC+, and F1 are used to measure the models' accuracy in discriminative tasks. These data provide insights into the models' ability to handle disturbed inputs in discriminative tasks.

\begin{table*}[htbp]
\renewcommand\arraystretch{1.2}
\small
  \centering
  \caption{The results of discriminative task on image concatenation, cropping, and prompt misleading.~\cite{ding2024hallu} }
  \setlength{\tabcolsep}{1mm}{
    \begin{tabular}{l|ccc|ccc|ccc|ccc|ccc}   
    \hline
    \multicolumn{1}{c|}{\multirow{3}[3]{*}{MLLMs}} & \multicolumn{5}{c}{Image Concatenation} &       & \multicolumn{5}{c}{Image Cropping}    &       & \multicolumn{3}{c}{Prompt Misleading} \\
\cmidrule{2-16}          & \multicolumn{3}{c|}{Before} & \multicolumn{3}{c|}{After} & \multicolumn{3}{c|}{Before} & \multicolumn{3}{c|}{After} & \multicolumn{3}{c}{After} \\

\cmidrule{2-16}
          & \multicolumn{1}{l}{ACC} & \multicolumn{1}{l}{ACC+} & \multicolumn{1}{l|}{F1} & \multicolumn{1}{l}{ACC} & \multicolumn{1}{l}{ACC+} & \multicolumn{1}{l|}{F1} & \multicolumn{1}{l}{ACC} & \multicolumn{1}{l}{ACC+} & \multicolumn{1}{l|}{F1} & \multicolumn{1}{l}{ACC} & \multicolumn{1}{l}{ACC+} & \multicolumn{1}{l|}{F1} & \multicolumn{1}{l}{ACC} & \multicolumn{1}{l}{ACC+} & \multicolumn{1}{l}{F1} \\
    \hline
    CogVLM~\cite{wang2023cogvlm} &  69.9  & \textbf{49.0}  &  74.4  & \textbf{67.2}  & \textbf{42.0}  & \textbf{73.1}  & 50.0  & 0.0   &  66.7  & 50.0  & 0.0   & \textbf{66.7}  & 56.7  & 33.3  & 51.9  \\
    Multi-GPT~\cite{achiam2023gpt} & 46.8  & 13.3  & 52.4  & 41.8  & 16.3  & 48.9  & 48.3  & 0.0   & 65.2  & 45.0  & 0.0   & 62.1  & 28.3  & 6.7   & 41.1  \\
    LLava~\cite{liu2024visual} & 51.5  & 6.3   & 57.2  & 50.3  & 1.0   & 54.0  & 50.0  & 0.0   &  66.7  & 50.0  & 0.0   & \textbf{66.7}  & 1.7   & 0.0   & 3.2  \\
    LLava1.5~\cite{liu2024visual} & \textbf{70.5}  & {43.0}  & \textbf{76.1}  & 51.7  & 8.0   & 61.7  & 51.7  & 6.7   & 56.7  & 48.3  & 6.7   & 45.6  & 40.0  & 3.3   & 5.2  \\
    MiniGPT-4~\cite{zhu2023minigpt} & 43.0  & 16.0  & 47.6  & 30.2  & 7.7   & 25.4  & 38.3  & 0.0   & 55.4  & 30.0  & 0.0   & 46.2  & 20.0  & 0.0   & 33.4  \\
    MiniGPT-v2~\cite{chen2023minigpt} & 55.8  & 28.3  & 56.4  & 48.2  & 21.3  & 41.3  & 55.0  &  26.7  & 62.0  & 48.3  & \textbf{23.3}  & 47.5  & 88.3  & 80.0  & 88.8  \\
    mPLUG2~\cite{xu2023mplug} & 62.3  & 38.0  & 68.3  & 51.5  & 27.3  & 54.5  & 50.0  & 13.3  & 62.5  & 48.3  & 13.3  & 59.7  & 43.3  & 13.3  & 34.6  \\
    InternLM~\cite{cai2024internlm2} & 68.2  & 48.3  & 70.8  &  61.2  & 37.0  & 55.9  & 50.0  & 3.3   & 60.5  &  51.7  & 6.7   & 61.3  & 75.0  & 50.0  & 68.1  \\
    Qwen-VL~\cite{bai2023qwen}  & 62.5  & 39.3  & 62.0  & 55.7  & 18.3  & 52.4  &  58.3  & 23.3  & 65.7  & 48.3  & 16.7  & 53.7  & 93.3 & 86.7  &  92.9  \\
    VisualGLM~\cite{du2022glm} & 46.3  & 5.3   & 50.9  & 43.3  & 0.3   & 45.0  & 50.0  & 0.0   &  66.7  & 50.0  & 0.0   & \textbf{66.7}  & 30.0  & 13.3  & 36.3  \\
    Gemini~\cite{team2023gemini} & 65.7  & 46.0  & 64.1  & 60.0  & 33.7  &  63.2 & {56.7}  & 16.7  & \textbf{67.5}  & 
\textbf{53.3}  & 10.0  & \textbf{66.7}  & 53.3  & 13.3  & 33.3  \\
    GPT-4V~\cite{achiam2023gpt} & 66.7  & 47.3  & 66.1  & 59.8  & 34.3  & 55.8  & \textbf{61.7}  & \textbf{33.3}  &  66.7  & \textbf{53.3}  &  20.0  & 62.5  & \textbf{95.0}  & \textbf{90.0}  & \textbf{94.7}  \\
    \hline
    \end{tabular}}
  \label{hallupi_tab4}%
\end{table*}%

\subsubsection{ReForm-Eval: Evaluating MLLMs via Unified Re-Formulation of Task-Oriented Benchmarks}

\label{appendix_ReForm-Eval}

MLLMs have made significant progress in understanding and reasoning about visual information~\cite{achiam2023gpt,zhu2023minigpt,liu2024visual,ye2023mplug,chen2023shikra}. However, this has posed challenges for the automatic evaluation of free-form text outputs from MLLMs. To leverage annotations from existing benchmarks and reduce the manual effort required to construct new benchmarks, Zejun Li et al.~\cite{li2024reform} proposed a method for reformatting existing benchmarks into a unified format compatible with MLLMs. Through systematic data collection and reformatting, they introduced the ReForm-Eval benchmark, which is designed to comprehensively and quantitatively assess the capabilities of MLLMs. This approach overcomes the structural differences between existing task-oriented multimodal benchmarks and MLLMs.

\begin{figure}[htbp]
    \centering
    \includegraphics[width=0.9\linewidth,height=0.9\linewidth]{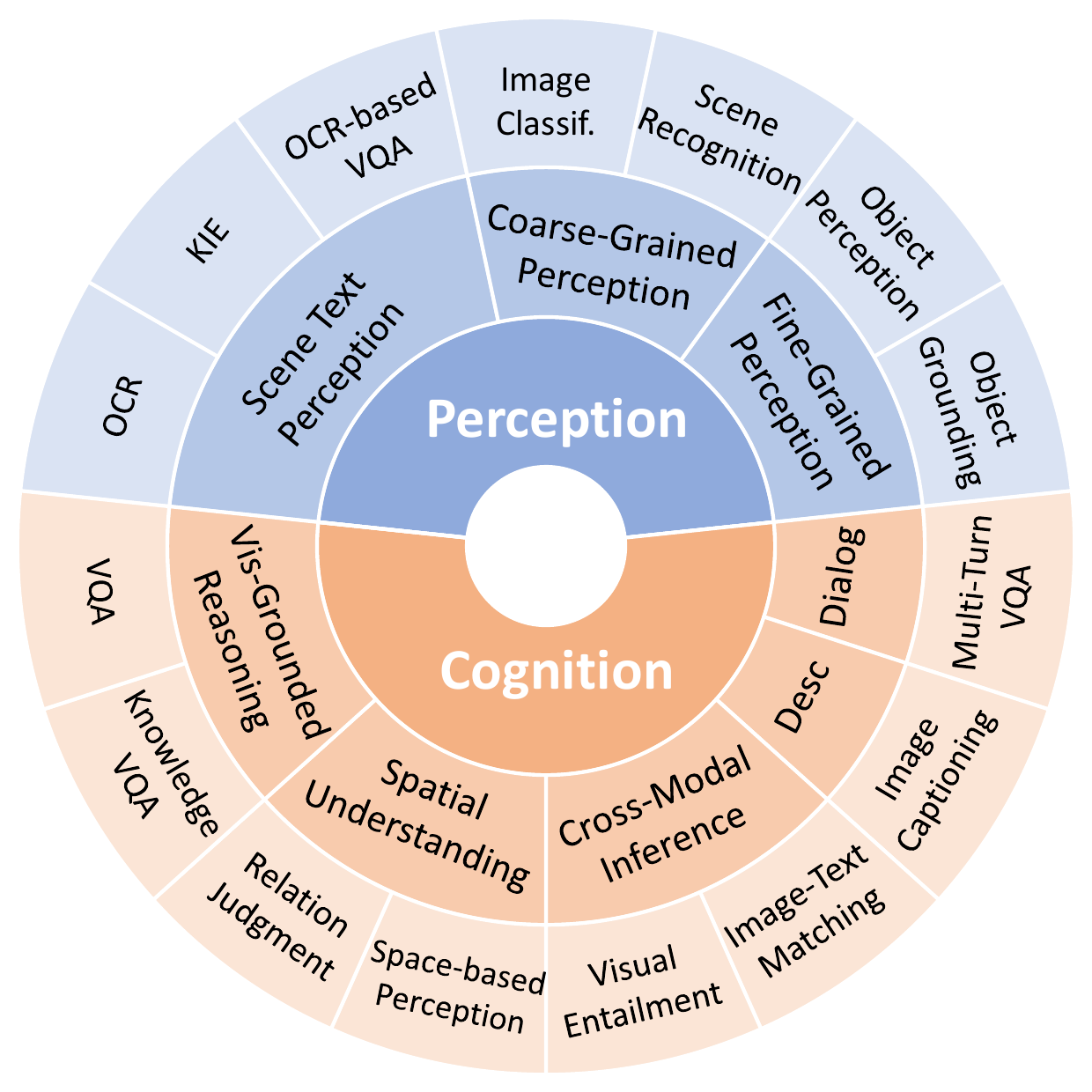}
    \caption{Assessed capability dimensions and tasks in ReForm-Eval. ''Desc'' and ''Classif'' are respectively short for description and classification.~\cite{li2024reform}}
    \label{reform_eval_fig}
\end{figure}

Figure~\ref{reform_eval_fig} illustrates the capabilities and task dimensions of the ReForm-Eval benchmark. It categorizes the evaluation dimensions into two major categories with eight subcategories: 1)Visual Perception Tasks: Coarse-Grained Perception (CG), Fine-Grained Perception (FG), Scene Text Perception (STP). 2)Visual Cognition Tasks: Visually Grounded Reasoning (VGR), Spatial Understanding (Spatial), Cross-Modal Inference (CMI), Visual Description (Desc), Multi-Turn Dialogue (Dialog).

These categories and subcategories comprehensively cover different aspects of MLLMs' visual understanding and reasoning capabilities, providing a comprehensive benchmark  for evaluating model performance.

Table~\ref{reform_eval_tab} shows a comprehensive performance evaluation of 16 open-source MLLMs across different capability dimensions, based on the ReForm-Eval benchmark.

\begin{table*}[t]
\renewcommand\arraystretch{1.2}
    \centering
        \caption{General evaluation results of MLLMs across different capability dimensions. ``CG'', ``FG'', ``CMI'', and ``Desc'' are respectively short for coarse-grained perception, fine-grained perception, cross-modal inference, and description. ``$\bar{R}$'' represents the average rank across dimensions.~\cite{li2024reform}}
    \resizebox{\textwidth}{!}{
    \begin{tabular}{c|ccc|ccccc|c|cc|cccc|c}
    \hline
    &  \multicolumn{9}{c}{\textbf{Generation Evaluation}}
    &  \multicolumn{7}{c}{\textbf{Likelihood Evaluation}} \\ \cmidrule{2-17}  
    \textbf{Model}
    & \multicolumn{3}{c|}{\textbf{Perception}}
    & \multicolumn{5}{c|}{\textbf{Cognition}} 
    & \multirow{2}{*}{$\bar{R}$}
    & \multicolumn{2}{c|}{\textbf{Perception}}
    & \multicolumn{4}{c|}{\textbf{Cognition}} 
    & \multirow{2}{*}{$\bar{R}$}  \\  \cmidrule{2-9}  \cmidrule{11-16}  
    & CG & FG & STP  & Spatial & VGR  & Dialog & CMI & Desc 
    &    & CG & FG   & Spatial & VGR  & Dialog & CMI &  \\ \midrule
    $\text{BLIP-2}_F$~\cite{li2023blip} 			& 69.4 & 76.6 & 38.1  & 43.2 & 73.3 & \textbf{61.8} & 66.9 & \textbf{74.3} & 2   & 60.7  & 74.4  & 51.1  & 69.8  & 62.6  & 58.9 & 4 \\
    $\text{InstructBLIP}_F$~\cite{dai2023instructblip}		& \textbf{71.2} & \textbf{78.1} & \textbf{41.2}  & \textbf{46.1} & \textbf{73.9} & 60.6  & \textbf{71.4} & 43.8 & \textbf{2}   & 60.4  & 75.6  & 51.2  & 71.0  & 67.2  & 55.5 & 4 \\
    $\text{InstructBLIP}_V$~\cite{dai2023instructblip} 	& 69.1 & 70.8 & 40.7  & 44.4 & 63.0 & 48.6  & 53.8 & 27.3 & 4   & 58.5  & 77.8  & 52.3  & \textbf{73.5}  & \textbf{68.7}  & 55.4 & 3 \\
    $\text{LLaVA}_V$~\cite{liu2024visual}  			& 28.7 & 34.4 & 18.4  & 28.7 & 44.0 & 35.6  & 47.3 & 36.8 & 11  & 61.0  & 70.3  & 42.4  & 58.9  & 52.3  & 48.0 & 8 \\
    $\text{LLaVA}_{L_2}$~\cite{liu2024visual}  		& 48.3 & 59.8 & 21.5  & 41.2 & 59.7 & 46.3  & 49.9 & 39.5 & 6   & 49.9  & 65.6  & 47.4  & 56.7  & 48.6  & 49.7 & 11 \\
    MiniGPT4~\cite{zhu2023minigpt}					& 46.2 & 53.2 & 33.0  & 34.6 & 45.6 & 39.5  & 45.4 & 47.5 & 7   & 54.9  & 70.6  & 49.2  & 57.3  & 54.1  & 50.9 & 8 \\
    mPLUG-Owl~\cite{ye2023mplugowl2} 					& 42.0 & 37.2 & 39.8  & 26.8 & 37.5 & 35.2  & 40.4 & 44.7 & 11  & 57.9  & 66.1  & 48.6  & 54.3  & 45.5  & 49.8 & 10 \\
    PandaGPT~\cite{su2023pandagpt}					& 28.2 & 34.6 & 4.5   & 33.3 & 41.9 & 34.1  & 36.6 & 1.6  & 14  & 42.3  & 47.4  & 39.4  & 43.3  & 41.5  & 37.0 & 16 \\
    IB-LLM~\cite{han2023imagebind}					& 29.2 & 32.7 & 8.2   & 35.6 & 36.7 & 35.3  & 36.6 & 27.6 & 13  & 49.6  & 54.4  & 46.1  & 50.3  & 39.5  & 45.6 & 15 \\
    LA-V2~\cite{gao2023llama}						& 33.2 & 30.8 & 24.2  & 23.8 & 36.3 & 35.4  & 41.1 & 36.0 & 13  & 42.7  & 61.4  & 48.6  & 54.1  & 43.4  & 49.9 & 12 \\
    mmGPT~\cite{gong2023multimodal}						& 30.4 & 30.3 & 16.7  & 26.9 & 33.0 & 31.8  & 38.2 & 27.7 & 14  & 52.6  & 62.4  & 47.2  & 56.2  & 43.1  & 44.1 & 13 \\
    Shikra~\cite{chen2023shikra}						& 47.2 & 47.5 & 8.3   & 33.3 & 41.2 & 35.2  & 44.5 & 31.8 & 11  & 60.9  & 66.8  & 45.5  & 58.5  & 59.5  & \textbf{59.3} & 7 \\
    Lynx~\cite{zeng2024matters}						& 59.5 & 62.6 & 18.6  & 40.2 & 58.4 & 47.0  & 53.0 & 60.7 & 5   & \textbf{66.1}  & 76.2  & \textbf{53.9}  & 69.9  & 60.0  & 57.4 & 3 \\
    $\text{Cheetor}_V$~\cite{li2023empowering} 			& 52.0 & 50.3 & 25.9  & 30.6 & 49.9 & 40.3  & 47.4 & 61.6 & 7   & 56.1  & 69.0  & 48.4  & 58.7  & 57.6  & 50.6 & 8 \\
    $\text{Cheetor}_{L_2}$~\cite{li2023empowering}		& 46.5 & 51.4 & 18.8  & 34.5 & 54.4 & 40.6  & 44.0 & 43.9 & 8   & 61.6  & 56.1  & 48.7  & 57.5  & 46.8  & 47.2 & 11 \\
    BLIVA~\cite{hu2024bliva}						& 41.7 & 43.4 & 40.8  & 33.3 & 42.4 & 39.8  & 45.2 & 52.5 & 8   & 64.9  & \textbf{78.2}  & 51.7  & 72.9  & 68.1  & 53.7 & \textbf{2}     
    \\ \hline
    \end{tabular}}
    \label{reform_eval_tab}
\end{table*}

\subsubsection{VisionGraph: Graph Theory Problems Benchmark in Visual Context}

\label{appendix_VisionGraph}

MLLMs have achieved significant success in visual understanding and reasoning~\cite{luo2023wizardmath,liu2024visual,chen2023shikra,achiam2023gpt}, but multimodal graph reasoning remains a challenging task~\cite{wang2024can}. It requires MLLMs to accurately understand graph structures and perform multi-step reasoning on visual graphs. To explore the ability of advanced MLLMs to address multimodal graph reasoning tasks, Yunxin Li et al.~\cite{li2024visiongraph} designed a benchmark called VisionGraph, which includes a series of graph reasoning problems aimed at testing MLLMs' understanding of graph structures and their multi-step reasoning capabilities.

Table~\ref{VisionGraph_tab1} presents the performance of different MLLMs on the VisionGraph benchmark, including evaluation metrics such as node recognition accuracy, edge recognition accuracy, and solution accuracy for specific graph theory problems. These results provide valuable insights for researchers into the models' abilities to understand and reason about graph structures.

\begin{table*}[t]
\renewcommand\arraystretch{1.20}
\caption{MLLMs results in the VisionGraph benchmark.~\cite{li2024visiongraph}}
\label{VisionGraph_tab1}
\centering
\scriptsize
\scalebox{0.95}{
\begin{tabular}{l|ccccccccc}
\hline
Model$\downarrow$ Task Types $\rightarrow$ & Connect & Cycle & Topo. Sort & Shortest Path & Max. Flow & Bipartite Graph & Hamilton Path & GNNs \\ 
\hline
 & \multicolumn{9}{c}{\textit{Node Recognition}} \\
\hline
%MiniGPT-4 $^{\clubsuit}$(Vicuna-7b, T=1.0) & 19.41 & 16.75 & 36.30 & 37.50 & 29.31 & 9.52 & 43.10 & 41.03 & \\
MiniGPT-4 (Vicuna-7b)~\cite{zhu2023minigpt} & 19.14 & 12.04 & 42.96 & 42.19 & 32.76 & 8.33 & 60.34 & 53.85 & \\
BLIP-2 (FlanT5-xxl)~\cite{li2023blip} & 37.74 & 52.88 & 47.41 & 81.25 & 67.24 & 22.62 & 62.07 & 61.54 &  \\
%mPLUG-Owl$^{\clubsuit}$ (Vicuna-7b) \\
%mPLUG-Owl$^{\clubsuit}$ (V-4-shot) \\
InstructBLIP (FlanT5-xl)~\cite{dai2023instructblip} & 36.12 & 47.64 & 46.67 & 75.00 & 56.90 & 36.90 & 53.45 & 74.36 & \\
InstructBLIP (FlanT5-xxl)~\cite{dai2023instructblip} & 35.31 & 52.88 & 61.48 & 85.94 & 77.59 & 17.86 & 65.52 & 61.54 &  \\
Sphinx~\cite{lin2023sphinx}  &	61.99	&98.95	&94.07&	100.00	&91.38	&55.95	&\textbf{100.00}&	97.44\\
Internlm~\cite{cai2024internlm2}  &	\textbf{67.92}	& \textbf{100.00}	&\textbf{97.78}	&\textbf{100.00}	&\textbf{98.25}	&\textbf{77.38}	&\textbf{100.00}	&\textbf{100.00}\\
Llava-v1.5-7b~\cite{liu2024visual}  &   64.15 & 96.86 & 92.59 & 100.00 & 93.10 & 13.10 & \textbf{100.00} & 94.87 \\
Llava-v1.5-13b~\cite{liu2024visual}  &   62.26 & 97.91 & 91.11 & 100.00 & 96.55 & 11.9 & \textbf{100.00} & 97.44 \\
Qwen-Plus (0-shot)~\cite{bai2023qwen}	& 2.96	&0.00	&0.00	&0.00	&5.17&	0.00	&0.00	&56.41\\
Qwen-max (0-shot)~\cite{bai2023qwen}	&29.11	&31.94	&30.37&	12.50	&3.45	&14.29	&29.31	&46.15\\
Gemini (0-shot)~\cite{geminiteam2024geminifamilyhighlycapable} &  40.97 & 42.93 & 47.41 & 67.19 & 72.41 & 10.71 & 65.52 & 35.90 \\
GPT-4V (0-shot)~\cite{achiam2023gpt} &   46.49 & 81.15 & 81.48 & 89.06 & 58.62 & 20.24 & 100.00 & 97.44 \\
\hline

& \multicolumn{9}{c}{\textit{Edge Recognition (Correct / Error)}} \\
\hline
MiniGPT-4 (Vicuna-7b)~\cite{zhu2023minigpt} & 11.78/31.78 & 0.68/1.59 & 12.54/58.89 & 4.78/87.20 & 0.61/61.15 & 14.45/47.53 & 28.48/34.69 & 37.48/55.05 & \\
BLIP-2 (FlanT5-xxl)~\cite{li2023blip} & 12.49/84.03 & 15.11/84.69 & 0.08/2.14 & 1.75/96.84 & 0.00/0.00 & 9.92/75.89 & 11.73/45.55 & 17.26/\textbf{88.84} & \\
Sphinx ~\cite{lin2023sphinx}	&44.76/66.69	&22.13/79.69	&37.84/73.07	&39.88/70.62	&20.68/86.57	&\textbf{83.93}/53.51	&66.26/71.15	&60.66/61.43\\
Internlm ~\cite{cai2024internlm2} &53.08/35.01&	40.78/60.05&	\textbf{55.70}/50.85	&\textbf{57.82}/45.02	&\textbf{23.45}/80.27	&71.21/42.34	&\textbf{73.98}/36.00	&\textbf{83.00}/19.69\\
InstructBLIP (FlanT5-xl)~\cite{dai2023instructblip} & 17.24/\textbf{87.62} & 26.02/\textbf{88.06} & 0.00/0.00 & 5.70/93.93 & 0.00/0.00 & 12.72/83.13 & 37.07/\textbf{82.85} & 49.18/81.28 & \\
InstructBLIP (FlanT5-xxl)~\cite{dai2023instructblip} & 16.34/81.50 & 16.04/85.54 & 0.00/0.00 & 3.58/\textbf{98.31} & 0.00/0.00 & 13.26/\textbf{76.86} & 32.05/65.84 & 37.70/67.57 & \\
Llava-v1.5-7b~\cite{liu2024visual}    & 46.81/58.13 & 23.23/77.63 & 36.56/72.97 & 38.76/66.47 & 9.80/91.56 & 63.10/54.70 & 80.14/48.06 & 69.85/32.92 \\

Llava-v1.5-13b~\cite{liu2024visual}    & 51.18/53.41 & 22.60/76.91 & 38.80/70.26 & 41.93/63.50 & 9.89/91.72 & 67.88/54.21 & 76.26/45.21 & 67.40/33.59 \\

Qwen-Plus~\cite{bai2023qwen}	&30.46/64.78	&27.42/82.37	&10.59/68.46&	6.16/81.60	&1.32/64.62	&75.93/58.65	&48.63/50.41	&33.71/60.56 \\
Qwen-max~\cite{bai2023qwen}	& 25.71/63.21&	20.92/83.50&	16.70/\textbf{76.00}	&1.63/95.70	&1.12/\textbf{96.58}	&42.59/55.55&	40.47/51.61&	35.17/55.81\\
Gemini (0-shot)~\cite{geminiteam2024geminifamilyhighlycapable}  & 23.26/52.35 & 21.65/80.09 & 19.11/66.94 & 16.18/83.09 & 4.79/94.78 & 66.01/53.90 & 39.40/37.80 & 40.83/52.60 \\
GPT-4V (0-shot)~\cite{achiam2023gpt}   & 14.10/23.09 & 17.50/72.97 & 9.64/30.58 & 23.01/66.85 & 5.31/43.62 & 24.13/32.33 & 29.22/38.03 & 46.14/42.74 \\
GPT-4V (4-shot)~\cite{achiam2023gpt}  & 20.63/34.52 & 26.25/69.95 & 13.19/51.75 & 23.40/61.90 & 6.12/84.94 & 46.33/51.69 & 58.49/49.79 & 48.06/35.01 \\
\hline

 & \multicolumn{9}{c}{\textit{Accuracy on Specific Graph Theory Problems}} \\
\hline
%MiniGPT-4 $^{\clubsuit}$(Vicuna-7b, T=1.0) & 54.45 & 51.83 & 0.00 & 0.00 & 0.00 & 1.19 & 0.00 & 0.00 \\
MiniGPT-4 (Vicuna-7b)~\cite{zhu2023minigpt} & 50.67 & 48.69 & 0.00 & 0.00 & 0.00 & \textbf{5.95} & 0.00 & 0.00 \\
BLIP-2 (FlanT5-xxl)~\cite{li2023blip} & 46.63 & \textbf{61.26} & 0.00 & 0.00 & \textbf{13.79} & 0.00 & 0.00 & 0.00 \\
%mPLUG-Owl$^{\clubsuit}$ (Vicuna-7b) \\
%mPLUG-Owl$^{\clubsuit}$ (V-4-shot) \\
InstructBLIP (FlanT5-xl)~\cite{dai2023instructblip} & 48.79 & 47.12 & 0.00 & 0.00 & 6.90 & 0.00 & 0.00 & 0.00 \\
InstructBLIP (FlanT5-xxl)~\cite{dai2023instructblip} & 48.25 & 52.88 & 0.00 & 0.00 & 12.07 & 0.00 & 0.00 & 0.00 \\
Llava-v1.5-7b~\cite{liu2024visual}     & 53.37 & 47.12 & 0.00 & 3.12 & 1.72 & 0.00 & 0.00 & 0.00\\

Llava-v1.5-13b~\cite{liu2024visual}   & 52.83 & 47.12 & 0.00 & 4.69 & 3.45 & 0.00 & 0.00 & 0.00\\

 Gemini (0-shot)~\cite{geminiteam2024geminifamilyhighlycapable}  & 55.52 & 48.69 & 0.00 & 0.00 & 3.45 & 1.72 & 0.00 & 0.00 &\\
GPT-4V (0-shot)~\cite{achiam2023gpt}  & 38.81 & 49.21 & - & 3.12 & - & - & 0.00 & -\\
GPT-4V (2-shot)~\cite{achiam2023gpt}  & 54.98 & 52.35 & - & 6.25 & - & - & 0.00 & -\\
GPT-4V (0-COT)~\cite{achiam2023gpt}  & 30.45 & 50.26 & - & \textbf{7.69} & - & - & 0.00 & -\\
GPT-4V (2-COT)~\cite{achiam2023gpt}  & 54.71 & 52.87 &  -& 6.25 & - & - & 0.00 & -\\

%\\没有GPT-4V (4-shot)
\hline
\end{tabular}
}
\end{table*}

Table~\ref{VisionGraph_tab2} shows the performance improvements of models on three representative graph theory problems (Connectivity, Cycle, and Shortest Path) after applying the Description-Program-Reasoning (DPR) method. The DPR approach enhances MLLMs' multi-step reasoning abilities by combining natural language processing and programming logic.

\begin{table*}[t]
\renewcommand\arraystretch{1.2}
\caption{Model performance on three common graph theory problems in VisionGraph.  ~\cite{li2024visiongraph}}
\label{VisionGraph_tab2}
\centering
\scriptsize
\begin{tabular}{l|cccc|cccc|ccc}
\hline
Task Types $\rightarrow$ & \multicolumn{4}{c|}{Connectivity } & \multicolumn{4}{c|}{Cycle } & \multicolumn{3}{c}{Shortest Path } \\ 
 \cmidrule(r){2-12}
Model$\downarrow$ & Easy & Medium & Hard & Avg. & Easy & Medium & Hard  & Avg. & Easy & Hard  & Avg.\\
\hline
MiniGPT-4 (Vicuna-7b)~\cite{zhu2023minigpt} & 60.71 & 53.57 & 52.94 & 54.45 & 36.00 & 51.40 & \textbf{59.32} & 51.83 & 0.00 & 0.00 & 0.00 \\
%MiniGPT-4 $^{\clubsuit}$(Vicuna-7b, T=0.2) & 51.79 & 55.61 & 42.02 & 50.67 & 36.00 & 49.53 & 52.54 & 48.69 & 0.00 & 0.00 & 0.00 \\
BLIP-2 (FlanT5-xxl)~\cite{li2023blip} &  37.50&  43.37&  \textbf{56.30}&  46.63&  \textbf{88.00}&  \textbf{63.55}&  45.76&  \textbf{61.26}&  0.00&  0.00&  0.00\\
%mPLUG-Owl$^{\clubsuit}$ (Vicuna-7b) \\
%mPLUG-Owl$^{\clubsuit}$ (V-4-shot) \\
%InstructBLIP$^{\clubsuit}$ (Vicuna-13b) &  &  &  &  &  &  &  &  & \\
InstructBLIP (FlanT5-xl)~\cite{dai2023instructblip} &  46.43&  46.43&  53.78&  48.79&  36.00&  50.47&  45.76&  47.12&  0.00&  0.00&  0.00\\
Sphinx~\cite{lin2023sphinx}	&39.29	&45.41	&52.10&	46.63&	64.00	&49.53	&54.24	&52.88	&6.90	&0.00&	3.12\\

Internlm~\cite{cai2024internlm2}	&78.57&	\textbf{66.33}	&52.10&	52.94	& 52.00	&55.14&	\textbf{59.32}	&56.02&	0.00&	0.00&	0.00\\

Llava-v1.5-7b~\cite{liu2024visual}   & 64.29 & 50.00 & 53.78 & 53.27 & 36.00 & 50.47 & 45.76 & 47.12 & 6.90 & 0.00 & 3.12 \\

Llava-v1.5-13b~\cite{liu2024visual}   & \textbf{71.43} & 49.49 & 49.58 & 52.83 & 36.00 & 50.47 & 45.76 & 47.12 & 10.34 & 0.00 & 4.69 \\

Gemini (0-shot)~\cite{geminiteam2024geminifamilyhighlycapable}  & 69.64 & 56.63 & 47.06 & 55.52 & 60.00 & 47.66 & 45.76 & 48.69 & 0.00 & 0.00 & 0.00  \\

Gemini (DPR)~\cite{geminiteam2024geminifamilyhighlycapable}  & 66.07 & 52.04 & 36.97 & 49.32 & 76.00 & 27.10 & 22.03 & 31.93 & 0.00 & 0.00 & 0.00  \\
Qwen-plus~\cite{bai2023qwen}	&62.50 &	56.63	&47.06&	54.45	&64.00	&49.53	&54.24&	52.88&	0.00&	0.00	&0.00 \\

Qwen-max~\cite{bai2023qwen}	&62.50&	56.63&	46.22&	54.18	&64.00&	49.53	&54.24&	52.88	&0.00	&0.00	&0.00\\

GPT-4V (0-shot)~\cite{achiam2023gpt}  & 69.64 & 42.86 & 17.65 & 38.81 & 60.00 & 48.60 & 45.76 & 49.21 & 6.90 & 0.00 & 3.12 \\
GPT-4V (2-shot)~\cite{achiam2023gpt}  & 67.86 & 56.12 & 47.06 & 54.98 & 64.00 & 48.60 & 54.24 & 52.35 & 13.79 & 0.00 & 6.25 \\
GPT-4V (0-COT)~\cite{achiam2023gpt}  & 64.29 & 34.69 & 7.56 & 30.45 & 64.00 & 47.66 & 49.15 & 50.26 & 17.24 & 0.00 & 7.69 \\
GPT-4V (2-COT)~\cite{achiam2023gpt}  & 67.86 & 56.63 & 45.38 & 54.71 & 64.00 & 49.53 & 54.24 & 52.87 & 13.79 & 0.00 & 6.25  \\
GPT-4V (DPR)~\cite{achiam2023gpt} & 92.86 & 58.67 & 36.97 & \textbf{56.87} & 76.00 & 48.60 & 45.76 & 51.30 & \textbf{24.14} & \textbf{2.86} & \textbf{12.50} \\

%\\没有GPT-4V (4-shot)
\hline
\end{tabular}

\end{table*}

\subsection{Applications of MLLMs}

\label{applications_MLLM}

Zebang Cheng et al.~\cite{cheng2024emotion} proposed Emotion-LLaMA, which integrates audio, visual, and text inputs through an emotion-specific encoder, and significantly improves emotion recognition and reasoning accuracy through instruction tuning. This approach enhances the model’s ability to understand and reason about emotional content across different modalities.

Xun Wu et al.~\cite{wumultimodal} created the VisionPrefer dataset, which includes fine-grained human preference annotations. They then trained the VP-Score reward model on this dataset to guide the training of image generation models, improving the alignment between images and text prompts. Finally, they fine-tuned the model using reinforcement learning to make the generated images more aligned with human aesthetics and preferences.

Zhenyu Wang et al.~\cite{wang2024genartist} proposed the GenArtist system, which enables unified image generation and editing coordinated by a multimodal large language model. The system introduces location-aware tool execution and integrates tool libraries, enhancing the model's flexibility and applicability.

Yushi Hu et al.~\cite{hu2024visual} proposed the Visual SKETCHPAD framework, enabling multimodal language models to draw sketches and perform reasoning based on visual artifacts. This significantly enhances the model's performance in mathematical and visual tasks.

Haoyu Chen et al.~\cite{chen2024restoreagent} proposed an MLLM-based intelligent image restoration system, RestoreAgent, which can automatically assess degradation, determine tasks, select models, and perform restoration.

Haodong Chen et al.~\cite{chen2024finecliper} proposed the FineCLIPER framework, which enhances facial expression recognition performance by incorporating text description augmentation, hierarchical information mining, and parameter-efficient fine-tuning to achieve multimodal feature fusion and cross-modal contrastive learning.

Shuo Ma et al.~\cite{ma2024sleepmg} proposed SleepMG, which addresses the classification and domain-discrepancy performance issues in sleep staging by quantifying modal performance differences and adaptively adjusting gradients to achieve multimodal balance. This method specifically tackles the challenges posed by the classification of multimodal physiological signals, such as EEG, EOG, EMG, and ECG.

Yifeng Xie et al.~\cite{xie2024moba} proposed the MoBA model, which employs bidirectional adapters and a mixture of experts system to achieve efficient cross-modal interaction with a low parameter count. This approach addresses the issues of large parameter sizes and low fine-tuning efficiency in multimodal sarcasm detection.

Pinxue Guo et al.~\cite{guo2024x} proposed the X-Prompt framework, which pretrains an RGB-based model and then adapts it to downstream tasks using multimodal prompts and specialized expert adapters. This approach addresses the limitations of traditional video object segmentation in complex scenarios such as extreme lighting and fast motion.

Daiqing Wu et al.~\cite{wu2024robust} proposed the DRF method, which addresses the issues of poor modality quality and missing data in sentiment analysis of image-text pairs on social media by approximating modality distributions using feature queues.

Lv Tang et al.~\cite{tang2024chain} proposed the MMCPF framework and CoVP strategy based on MLLMs, which effectively detect camouflaged objects without labeled data, addressing the issue of weak generalization in supervised learning models for zero-shot camouflaged object detection.

Deji Zhao et al.~\cite{zhao2024autograph} proposed AutoGraph, an automatic method for constructing visual context graphs. They designed a graph sampling syntax and employed a two-stage fine-tuning strategy to enhance the visual dialogue capabilities of LLMs.

Kangzheng Liu et al.~\cite{liu2024dysarl} proposed DySarl, which effectively enhances multimodal knowledge graph reasoning performance through dual-space multi-hop structural learning and interactive symmetric attention fusion.

Bowen Zhao et al.~\cite{zhao2024ct2c} proposed the CT2C-QA dataset and the AED multi-agent system. The former includes three modalities, while the latter unifies multimodal data processing through collaborative agents and introduces new evaluation metrics to enhance question-answering performance.

Linhui Xiao et al.~\cite{xiao2024hivg} proposed the HiVG framework, which includes multi-level adaptive cross-modal bridges and hierarchical low-rank adaptation. This framework enables fine-grained multimodal feature modulation, enhancing the accuracy and efficiency of visual localization.

Ruofan Wang et al.~\cite{wang2024white} proposed a multimodal attack strategy with dual optimization objectives, which jointly attacks both the text and image modalities to increase the success rate of attacking MLLMs.

Feihong Lu et al.~\cite{lu2024miko} proposed the Miko framework, which combines LLMs and MLLMs to automatically capture user intentions by analyzing text and images, and constructs an intention knowledge base to enhance intention understanding in social media.

Pinhan Fu et al.~\cite{fu2024core} proposed CoMO-NAS, which guides multi-objective search through core structure optimization to balance model complexity and performance, improving search efficiency and meeting the diverse needs of users.

Jianing Zhao et al.~\cite{zhao2024hawkeye} addressed the challenge of detecting implicit abnormal emotions in reconnaissance videos by proposing the scene-enhanced MLLM, Hawkeye, for the IasDig task. It integrates graph-structured scene modeling with a balanced heterogeneous MoE module to optimize scene information modeling and balance, effectively reducing false alarm rates and improving detection efficiency.

Xian Zhang et al.~\cite{zhang2024differential} proposed the FINER-MLLM model, which enhances image feature extraction capabilities by fine-tuning the image encoder with LoRA and applying dual feature constraints. The model also introduces a retrieval-augmented mechanism to assist in generating accurate change descriptions.

Zhanyu Wang et al.~\cite{wang2024gpt4video} proposed the GPT4Video framework, which aims to enhance the capabilities of large language models in video understanding and generation, enabling them to better handle multimodal inputs and efficiently generate video content.

Xiuliang Duan et al.~\cite{duan2024reason} proposed the Reason-and-Execute prompting method, which enhances the model's ability to solve geometric problems by combining reasoning templates and execution templates.

Xuechen Guo et al.~\cite{guo2024llava} proposed the LLaVA-Ultra model, which introduces a fine-grained visual encoder and an adaptive sampling module through architecture improvements, addressing the performance limitations of current multimodal large language models in medical visual question answering (Med-VQA).

Yi Bin et al.~\cite{bin2024gallerygpt} constructed the large-scale painting analysis dataset, PaintingForm, and proposed the GalleryGPT model. By fine-tuning for tasks focused on visual feature analysis, the model significantly improved the performance and generalization ability of art analysis.

Dan Kondratyuk et al.~\cite{kondratyuk2023videopoet} proposed VideoPoet, a zero-shot video generation model based on LLMs. It uses a decoder architecture to process multimodal inputs and enables high-quality video synthesis, demonstrating the ability to generate complex dynamic scenes.

Yongshuo Zong et al.~\cite{zong2024safety} proposed post hoc and hybrid fine-tuning strategies to effectively enhance the safety of MLLMs, addressing the issues of harmful content generation and susceptibility to attacks in MLLMs.

Yang Jin et al.~\cite{jin2024video} proposed the Video-LaVIT framework, which achieves efficient video decomposition using keyframes and motion vectors. This approach enables unified pretraining for video, image, and text, improving the safety and efficiency of MLLMs.

Long Qian et al.~\cite{qian2024momentor} proposed the Momentor model, which incorporates a time-aware module and event-based sequence modeling to achieve fine-grained temporal understanding and video segment-level reasoning.

Zhisheng Zheng et al.~\cite{zheng2024bat} designed the SPATIAL-AST encoder, which jointly performs sound event detection, spatial localization, and distance estimation. By integrating SPATIAL-AST with LLaMA-2, they constructed the BAT model, capable of answering questions about sound source relationships in 3D environments. The model utilizes a multi-stage training strategy to progressively enhance its spatial audio perception and reasoning capabilities.

Guangzhi Sun et al.~\cite{sun2024video} proposed Video-SALMONN, the first unified model to simultaneously process video, speech, and music. They designed the MRC Q-Former structure to achieve multi-resolution information extraction, enhancing the ability of AV-LLMs to integrate speech information for comprehensive video content understanding.

Ling Li et al.~\cite{ligeoreasoner} introduced the concept of "localizability" to quantify street view images and filter high-quality data. They proposed the GeoReasoner model, which combines human reasoning knowledge and employs a two-stage fine-tuning approach to achieve geographic localization and reasoning, addressing the challenges of geographic localization in street view images.

Yunheng Li et al.~\cite{li2024cascade} proposed the Cascade-CLIP framework, which aligns multi-level visual features with text embeddings in a cascading manner. By introducing independent decoders to handle features at different levels, the framework enhances the transferability to new categories. This approach addresses the issue where the pre-trained model CLIP fails to fully leverage intermediate visual feature information in zero-shot semantic segmentation tasks.

Zhijian Huang et al.~\cite{huang2025making} proposed the RDA-Driver model, which ensures the consistency between reasoning and decision-making in MLLMs through reasoning-decision alignment constraints and a redesigned Chain-of-Thought (CoT) framework. This approach enhances the interpretability and performance of autonomous driving systems.

\section{Continue Learning}
\label{Appendix_CL}
\subsection{Non-Large Language Model Unimodal Continual Learning}

\label{appendix_cl_nlu}

\subsubsection{Framework Innovation}

Xiaoxue Han et al.~\cite{han2024topology} proposed the TACO framework, which combines graph coarsening and continual learning to dynamically store information from previous tasks. They designed an efficient graph coarsening algorithm, RePro, based on node similarity, and introduced a node fidelity preservation strategy. The effectiveness of this approach in preventing the disappearance of minority classes was theoretically validated.

Ari S. Benjamin et al.~\cite{benjamin2024continual} proposed the Neural Tangent Ensemble (NTE) framework, which views a neural network as an ensemble of fixed experts. They derived its posterior update rule, which is equivalent to a specific form of stochastic gradient descent (SGD), offering a novel perspective for understanding and mitigating catastrophic forgetting.

Daehee Lee et al.~\cite{lee2024incremental} proposed the IsCiL framework, which improves sample efficiency and task adaptability by incrementally learning shared skills. They introduced prototype-based skill retrieval and adapter learning to enable effective knowledge sharing across different tasks.

Kunlun Xu et al.~\cite{xu2024mitigate} proposed the CKP framework, which purifies data through the CDP and ILR modules, and filters out erroneous knowledge using the EKF algorithm. This approach addresses the performance degradation issue caused by incorrect labels in the Lifelong Person Re-Identification task.

Lei Liu et al.~\cite{liu2024prior} proposed the PBR framework, which operates without prior knowledge. It reduces forgetting and enhances long-tail continual learning performance through an uncertainty-guided sampling strategy and two prior-free constraints.

Yusong Hu et al.~\cite{hutask} proposed the Task-Aware Orthogonal Sparse Network (OSN), which explores shared knowledge between old and new tasks through parameter sharing. They introduced sharpness-aware orthogonal projections to optimize the update of shared parameters and reduce interference with old tasks.

Daeun Lee et al.~\cite{lee2024becotta} proposed the Mixture-of-Domain Low-rank Experts (MoDE) framework, which includes domain-adaptive routing and domain-expert collaborative loss. This framework enables input-dependent online expert fusion, improving adaptation to new domains while preserving old knowledge.

Meng Ding et al.~\cite{ding2024understanding} proposed a theoretical analysis framework for linear regression applicable to different parameterization scenarios. They revealed the impact of task sequences and algorithm parameters on forgetting and experimentally validated the theoretical findings.

Soochan Lee et al.~\cite{lee2024learning} proposed the SB-MCL framework, which achieves continual learning through sequential Bayesian updates. The neural network is fixed to prevent forgetting, and the framework is domain- and model-agnostic.

Mikel et al.~\cite{malagonself} proposed CompoNet, a modular neural network with linearly growing parameters. By combining strategies, it prevents forgetting while achieving efficient knowledge transfer and scalability.

Raymond L. Wang et al.~\cite{wangrapid} proposed a Vector-HaSH-based neural model that combines hetero-associative memory and spatially invariant CNNs to enable fast learning and continual memory. They introduced the vHSN method, which utilizes attention mechanisms and grid encoding to prevent catastrophic forgetting and enhance generalization across different environments.

Jinglin Liang et al.~\cite{liang2025diffusion} proposed the DDDR framework, which utilizes diffusion models to generate historical data. By employing contrastive learning, the framework enhances the model's generalization ability on both generated and real data, addressing the issue of catastrophic forgetting in federated continual learning.

Fernando Julio Cendra et al.~\cite{cendra2025promptccd} proposed the PromptCCD framework, which uses GMM as a prompting method to address the CCD problem. They introduced the GMP module, which dynamically generates prompts to adapt to new classes, thus solving the problem of automatically discovering new classes in continuous data streams while mitigating catastrophic forgetting.

Dong Li et al.~\cite{li2024efficient} proposed the Mecoin framework, which employs Structured Memory Units (SMU) and a Memory Construction Module (MeCo) for efficient storage and updating of class prototypes. They introduced the Memory Representation Adaptation Module (MRaM) and the Graph Knowledge Interchange Module (GKIM) to reduce parameter fine-tuning, lower the forgetting rate, and enhance the model's generalization ability.

Linglan Zhao et al.~\cite{zhao2024safe} proposed the SAFE framework, which, in the first session, inherits the knowledge of the pre-trained model through knowledge transfer loss. In subsequent sessions, the framework balances model stability and adaptability by fixing slow parameters and updating fast parameters. It introduces an entropy-based aggregation strategy to dynamically fuse the advantages of two types of learners. This approach enables the efficient use of the rich knowledge from pre-trained models in continual learning while maintaining the model's adaptability and stability when facing new data.

Wenju Sun et al.~\cite{sun2024incremental} proposed the RP2F framework, which directly combines the posterior parameters of new and old tasks. They introduced a parameter robustness prior and used perturbation methods to approximate the Hessian matrix, enabling effective knowledge sharing and backward knowledge transfer.

Xiaoqian Liu et al.~\cite{liu2024hierarchical} proposed the HAMMER framework, which identifies shared knowledge and guides multilingual learning through online knowledge analysis and a hierarchical language evaluation mechanism, effectively alleviating the forgetting problem.

Hao Yu et al.~\cite{yu2024overcoming} proposed the FedCBC framework, which overcomes forgetting through category-specific binary classifiers and selective knowledge fusion.

Xiaochen Li et al.~\cite{xiaochen2024ts} proposed the TS-ILM framework, which includes a task-level temporal pattern extractor and a time-sensitive example selector. This framework effectively captures cross-task temporal patterns, selects representative frames for replay, reduces information redundancy, and enhances memory retention.

Depeng Li et al.~\cite{li2024harnessing} proposed the AutoActivator model, which dynamically adapts neural units to new tasks, enabling on-demand network expansion. This approach addresses the issue of forgetting old classes when learning new classes incrementally in class-incremental learning.

Tom Fischer et al.~\cite{fischer2024inemo} proposed iNeMo, an incremental neural grid model, which achieves efficient class-incremental learning through latent space initialization and position regularization.

\subsubsection{Method Innovation}

Huiping Zhuang et al.~\cite{zhuang2024gacl} proposed the sample-free Generalized Analytical Continual Learning (GACL) technique, which avoids catastrophic forgetting through analytical learning. It establishes the equivalence between incremental learning and joint training, effectively addressing the challenges of handling mixed data categories.

Ang Bian et al.~\cite{bian2024make} proposed the C-Flat method, which enhances continual learning (CL) performance by optimizing the flatness of the loss landscape. The method is easy to integrate and outperforms traditional approaches comprehensively.

Yan Fan et al.~\cite{fan2024dynamic} proposed the Dynamic Subgraph Distillation (DSGD) method, which uses structural and semantic information for stable knowledge distillation. This approach enhances the model's robustness to distribution shifts and adapts to different supervision settings, addressing the practical deployment challenges in continual learning that arise from relying on a large number of labeled samples.

Li Jiao et al.~\cite{jiao2024vector} proposed the VQ-Prompt method, which utilizes vector quantization to achieve end-to-end optimization of discrete prompt selection. They introduced gradient estimation, regularization terms, and representation statistics to stabilize task knowledge learning and improve continual learning performance.

Ameya Prabhu et al.~\cite{prabhurandom} proposed the RanDumb method, which uses random transformations and linear classifiers to investigate whether the representations produced by continual learning algorithms are truly effective in online continual learning.

Yue Lu et al.~\cite{lu2024visual} proposed two consistency conditions and an invariant prompt distribution constraint to reduce interference from new tasks on old tasks, overcoming catastrophic forgetting.

Botos Csaba et al.~\cite{csabalabel} proposed the IWMS method, which addresses label delay by prioritizing the memory of samples similar to new data. This approach helps mitigate the label delay issue in online continual learning.

Qiwei Li et al.~\cite{li2024progressive} proposed the Progressive Prototype Evolution (PPE) method, which learns class prototypes during the online learning phase to alleviate forgetting. They introduced prototype similarity preservation and prototype-guided gradient constraint modules, effectively combating dual forgetting.

Chengyi Yang et al.~\cite{yang2024introducing} proposed the Gradient Projection Common Null Space (GPCNS), which enhances plasticity by utilizing gradient information from old tasks. They integrated feature and gradient information through a collaborative framework, improving the performance of continual learning.

Zeyang Zhang et al.~\cite{zhangdisentangled} introduced a factor-based task-module router to optimize task routing and reduce forgetting. They designed an invariance-based architecture search mechanism to capture shared knowledge between tasks, enhancing knowledge sharing. This approach addresses the static assumptions and catastrophic forgetting issues in Graph Neural Architecture Search (GNAS) when handling continuous graph tasks.

Jeevan Thapa et al.~\cite{thapabayesian} proposed a non-parametric Bayesian method that infers network depth using a Beta process and adapts the width through a conjugate Bernoulli process. This approach enables joint inference of both network structure and weights, enhancing continual learning performance.

Nicolas Michel et al.~\cite{michel2023rethinking} proposed a new method based on momentum knowledge distillation, which dynamically updates the teacher model using exponential moving averages. This approach effectively overcomes the challenges of data stream processing and catastrophic forgetting in online continual learning.

Yichen Wen et al.~\cite{wen2024provable} proposed the CILA algorithm, which improves model performance in continual tasks through an adaptive distillation coefficient and theoretical performance guarantees.

Yichen Wu et al.~\cite{wumitigating} proposed the POCL algorithm, which models task relationships through Pareto optimization and dynamically adjusts weights to reduce forgetting.

Hongming Piao et al.~\cite{piaofederated} proposed the Powder algorithm, which enables prompt-based dual knowledge transfer. By selectively transferring knowledge based on task relevance, it reduces communication costs, addressing the challenge of cross-task and cross-client knowledge transfer in federated continual learning.

Weichen Lin et al.~\cite{lin2024effective} proposed the Dynamic Gradient Calibration (DGC) method, which effectively utilizes historical data to calibrate gradients. By combining it with existing continual learning methods, DGC helps alleviate the issue of catastrophic forgetting caused by data stream updates in continual learning.

Doyoung Kim et al.~\cite{kim2023one} proposed an adaptive prompting method, AdaPromptCL, which effectively adapts to varying degrees of semantic change through dynamic semantic grouping and prompt adjustment. This approach addresses the challenge of task-specific semantic variations in continual learning that fixed prompting strategies face.

Jason Yoo et al.~\cite{kim2023one} proposed the Layerwise Proximal Replay (LPR) method, which adjusts the optimization geometry to balance the learning of new and old data, enabling progressive changes. This approach reduces catastrophic forgetting and underfitting, improving the model's adaptability to both new and old data.

Zhen Zhu et al.~\cite{zhu2025anytime} proposed a dynamic weight prediction method and attention-weighted PCA feature compression, enabling efficient updates and storage compression in continual learning. This approach enhances model accuracy and flexibility.

Yanshuo Liang et al.~\cite{liang2024inflora} proposed the InfLoRA method, which injects parameter reparameterization into pre-trained weights, effectively fine-tuning within a subspace. The method designs subspace elimination to prevent new tasks from interfering with old tasks, addressing the issue of forgetting old tasks when adapting to new tasks in continual learning.

Chaoxi Niu et al.~\cite{niu2024replay} proposed a Laplace smoothing-based graph task analysis and prompting method, which enables accurate prediction of task IDs and learning of task-specific knowledge without the need for data replay. This approach effectively prevents forgetting and improves classification accuracy.

Huiping Zhuang et al.~\cite{zhuangf} proposed a forward online analytical learning method, F-OAL, which does not rely on backpropagation. It updates the linear classifier using recursive least squares, helping to alleviate the issue of catastrophic forgetting in online class-incremental learning.

Wuxuan Shi et al.~\cite{shiprospective} proposed Prospective Representation Learning (PRL), which aligns reserved space and latent space to adapt new class features to the reserved space. This method balances new and old classes, improving performance in non-sample class-incremental learning.

Zitong Huang et al.~\cite{huang2024class} proposed the ACIL task and CBS strategy, which implement class balancing through clustering and greedy selection, enhancing performance in incremental learning.

Xuze Hao et al.~\cite{hao2024addressing} proposed the CIL-balanced classification loss and distribution margin loss to reduce classifier bias and enhance class separability. This approach addresses the issue of catastrophic forgetting in class-incremental learning for medical image classification.

Zhiwen Yang et al.~\cite{yang2024domain} proposed the DSSP method, which leverages domain sharing and task-specific prompt learning, along with the S²-Adapter to adapt to deep space variations. This approach eliminates the need for sample replay and effectively mitigates catastrophic forgetting.

Shiye Wang et al.~\cite{wang2024importance} proposed Shared Parameter Subspace Learning, which combines momentum updates and an importance-aware mechanism, along with cross-domain contrast and orthogonality constraints, to capture cross-domain shared information and reduce forgetting.

Bowen Zheng et al.~\cite{zhengmulti} proposed the MRFA method, which optimizes the entire layer margin by enhancing the features of review samples. By increasing the margin, this approach helps reduce catastrophic forgetting.

Kishaan Jeeveswaran et al.~\cite{jeeveswaran2024gradual} proposed the DARE method, which reduces representation drift through a three-stage training process. They introduced the IRS strategy to optimize buffer sampling, thereby improving the model's performance on old tasks.

Dawei Zhou et al.~\cite{zhou2024expandable} proposed the EASE method, which constructs task-specific subspaces using lightweight adapters and synthesizes new features for old classes by leveraging semantic information. This approach effectively alleviates catastrophic forgetting.

\begin{table*}[ht]
	\caption{The statistic of collected datasets and instructions in CoIN benchmark.~\cite{chen2024coin}}
	\label{coin_tab1}
	\centering
	\resizebox{\linewidth}{!}{
		\begin{tabular}{c|c|c|c|cc}
			\hline
			\textbf{Task} &
			\textbf{Dataset} &
			\textbf{Instruction} &
			\makecell[c]{\textbf{Train} \\ \textbf{Number}}
			& \makecell[c]{\textbf{Test} \\ \textbf{Number}} \\
			\hline
			\multicolumn{1}{c|}{\makecell[c]{\textbf{Grounding}}} & \makecell[c]{RefCOCO\;\;\; \\ RefCOCO+ \\ RefCOCOg\;} & \makecell[c]{Please provide the bounding \\ box coordinate of the region \\ this sentence describes} &  55k & 31k \\ \hline
			
			\multicolumn{1}{c|}{\textbf{Classification}} & ImageNet & \makecell[c]{What is the object in the image? \\ Answer the question using a \\ single word or phrase} &  129k & 5k \\ \hline
			
			\textbf{Image Question Answering (IQA)} & VQAv2 & \makecell[c]{Answer the question using a \\ single word or phrase} &  82k & 107k \\ \hline
			
			\makecell[c]{\textbf{Knowledge Grounded IQA}} & ScienceQA & \makecell[c]{Answer with the option's letter \\ from the given choices directly} &  12k & 4k \\ \hline
			
			\makecell[c]{\textbf{Reading Comprehension IQA}} & \makecell[c]{TextVQA} & \makecell[c]{Answer the question using a \\ single word or phrase} &  34k & 5k \\ \hline
			
			\makecell[c]{\textbf{Visual Reasoning IQA}} & GQA & \makecell[c]{Answer the question using a \\ single word or phrase} &  72k & 1k \\ \hline
			
			\makecell[c]{\textbf{Blind People} IQA} & VizWiz & \makecell[c]{Answer the question using a \\ single word or phrase} &  20k & 8k \\ \hline
			
			\makecell[c]{\textbf{OCR IQA}} & OCR-VQA & \makecell[c]{Answer the question using a \\ single word or phrase} &  165k & 100k \\ 
			\bottomrule
		\end{tabular}
	}
\end{table*}

Table~\ref{coin_tab2} shows the results of Truth Alignment ability for different methods on the CoIN benchmark. These methods include multitask training, zero-shot learning, and fine-tuning. The table lists the performance of each method on individual tasks, as well as the average performance across all tasks, including metrics such as MAA, and BWT.

\begin{table*}
	\caption{The results evaluating the \textit{Truth Alignment} ability are presented below. 
		The first line of \textbf{Sequential Finetune} are the results for each task evaluated when just tuned on the corresponding task, and the second line displays the final results of each task after fine-tuning on the last task.~\cite{chen2024coin}}
	\label{coin_tab2}
	\renewcommand\arraystretch{1.3}
	\centering
	\resizebox{\linewidth}{!}{
		\begin{tabular}{c|c|cccccccc|cc}
			\hline
			\multirow{2}{*}{\textbf{MLLM}} & \multirow{2}{*}{\textbf{Method}} & \multicolumn{8}{c|}{\textbf{Accuracy on Each Task}} & \multicolumn{2}{c}{\textbf{Overall Results}} \\ \cline{3-12} 
			&  & ScienceQA & TextVQA & ImageNet & GQA & VizWiz & Grounding & VQAV2 & OCR-VQA & MAA & BWT \\ \hline
			
			\multirow{4}{*}{LLaVA~\cite{liu2024visual}} & Multi-task & 56.77 & 49.35 & 95.55 & 56.65 & 53.90 & 30.09 & 59.50 & 55.65 & \textbf{57.18} & - \\  \cmidrule(r){2-12}
			& Zero-shot & 49.91 & 2.88 & 0.33 & 2.08 & 0.90 & 0.00 & 0.68 & 0.17 & 7.12 & -  \\ \cmidrule(r){2-12}
			
			& \multirow{2}{*}{\makecell[c]{Sequential \\ Finetune}} & \textbf{82.45} & 49.99 & \textbf{96.05} & 56.40 & \textbf{55.45} & 31.27 & 62.20 & 57.08 & \multirow{2}{*}{32.97} & \multirow{2}{*}{-32.62} \\ 
			&  & 21.26 & 28.74 & 10.25 & 36.78 & 32.45 & 0.83 & 42.50 & 57.08  \\ \hline
			
			\multirow{4}{*}{Qwen-VL~\cite{bai2023qwen}} & Multi-task & 25.70 & 60.88 & 17.05 & 56.77 & 35.58 & 6.78 & 68.67 & \textbf{63.50} & 41.87 & - \\ \cmidrule(r){2-12}
			& Zero-shot & 64.56 & 48.15 & 11.82 & 44.50 & 9.57 & 0.00 & 64.10 & 27.50 & 33.78 & - \\ \cmidrule(r){2-12}
			& \multirow{2}{*}{\makecell[c]{Sequential \\ Finetune}} & 67.69 & \textbf{66.36} & 53.70 & \textbf{59.30} & 36.38 & \textbf{63.10} & \textbf{71.00} & 47.80 & \multirow{2}{*}{43.35} & \multirow{2}{*}{-16.94} \\
			& & 31.05 & 42.45 & 29.57 & 55.57 & 15.30 & 40.33 & 67.75 & 47.80 &  & \\ \hline
			
			\multirow{4}{*}{MiniGPT-v2~\cite{chen2023minigpt}} & Multi-task & 43.55 & 19.24 & 10.57 & 28.43 & 41.62 & 0.00 & 27.12 & 1.45 & 21.50 & -  \\ \cmidrule(r){2-12}
			& Zero-shot & 32.16 & 6.83 & 0.07 & 11.58 & 35.20 & 0.00 & 12.20 & 0.03 & 12.26 & - \\ \cmidrule(r){2-12}
			& \multirow{2}{*}{\makecell[c]{Sequential \\ Finetune}} & 28.81 & 10.40 & 7.25 & 31.55 & 41.35 & 0.00 & 36.10 & 6.15 & \multirow{2}{*}{25.45} & \multirow{2}{*}{6.04} \\
			& & 44.35 & 29.89 & 11.90 & 36.95 & 42.58 & 0.00 & 38.10 & 6.15 &  &  \\ \hline 
		\end{tabular}
	}
\end{table*}

Table~\ref{coin_tab3} presents the results of Reasoning Capability for different methods on the CoIN benchmark. Similar to Table~\ref{coin_tab2}, these results provide a comprehensive evaluation of the model's understanding and reasoning capabilities across different tasks.

\begin{table*}
	\caption{The evaluation results of \textit{Reasoning Capability} are presented below. ~\cite{chen2024coin}}
	\label{coin_tab3}
	\renewcommand\arraystretch{1.3}
	\renewcommand\tabcolsep{4.0pt}
	\centering
	\resizebox{\linewidth}{!}{
		\begin{tabular}{c|c|cccccccc|cc}
			\hline
			\multirow{2}{*}{\textbf{MLLM}} &
			\multirow{2}{*}{\textbf{Method}} &
			\multicolumn{8}{c|}{\textbf{Accuracy on Each Task}} &
			\multicolumn{2}{c}{\textbf{Overall Results}} \\ \cline{3-12}
			&  & ScienceQA & TextVQA & ImageNet & GQA & VizWiz & Grounding & VQAV2 & OCR-VQA & MAA & BWT \\ 
			\hline
			\multirow{4}{*}{LLaVA~\cite{liu2024visual}} & Multi-task & 80 & 75 & \textbf{97} & 72 & 42 & 86 & 73 & 79 & 75.50 & - \\ \cmidrule(r){2-12}
			& Zero-shot & 93 & \textbf{83} & 69 & 64 & 48 & 35 & 64 & 66 & 65.25& - \\ \cmidrule(r){2-12}
			& \multirow{2}{*}{\makecell[c]{Sequential \\ Finetune}} & 92 & 75 & \textbf{97} & 72 & 42 & 58 & 75 & 78 & \multirow{2}{*}{71.28}  & \multirow{2}{*}{-10.88} \\
			& & 82 & 74 & 55 & 56 & 47 & 52 & 58 & 78  \\ 
			% &\multirow{2}{*}{MoELoRA} & 86 & 76 & 98 & 72 & 59 & 59 & 70 & 74 & \multirow{2}{*}{70.11} & \multirow{2}{*}{65.38} & \multirow{2}{*}{-8.88}  \\ 
			% & & 82 & 76 & 55 & 61 & 54 & 58 & 63 & 74  \\ 
			\cline{1-12}
			
			\multirow{4}{*}{Qwen-VL~\cite{bai2023qwen}} & Multi-task & \textbf{98} & 82 & 68 & 77 & 50 & 51 & \textbf{82} & \textbf{88} & 74.50 & - \\ \cmidrule(r){2-12}
			& Zero-shot & 97 & 81 & 78 & 74 & \textbf{54} & 58 & 81 & 74 & 74.63 & - \\ \cmidrule(r){2-12}
			& \multirow{2}{*}{\makecell[c]{Sequential \\ Finetune}} & 96 & \textbf{83} & 86 & \textbf{78} & 51 & 82 & \textbf{82} & 75 & \multirow{2}{*}{\textbf{80.97}} & \multirow{2}{*}{-3.25} \\
			& &  95 & 78  & 77 & 77  & 47 & 76 & \textbf{82}  & 75  \\ 
			% &\multirow{2}{*}{MoELoRA} &   &   &   &   &   &   &  &   & \multirow{2}{*}{} & \multirow{2}{*}{} & \multirow{2}{*}{}  \\ 
			% &&   &   &   &   &   &   &   &  \\
			\cline{1-12}
			
			\multirow{4}{*}{MiniGPT-v2~\cite{chen2023minigpt}} & Multi-task & 96 & 76 & 58 & 62 & 44 & 89 & 63 & 59 & 68.38 & - \\ \cmidrule(r){2-12}
			& Zero-shot & \textbf{98} & 72 & 48 & 63 & 48 & 80 & 64 & 61 & 66.75 & - \\ \cmidrule(r){2-12}
			& \multirow{2}{*}{\makecell[c]{Sequential \\ Finetune}} & 97 & 71 & 55 & 61 & 44 & 91 & 63 &  52 & \multirow{2}{*}{75.05}  & \multirow{2}{*}{0.00} \\
			& & 89 & 73 & 59 & 60 & 44 & \textbf{94} & 63 & 52 \\ 
			% \cdashline{2-13}
			% &\multirow{2}{*}{MoELoRA} &   &   &   &   &   &   &  &   & \multirow{2}{*}{} & \multirow{2}{*}{} & \multirow{2}{*}{}  \\ 
			% &&   &   &   &   &   &   &   &  \\
			\hline
		\end{tabular}
	}
\end{table*}

Table~\ref{coin_tab4} explores the impact of different data volumes on MLLMs' instruction following ability on the CoIN benchmark. By randomly selecting varying proportions of samples from each dataset, Table~\ref{coin_tab4} illustrates how the volume of data affects the model's performance.

\begin{table*}[ht]
	\caption{The results of LLaVA about \textbf{different data volumes} are presented below. ~\cite{chen2024coin}}
	\label{coin_tab4}
	\renewcommand\arraystretch{1.3}
	\renewcommand\tabcolsep{4.0pt}
	\centering
	\resizebox{\linewidth}{!}{
		\begin{tabular}{c|cccccccc|cc}
			\hline
			\multirow{2}{*}{\textbf{Volume}} &
			\multicolumn{8}{c|}{\textbf{Accuracy on Each Task}} &
			\multicolumn{2}{c}{\textbf{Overall Results}} \\ \cline{2-11}
			& ScienceQA & TextVQA & ImageNet & GQA & VizWiz & Grounding & VQAV2 & OCR-VQA & MAA & BWT \\ 
			\hline
			\multirow{2}{*}{0.1} & 70.00 & 42.88 & 93.45 & 36.93 & 43.7 & 3.73 & 40.48 & 45.62 & \multirow{2}{*}{30.27} & \multirow{2}{*}{-16.17} \\
			& 53.71 & 32.62 & 5.38 & 33.50 & 36.98 & 2.85 & 36.77 & 45.62  \\  \cmidrule(r){1-11}
			
			\multirow{2}{*}{0.2} & 69.86 & 46.86 & 94.38 & 44.98 & 44.15 & 4.81 & 32.55 & 52.10 & \multirow{2}{*}{30.33} & \multirow{2}{*}{-19.89} \\
			& 41.12 & 33.25 & 5.53 & 33.80 & 25.85 & 1.77 & 37.10 & 45.62  \\ \cmidrule(r){1-11}
			
			\multirow{2}{*}{0.4} & 75.33 & 47.06 & 94.95 & 52.95 & 50.77 & 10.25 & 56.73 & 55.33 & \multirow{2}{*}{33.18} & \multirow{2}{*}{-24.85} \\
			& 49.96 & 23.60 & 7.22 & 36.12 & 33.05 & 0.09 & 39.20 & 55.33  \\ \cmidrule(r){1-11}
			
			\multirow{2}{*}{0.6} & 78.09 & 47.65 & 95.85 & 55.93 & 53.08 & 10.00 & 59.17 & 46.33 & \multirow{2}{*}{31.47} & \multirow{2}{*}{-32.57} \\
			& 27.42 & 19.54 & 7.03 & 33.52 & 13.15 & 0.05 & 38.48 & 46.33 \\ \cmidrule(r){1-11}
			
			\multirow{2}{*}{0.8} & 80.02 & 48.13 & 95.45 & 54.00 & 49.85 & 28.33 & 58.35 & 56.67 & \multirow{2}{*}{30.00} & \multirow{2}{*}{-33.60} \\
			& 11.74 & 16.94 & 8.85 & 32.62 & 35.50 & 0.00 & 39.67 & 56.67  \\ \cmidrule(r){1-11}
			
			\multirow{2}{*}{1.0} & \textbf{82.45} & \textbf{49.99} & \textbf{96.05} & \textbf{56.40} & \textbf{55.45} & \textbf{31.27} & \textbf{62.20} & \textbf{57.08} & \multirow{2}{*}{\textbf{32.97}} & \multirow{2}{*}{-32.62} \\
			& 21.26 & 28.74 & 10.25 & 36.78 & 32.45 & 0.83 & 42.50 & \textbf{57.08}
			\\ \hline
		\end{tabular}
	}
\end{table*}

\subsection{Non-large Language Model Multimodal Continual Learning}
\label{appendix_cl_nlm}

\subsubsection{Framework Innovation}

Bo Yuan et al.~\cite{yuan2024continual} proposed the CPP model for multi-task joint learning, which incorporates the CCE, TKD, and TPL mechanisms to achieve end-to-end multimodal general vision perception, significantly enhancing the efficiency of continual learning.

Yu Feng et al.~\cite{feng2024cp} proposed the CP-Prompt framework, which utilizes a dual-prompt strategy and parameter-efficient adjustments to achieve domain-specific knowledge extraction and inter-domain knowledge sharing, significantly reducing the forgetting rate.

Xianghu Yue et al.~\cite{yue2024mmal} proposed the MMAL framework, which includes the modality fusion module and MSKC module. It effectively integrates audio-visual information without requiring samples, reducing forgetting and enhancing incremental learning performance.

Yuchu Yu et al.~\cite{yu2025select} proposed a selective dual-teacher knowledge transfer framework, which utilizes unlabeled data to identify teacher networks, thereby ensuring knowledge retention and maintaining zero-shot capability.

Xiang Chen et al.~\cite{chen2023continual} proposed the MSPT framework, which optimizes multimodal learning through gradient modulation and attention distillation. It balances knowledge retention and new data integration, effectively mitigating catastrophic forgetting.

Jiazuo Yu et al.~\cite{yu2024boosting} proposed a dynamic expansion framework based on MoE adapters and DDAS, enabling parameter-efficient and zero-shot continual learning.

Yiwen Ye et al.~\cite{ye2024continual} proposed MedCoSS, a staged multimodal self-supervised learning framework that avoids modality conflicts. It introduces rehearsal strategies and feature distillation, effectively preventing catastrophic forgetting and enhancing knowledge retention.

\subsubsection{Method Innovation}

Jieren Deng et al.~\cite{deng2024zero} proposed the ZiRa method, which effectively alleviates the challenge of adapting visual-language object detection models to new domains while retaining zero-shot generalization capabilities in incremental learning. This is achieved through zero-interference loss and a reparameterized dual-branch structure, without increasing memory burden.

Tao Jin et al.~\cite{jin2024calibrating} proposed a historical prompt calibration strategy, which includes intra-modal correlation estimation and inter-modal consistency alignment to calibrate prompts in pre-trained models. This enhances the task and modality relationships, addressing the issues of task unfamiliarity and modality heterogeneity in multimodal continual learning.

Jaewoo Lee et al.~\cite{leestella} proposed a localized patch importance scoring method, emphasizing the semantic interweaving of audio-visual patches. The replay-guided relevance assessment reduces forgetting of previously learned knowledge.

Longrong Yang et al.~\cite{yang2024rcs} proposed the RCS-Prompt method, which reduces category space overlap and establishes clear boundaries between sessions through bidirectional prompt optimization and prompt magnitude normalization. This addresses the issue of overlap between old and new category spaces in continual learning.

Zangwei Zheng et al.~\cite{zheng2023preventing} proposed the ZSCL method, which mitigates forgetting through feature space distillation and parameter space weight integration.

Kaiyang Zhou et al.~\cite{zhou2022conditional} proposed the CoCoOp method, which generates dynamic prompts using a lightweight neural network to enhance model generalization. This addresses the issue of insufficient zero-shot generalization to unseen categories when pre-trained vision-language models adapt to new tasks.

Martin Menabue et al.~\cite{menabue2025semantic} proposed a dual-level prompt mechanism and semantic residual prompts, combined with multimodal generative replay, to enhance the stability and adaptability of models in continual learning.

Yicheng Xu et al.~\cite{xu2024advancing} proposed the RAIL method, which uses recursive ridge regression and a no-training fusion module, along with the introduction of the X-TAIL setup, aiming to address the challenge of improving cross-domain classification capabilities in vision-language models during continual learning.

Linlan Huang et al.~\cite{huang2025class} proposed an adaptive representation adjustment and parameter fusion method, which adjusts the representations of old categories affected by new categories using text features. Additionally, they employ a decomposition-based parameter fusion strategy to reduce forgetting.

Through continuously innovative frameworks and methods, multimodal continual learning in non-large models has achieved a certain level of effective integration and learning across different modalities. However, with the diversification of data types and application scenarios, non-large model multimodal continual learning will face more complex tasks and dynamic environments, necessitating more flexible and efficient solutions.

\subsection{Continual Learning in Large Language Model}

\label{appendix_cl_llm}
\subsubsection{Model Innovation}

Yeongbin Seo et al.~\cite{seo2024train} proposed the TAALM method, which uses meta-learning to dynamically predict token importance, enabling targeted knowledge updates and reducing forgetting.

Haoran Que et al.~\cite{que2024d} proposed the D-CPT Law and Cross-Domain D-CPT Law, which predict the optimal training ratio to address the issue of selecting the mixed corpus ratio during continual pre-training of large language models. These methods reduce GPU resource consumption and improve domain adaptability.

Srikanth Malla et al.~\cite{malla2024copal} proposed the COPAL algorithm, which enables continual pruning without the need for retraining, thereby avoiding model retraining. This solution addresses the high computational demands and model adaptability limitations faced by large language models when adapting to new domains.

Daniel Marczak et al.~\cite{marczak2025magmax} proposed the MagMax method, which achieves effective cross-task knowledge integration through sequential fine-tuning and maximum magnitude weight selection. This approach mitigates the problem of catastrophic forgetting of old knowledge in large pre-trained models during continual learning, enabling adaptation to the continuously evolving data stream.

Weixiang Zhao et al.~\cite{zhao2024sapt} proposed the SAPT framework, which aligns the learning and selection of PET blocks through a shared attention mechanism. They introduced the ARM module to recall old tasks using pseudo-samples, enabling effective knowledge retention and transfer.

Jianheng Huang et al.~\cite{huang2024mitigating} proposed the SSR framework, which utilizes LLM-generated synthetic instances for rehearsal. This approach effectively mitigates forgetting, improves data efficiency, and maintains the model's generalization ability.

Shihan Dou et al.~\cite{dou2024loramoe} proposed the LoRAMoE framework, which integrates LoRA and router networks, introducing local balance constraints to effectively mitigate the forgetting of world knowledge while enhancing multi-task handling capabilities.

Shiwen Ni et al.~\cite{ni2023forgetting} proposed the F-Learning paradigm, which first forgets old knowledge before learning new knowledge. Experiments show that it outperforms traditional fine-tuning, and the LoRA parameter reduction method achieves results comparable to full-parameter fine-tuning.

Junhao Zheng et al.~\cite{zheng2023learn} proposed the SEQ method, which enhances the performance of LLMs in incremental learning through simple strategies, reducing both parameters and training time.

\subsubsection{Instruction Fine-tuning}

To mitigate catastrophic forgetting, Continual-T0~\cite{scialom2022fine} uses a memory buffer for rehearsal~\cite{shin2017continual}, storing data from previous tasks and replaying them during training.

ConTinTin~\cite{yin2022contintin} proposed InstructionSpeak, which includes two strategies that fully leverage task instructions to improve both forward and backward transfer. The first strategy involves learning from negative outputs, while the second focuses on revisiting the instructions of previous tasks.

ELM~\cite{jang2023exploring} trains a small expert adapter for each task on top of the LLM. It then adopts a retrieval-based approach to select the most relevant expert LLM for each new task.

Based on the parameter-efficient tuning (PET) framework, OLoRA~\cite{wang2023orthogonal} introduces orthogonal low-rank adaptation for CIT. O-LoRA gradually learns new tasks in orthogonal subspaces while preserving the LoRA parameters learned from past tasks, thereby minimizing catastrophic forgetting.

DAPT~\cite{zhao2024dapt} introduces an innovative dual-attention framework, which coordinates the learning and selection of LoRA parameters through a dual-attention learning and selection module.

LLaMA PRO~\cite{wu2024llama} introduces an innovative block expansion technique that allows new knowledge to be injected into the LLM while efficiently retaining the initial functionality through post-training.

AdaptLLM~\cite{cheng2023adapting} adapts the LLM to different domains by enriching the original training corpus with a series of content-related reading comprehension tasks. These tasks are designed to help the model leverage domain-specific knowledge while enhancing prompt performance.

\cite{zhang2023reformulating} designed an adapt-retrieve-revise process to enable the LLM to adapt to new domains.

\cite{dong2023abilities} analyzed LLMs that continuously adapt to different domains and found that the order of training data has a significant impact on the performance of LLMs.

DynaInst~\cite{mok2023large} proposes a hybrid approach that combines dynamic instruction replay with a local minima-inducing regularizer. These two components enhance the generalization of the LLM while reducing memory and computational usage in the replay module.

\section{Continual Learning in Multimodal Large Language Model}
\label{Appendix_MLLMCL}

\subsection{Benchmark}

\label{MLLMCLBenchmark}

\subsubsection{CoIN: Continual Instruction Tuning Benchmark}

\label{appendix_CoIN}

MLLMs adapt to new tasks and users' evolving needs through instruction tuning. However, these models face challenges in adapting to the constantly changing knowledge requirements of users. To address this, Cheng Chen et al.~\cite{chen2024coin} proposed the CoIN benchmark to evaluate MLLMs' performance under the sequential instruction tuning paradigm. They also introduced the MoELoRA method to help MLLMs retain previous instruction alignment, reducing catastrophic forgetting.

CoIN consists of 10 commonly used datasets, covering 8 different task categories, ensuring diversity in both instructions and tasks. Table~\ref{coin_tab1} provides a detailed list of the datasets included in the CoIN benchmark, along with their corresponding instruction types, training sample sizes, and test sample sizes. The datasets cover a variety of task types, including Referring Expression Comprehension (REC), Classification, Image Question Answering (IQA), and Knowledge Grounded IQA, among others. Each task has two versions of instructions, Type1 and Type2, to ensure the diversity and comprehensiveness of the evaluation. 

Furthermore, CoIN evaluates MLLMs from two perspectives: 1) Truth Alignment. The ability to generate the correct result in the desired format to follow task instruc- tion is the basic requirement for instruction tuning. 2) Reasoning Capability. The performance of MLLMs depends not only on the instruction following but also on the knowledge maintained in MLLMs.
Three metrics are used to measure the performance of MLLMs: 1) Backward Transfer (BWT): Measures the catastrophic forgetting that occurs after learning all tasks. 2) Mean Average Accuracy (MAA): Assesses the model's performance throughout the entire training process.

\subsubsection{CliMB: The Continual Learning in Multimodality Benchmark}

\label{appendix_CliMB}

Existing multimodal large language models are typically fine-tuned separately for each downstream task, requiring a new model to be fine-tuned and stored for each task. In contrast, multitask learning involves training on a fixed set of tasks, but it cannot dynamically learn new tasks. To address this, Tejas Srinivasan et al.~\cite{srinivasan2022climb} proposed the CLiMB benchmark, designed to study the continual learning challenges faced by multimodal large models in multimodal tasks and to systematically evaluate how upstream continual learning can quickly generalize to new multimodal and unimodal tasks. The CLiMB benchmark includes vision-and-language input tasks, such as VQAv2, NLVR2, SNLI-VE, and VCR. Additionally, the evaluation phase of the CLiMB benchmark includes: 1) Upstream Continual Learning: The model is trained on a series of vision-language tasks, and its ability to forget old tasks and transfer knowledge to new tasks is evaluated after each task. 2) Downstream Low-Shot Transfer: After training on upstream tasks, the model's adaptability to new multimodal and unimodal tasks with limited samples is assessed.

Table~\ref{climb_tab1} presents the results of different continual learning algorithms for multimodal large models in upstream multimodal task learning. It compares the upstream knowledge transfer ($\knowledgetransfer{i}$) relative to direct fine-tuning, along with the task scores $[S_{\algorithm}^i]$.

\begin{table*}[t]
\renewcommand\arraystretch{1.2}
    \centering
    \caption{
        Upstream Knowledge Transfer $\knowledgetransfer{i}$ relative to direct fine-tuning on each task, along with task score $[S_{\algorithm}^i]$ (\%), for different CL algorithms $\algorithm$ applied to \vilt.
        No CL algorithms achieve notable positive Knowledge Transfer, while the majority in fact \emph{hurt} learning of new tasks.~\cite{srinivasan2022climb}
    }
    \begin{small}
    \begin{tabular}{l|r|r|r|r|rr}
    \hline
    \multirow{2}{*}{Alg $\algorithm$} & Params &  \multicolumn{1}{c|}{Task 1} & \multicolumn{1}{c|}{Task 2} & \multicolumn{1}{c|}{Task 3} & \multicolumn{1}{c}{Task 4} \\
    & Trained &  \multicolumn{1}{c|}{VQAv2} & \multicolumn{1}{c|}{NLVR2} & \multicolumn{1}{c|}{SNLI-VE} & \multicolumn{1}{c}{VCR} \\
    \hline
    Direct FT & 100\% &  [67.70]	& [73.07] & [76.31] & [61.31] \\
    \hline
    SeqFT~\cite{yigit2023enhancing}  & 100\% &  \transfer{0.13}{67.79}	& \transfer{-1.80}{72.66} & \transfer{-3.33}{74.89} & \transfer{-5.09}{59.47} \\
    Frozen Enc~\cite{srinivasan2022climb} & 7.88\% &  \transfer{-14.10}{58.15} & \transfer{-40.78}{63.66} & \transfer{-15.98}{69.45} & \transfer{-53.47}{41.90} \\
    Frozen B9~\cite{srinivasan2022climb} & 25.92\% &  \transfer{-0.58}{67.30} & \transfer{-0.58}{72.94} & \transfer{-3.31}{74.90} & \transfer{-15.49}{55.69} \\
    ER~\cite{chaudhry2019tiny} & 100\% &  \transfer{0.26}{67.87} & \transfer{0.56}{73.20} & \transfer{-2.89}{75.08} & \transfer{-4.45}{59.70} \\
    EWC~\cite{kirkpatrick2017overcoming} & 100\% &  \transfer{0.20}{67.84} & \transfer{-2.79}{72.39} & \transfer{-4.52}{74.38} & \transfer{-4.86}{59.55} \\
    Adapters~\cite{houlsby2019parameter} & 13.02\% &  \textbf{\transfer{0.59}{68.10}} & \textbf{\transfer{2.55}{73.66}} & \textbf{\transfer{-0.56}{76.08}} & \textbf{\transfer{-0.36}{61.18}} \\
    \hline
    \end{tabular}
    \end{small}
\label{climb_tab1}
\end{table*}

Table~\ref{climb_tab2} presents the Forgetting Transfer results for six continual learning algorithms applied to multimodal large models. It shows the performance degradation on previous tasks after training on subsequent tasks, indicating the extent of catastrophic forgetting.

\begin{table*}[ht]
\renewcommand\arraystretch{1.2}
    \centering
    \caption{Full numbers for forgetting transfer $\forgetting{j}{i}$ of previously seen tasks for each CL algorithm. We also show the  transfer score $[S_\algorithm^{j \leftarrow i}]$ when evaluated on that task after training on future task $i$. The first row contains task score $[S_\algorithm^j]$ after originally training on $j^{th}$ task.~\cite{srinivasan2022climb}
    }
\begin{tabular}{l|r|r|r}
    \multicolumn{4}{c}{CL Algorithm: Sequential Fine-tuning} \\
    \hline
    %\multicolumn{7}{c}{Comparison of Algorithms (Fixed Task Order, Encoder=ViLT)}\\
    %\toprule
    \multirow{2}{*}{\backslashbox{Checkpoint}{Evaluated on}} & \multicolumn{1}{c|}{Task 1} & \multicolumn{1}{c|}{Task 2} & \multicolumn{1}{c}{Task 3} \\
    &  \multicolumn{1}{c|}{VQAv2} & \multicolumn{1}{c|}{NLVR2} & \multicolumn{1}{c}{SNLI-VE} \\
    \hline
    After training on that task & [67.79]	& [72.66] & [74.89] \\
    \hline
    Task 2: NLVR2  &  \transfer{40.97}{40.02}	& - & - \\
    Task 3: SNLI-VE &   \transfer{39.25}{41.18} & \transfer{43.81}{62.73} & - \\
    Task 4: VCR &  \transfer{63.90}{24.47} & \transfer{93.74}{51.24} & \transfer{89.93}{37.52} \\
    \hline
    % \\
    \multicolumn{4}{c}{CL Algorithm: Frozen Encoder} \\
    \hline
    %\multicolumn{7}{c}{Comparison of Algorithms (Fixed Task Order, Encoder=ViLT)}\\
    %\toprule
    \multirow{2}{*}{\backslashbox{Checkpoint}{Evaluated on}} & \multicolumn{1}{c|}{Task 1} & \multicolumn{1}{c|}{Task 2} & \multicolumn{1}{c}{Task 3} \\
    &  \multicolumn{1}{c|}{VQAv2} & \multicolumn{1}{c|}{NLVR2} & \multicolumn{1}{c}{SNLI-VE} \\
    \hline
    After training on that task & [58.15]	& [63.66] & [69.45] \\
    \hline
    Task 2: NLVR2  &  \transfer{-0.38}{58.37}	& - & - \\
    Task 3: SNLI-VE &   \transfer{-0.38}{58.37} & \transfer{-0.31}{63.70} & - \\
    Task 4: VCR &  \transfer{-0.38}{58.37} & \transfer{-0.42}{63.72} & \transfer{0.00}{69.45} \\
    \hline
    % \\
    \multicolumn{4}{c}{CL Algorithm: Frozen Bottom-9} \\
    \hline
    %\multicolumn{7}{c}{Comparison of Algorithms (Fixed Task Order, Encoder=ViLT)}\\
    %\toprule
    \multirow{2}{*}{\backslashbox{Checkpoint}{Evaluated on}} & \multicolumn{1}{c|}{Task 1} & \multicolumn{1}{c|}{Task 2} & \multicolumn{1}{c}{Task 3} \\
    &  \multicolumn{1}{c|}{VQAv2} & \multicolumn{1}{c|}{NLVR2} & \multicolumn{1}{c}{SNLI-VE} \\
    \hline
    After training on that task & [67.30]	& [72.94] & [74.90] \\
    \hline
    Task 2: NLVR2  &  \transfer{16.97}{55.90}	& - & - \\
    Task 3: SNLI-VE &   \transfer{21.36}{52.93} & \transfer{29.32}{66.21} & - \\
    Task 4: VCR &  \transfer{71.61}{19.11} & \transfer{78.52}{54.93} & \transfer{35.01}{60.34} \\
    \hline
    % \\
    \multicolumn{4}{c}{CL Algorithm: Experience Replay} \\
    \hline
    %\multicolumn{7}{c}{Comparison of Algorithms (Fixed Task Order, Encoder=ViLT)}\\
    %\toprule
    \multirow{2}{*}{\backslashbox{Checkpoint}{Evaluated on}} & \multicolumn{1}{c|}{Task 1} & \multicolumn{1}{c|}{Task 2} & \multicolumn{1}{c}{Task 3} \\
    &  \multicolumn{1}{c|}{VQAv2} & \multicolumn{1}{c|}{NLVR2} & \multicolumn{1}{c}{SNLI-VE} \\
    \hline
    After training on that task & [67.87]	& [73.20] & [75.08] \\
    \hline
    Task 2: NLVR2  &  \transfer{12.88}{59.13}	& - & - \\
    Task 3: SNLI-VE &   \transfer{12.96}{59.07} & \transfer{17.10}{69.23} & - \\
    Task 4: VCR &  \transfer{43.62}{38.27} & \transfer{78.27}{55.04} & \transfer{33.45}{61.11} \\
    \hline
    % \\
    \multicolumn{4}{c}{CL Algorithm: Elastic Weight Consolidation} \\
    \hline
    %\multicolumn{7}{c}{Comparison of Algorithms (Fixed Task Order, Encoder=ViLT)}\\
    %\toprule
    \multirow{2}{*}{\backslashbox{Checkpoint}{Evaluated on}} & \multicolumn{1}{c|}{Task 1} & \multicolumn{1}{c|}{Task 2} & \multicolumn{1}{c}{Task 3} \\
    &  \multicolumn{1}{c|}{VQAv2} & \multicolumn{1}{c|}{NLVR2} & \multicolumn{1}{c}{SNLI-VE} \\
    \hline
    After training on that task & [67.84] &	[72.39]	& [74.38] \\
    \hline
    Task 2: NLVR2  &  \transfer{39.81}{40.83}	& - & - \\
    Task 3: SNLI-VE &   \transfer{31.52}{46.46} & \transfer{25.73}{66.66} & - \\
    Task 4: VCR &  \transfer{65.25}{23.58} & \transfer{81.03}{54.25} & \transfer{73.61}{43.34} \\
    \hline
    % \\
    \multicolumn{4}{c}{CL Algorithm: Adapters} \\
    \hline
    %\multicolumn{7}{c}{Comparison of Algorithms (Fixed Task Order, Encoder=ViLT)}\\
    %\toprule
    \multirow{2}{*}{\backslashbox{Checkpoint}{Evaluated on}} & \multicolumn{1}{c|}{Task 1} & \multicolumn{1}{c|}{Task 2} & \multicolumn{1}{c}{Task 3} \\
    &  \multicolumn{1}{c|}{VQAv2} & \multicolumn{1}{c|}{NLVR2} & \multicolumn{1}{c}{SNLI-VE} \\
    \hline
    After training on that task & [68.10]	& [73.66] & [76.08] \\
    \hline
    Task 2: NLVR2  &  \transfer{-0.01}{68.11}	& - & - \\
    Task 3: SNLI-VE &   \transfer{0.04}{68.07} & \transfer{3.51}{72.83} & - \\
    Task 4: VCR &  \transfer{0.67}{67.64} & \transfer{6.48}{72.13} & \transfer{0.89}{75.70} \\
    \hline
    \end{tabular}
\label{climb_tab2}
\end{table*}

Table~\ref{climb_tab3} illustrates the impact of different upstream task sequences on the upstream knowledge forgetting of multimodal large models.

\begin{table*}[ht]
\renewcommand\arraystretch{1.2}
    \centering
    \caption{Full forgetting results with different task orders.~\cite{srinivasan2022climb}
    }
\begin{tabular}{l|r|r|r}
    \multicolumn{4}{c}{Task Order: VQAv2 $\rightarrow$ NLVR2 $\rightarrow$ SNLI-VE $\rightarrow$ VCR} \\
    \hline
    %\multicolumn{7}{c}{Comparison of Algorithms (Fixed Task Order, Encoder=ViLT)}\\
    %\toprule
    \multirow{2}{*}{\backslashbox{Checkpoint}{Evaluated on}} & \multicolumn{1}{c|}{Task 1} & \multicolumn{1}{c|}{Task 2} & \multicolumn{1}{c}{Task 3} \\
    &  \multicolumn{1}{c|}{VQAv2} & \multicolumn{1}{c|}{NLVR2} & \multicolumn{1}{c}{SNLI-VE} \\
    \hline
    After training on that task & [67.79]	& [72.66] & [74.89] \\
   \hline
    Task 2: NLVR2  &  \transfer{40.97}{40.02}	& - & - \\
    Task 3: SNLI-VE &   \transfer{39.25}{41.18} & \transfer{43.81}{62.73} & - \\
    Task 4: VCR &  \transfer{63.90}{24.47} & \transfer{93.74}{51.24} & \transfer{89.93}{37.52} \\
    \hline
    % \\
    \multicolumn{4}{c}{Task Order: SNLI-VE $\rightarrow$ VCR $\rightarrow$ VQAv2 $\rightarrow$ NLVR2} \\
    \hline
    %\multicolumn{7}{c}{Comparison of Algorithms (Fixed Task Order, Encoder=ViLT)}\\
    %\toprule
    \multirow{2}{*}{\backslashbox{Checkpoint}{Evaluated on}} & \multicolumn{1}{c|}{Task 1} & \multicolumn{1}{c|}{Task 2} & \multicolumn{1}{c}{Task 3} \\
    &  \multicolumn{1}{c|}{SNLI-VE} & \multicolumn{1}{c|}{VCR} & \multicolumn{1}{c}{VQAv2} \\
    \hline
    After training on that task & [76.29]	& [60.75] & [63.27] \\
    \hline
    Task 2: VCR  &  \transfer{84.50}{39.99}	& - & - \\
    Task 3: VQAv2 &   \transfer{85.86}{39.40} & \transfer{91.47}{28.05} & - \\
    Task 4: NLVR2 &  \transfer{77.56}{42.97} & \transfer{86.11}{29.97} & \transfer{41.94}{36.73} \\
    \hline
    % \\
    \multicolumn{4}{c}{Task Order: NLVR2 $\rightarrow$ VQAv2 $\rightarrow$ VCR $\rightarrow$ SNLI-VE} \\
    \hline
    %\multicolumn{7}{c}{Comparison of Algorithms (Fixed Task Order, Encoder=ViLT)}\\
    %\toprule
    \multirow{2}{*}{\backslashbox{Checkpoint}{Evaluated on}} & \multicolumn{1}{c|}{Task 1} & \multicolumn{1}{c|}{Task 2} & \multicolumn{1}{c}{Task 3} \\
    &  \multicolumn{1}{c|}{NLVR2} & \multicolumn{1}{c|}{VQAv2} & \multicolumn{1}{c}{VCR} \\
    \hline
    After training on that task & [73.25]	& [66.55] & [59.10] \\
    \midrule
    Task 2: VQAv2  &  \transfer{58.06}{59.68}	& - & - \\
    Task 3: VCR &   \transfer{90.63}{52.16} & \transfer{68.69}{20.87} & - \\
    Task 4: SNLI-VE &  \transfer{91.75}{51.90} & \transfer{62.59}{24.94} & \transfer{34.04}{47.51} \\
    \hline
    \end{tabular}
    \label{subtab:askorder-forgetting}
\label{climb_tab3}
\end{table*}

\subsubsection{COAST: Continual Instruction Tuning Benchmark}

\label{appendix_COAST}

An ideal MLLM should be able to continuously adjust to new tasks in the face of task flow distributions across different domains, new capabilities, and new datasets, while minimizing forgetting of prior knowledge. However, most existing MLLMs are limited to single-task adaptation and lack performance evaluation standards for continual learning of new tasks. To comprehensively assess MLLMs' continual learning performance across different domains, capabilities, and datasets, Meng Cao et al.~\cite{cao2024continual} proposed the COAST benchmark. COAST includes three incremental learning settings: 1) Domain-incremental: Simulates scenarios where MLLMs continuously adapt to different domains.
Capability-incremental: Evaluates the ability of MLLMs to progressively acquire and integrate new capabilities. 2) Dataset-incremental: Assesses the ability of MLLMs to adapt to and generalize across varying dataset distributions. 3) By chaining and reusing existing benchmark tests, the COAST benchmark creates a streaming task distribution to evaluate the performance of MLLMs when continually learning new tasks.

Table~\ref{coast_tab1} presents the average accuracy (Avg.↑) and average forgetting rate (Fgt.↓) of different continual learning methods under the COAST-domain setting. These results reflect the performance of multimodal large models on new tasks and their ability to retain performance on previous tasks while learning new ones.

\begin{table*}[t]
\renewcommand\arraystretch{1.2}
\centering
%\small
%\setlength{\tabcolsep}{1.3mm}
%\setlength{\arraycolsep}{2pt}
%\caption{\textbf{Continual instruction tuning results (\%) on COAST-Domain.} ``Doc.", ``Med.", ``Avg." and ``Fgt." represent document, medical, average accuracy and average forgetting, respectively. ``Reh.", ``Seq." and ``Joint" denote rehearsal, sequential and joint training.}
\caption{\textbf{Evaluation results (\%) of continual instruction tuning on COAST-domain.} ``Avg." and ``Fgt." represent average accuracy and average forgetting, respectively. ``Reh.", ``Seq." and ``Joint" denote rehearsal, sequential and joint training.~\cite{cao2024continual}}
%\resizebox{0.99\linewidth}{!}{
%\begin{tabular}{lcx{28}x{28}|x{28}x{28}x{28}x{28}}
\begin{tabular}{l|ccccccc}
%\begin{tabular}{lcx{28}x{28}|cccc}
%\begin{tabular}{lccccccc}
	\hline
        \textbf{Methods}  & \textbf{Params} & \textbf{Avg.} & \textbf{Fgt.}  & \textbf{ChartQA} & \textbf{DocVQA} & \textbf{IconQA} & \textbf{MedicalQA}    \\
        %\textbf{Methods}  & \# \textbf{Param↓} & \textbf{Avg.↑} & \textbf{Fgt.↓}  & \textbf{ChartQA} & \textbf{DocVQA} & \textbf{IconQA} & \textbf{MedicalQA}    \\
        %\multirow{2}{*}{\textbf{Method}}  & \multirow{2}{*}{\#\textbf{Param}} & \multicolumn{2}{c}{\textbf{Final Results}} & \multicolumn{4}{c}{\textbf{Results of Each Task}} \\ \cline{3-8}
        %& & \textbf{Avg.↑} & \textbf{Fgt.↓} & \textbf{Chart} & \textbf{Doc.} & \textbf{Icon} & \textbf{Med.}   \\
	\hline
        %\textcolor{gray!70}{GPT-4o} & --- &  \\
        Joint~\cite{cao2024continual} & 6.76B & \textbf{42.79} & --- & \textbf{21.99} & \textbf{20.08} & \textbf{64.37} & \textbf{64.73} \\
        CODA~\cite{smith2023coda} & 0.75M & 36.06 & \textbf{2.72}  & 15.03 &  16.93 & 58.96 & 53.33  \\
        Dual~\cite{wang2022dualprompt} & 0.75M & 35.80 & 2.79  & 14.92 &  16.77 & 58.60 & 52.92  \\
        L2P~\cite{wang2022learning}  & 0.75M & 35.06 & 2.91  & 14.77 &  16.73 & 57.55 & 51.20  \\
        LWF~\cite{li2017learning}  & 6.76B & 27.06 & 15.05 &  14.07 & 13.19 & 37.93 & 43.05  \\
        EWC~\cite{kirkpatrick2017overcoming}  & 6.76B & 25.82 & 15.23 &  13.73 & 11.89 & 35.12 & 42.53  \\
        Reh.~\cite{bonicelli2022effectiveness} & 6.76B & 24.92 & 15.61 &  13.10 & 11.20 & 34.83 & 40.53  \\
        Seq.~\cite{cao2024continual} & 6.76B & 24.02 & 15.83 &  11.77 & 11.29 & 33.73 & 39.27  \\
        \hline
\end{tabular}
\label{coast_tab1}
\end{table*}

Table~\ref{coast_tab2} presents the performance of different methods on the continual instruction tuning tasks under the COAST-capability setting, focusing on the ability of MLLMs to acquire and integrate new capabilities. The table categorizes tasks into Conv. (Conversation), Desc. (Detail Description), Reason (Complex Reasoning), and Ref. (Referring qa).

\begin{table*}[t]

\renewcommand\arraystretch{1.2}
\centering
%\small
%\setlength{\tabcolsep}{1.3mm}
%\setlength{\arraycolsep}{2pt}
% \renewcommand\arraystretch{1.1}
\caption{\textbf{Evaluation results (\%) of continual instruction tuning on COAST-capability.} ``Conv.", ``Desc.", ``Reason" and ``Ref." represent conversation, detail description, complex reasoning, and referring qa, respectively. ``Reh.", ``Seq." and ``Joint" denote rehearsal, sequential, and joint training.~\cite{cao2024continual}}
%\resizebox{0.99\linewidth}{!}{
% \begin{tabular}{lcx{30}x{30}|x{30}x{30}x{30}x{30}}
\begin{tabular}{l|ccccccc}
	\hline
        %\makecell{\multirow{2}{*}{Methods}}  & \multirow{2}{*}{\#Params} & \multicolumn{6}{c|}{Accuracy (\%)}  \\
	  %& & Chart & Doc. & Icon & Med.  & Avg. & Fgt.  \\
        \textbf{Methods}  & \textbf{Params} & \textbf{Avg.} & \textbf{Fgt.}  & \textbf{Conv.} & \textbf{Desc.} & \textbf{Reason} & \textbf{Ref.}   \\
	\hline
        %\textcolor{gray!70}{GPT-4o} & --- &  \\
        Joint~\cite{cao2024continual} & 6.76B & \textbf{57.95} & --- & \textbf{62.48} & \textbf{43.45} & \textbf{74.02} & \textbf{51.84} \\
        CODA~\cite{smith2023coda}   &  0.75M & 54.21 & \textbf{4.99}  &  58.91 & 40.12 & 70.71 & 47.08  \\
        Dual~\cite{wang2022dualprompt}   &  0.75M & 53.62 & 5.01  &  58.09 & 39.85 & 70.03 & 46.52  \\
        L2P~\cite{wang2022learning}    &  0.75M & 53.31 & 5.04  &  57.90 & 39.33 & 69.70 & 46.32  \\
        LWF~\cite{li2017learning}    &  6.76B & 44.15 & 9.77  & 46.11 & 24.16 & 61.43 & 44.90 \\
        EWC~\cite{kirkpatrick2017overcoming}    &  6.76B & 43.69 & 9.72  & 46.23 & 24.20 & 60.11 & 44.20 \\
        Reh.~\cite{bonicelli2022effectiveness}   &  6.76B & 43.34 & 9.79  & 45.11 & 23.93 & 60.54 & 43.76 \\
        Seq.~\cite{cao2024continual}   &  6.76B & 41.51 & 10.56 & 44.29 & 23.25 & 58.39 & 40.13 \\
        \hline
 \end{tabular}
\label{coast_tab2}
\end{table*}

Table~\ref{coast_tab3} presents the performance of various methods on the continual instruction tuning task under the COAST-dataset setting, evaluating the ability of MLLMs to adapt to and generalize across dataset distributions. The terms "SciQA," "Text," "ImgNet," "GQA," "Viz," "REC," "VQA," and "OCR" in the table represent different visual question answering datasets.

\begin{table*}[t]
\centering
%\small
%\setlength{\tabcolsep}{1.3mm}
%\setlength{\arraycolsep}{2pt}
\renewcommand\arraystretch{1.1}
\caption{\textbf{Evaluation results (\%) of continual instruction tuning on COAST-dataset.} ``Reh.", ``Seq." and ``Joint" denote rehearsal, sequential, and joint training.~\cite{cao2024continual}}
%\resizebox{0.99\linewidth}{!}{
\scalebox{0.96}{
%\begin{tabular}{lcx{22}x{22}|x{22}x{22}x{22}x{22}x{22}x{22}x{22}x{22}}
\begin{tabular}{lcc|cccccccc}
	\toprule
        %\makecell{\multirow{2}{*}{Methods}}  & \multirow{2}{*}{\#Params} & \multicolumn{6}{c|}{Accuracy (\%)}  \\
        %\textbf{Methods}  & \#\textbf{Params}  & \textbf{Chart} & \textbf{Doc.} & \textbf{Icon} & \textbf{Med.}  & \textbf{Avg.↑} & \textbf{Fgt.↓}  \\
        \textbf{Methods} & \textbf{Avg.↑} & \textbf{Fgt.↓} & \textbf{SciQA} &  \textbf{Text} &  \textbf{ImgNet} &  \textbf{GQA} &  \textbf{Viz} & \textbf{REC} & \textbf{VQA} & \textbf{OCR} \\
	\midrule
        %\textcolor{gray!70}{GPT-4o} & --- &  \\
        Joint~\cite{cao2024continual} & 57.03 & --- & 61.74 & 52.14 & 60.93 & 65.56 & 47.46 & 21.86 & 67.54 & 79.04 \\
        CODA~\cite{smith2023coda}  &  50.27  &  9.70  & 54.80 & 44.55 & 53.64 & 58.43 & 39.07 & 14.97 & 62.63 &  74.08   \\
        Dual~\cite{wang2022dualprompt}  &  49.40  &  12.03 & 53.82 & 41.88 & 52.21 & 59.24 & 39.13 & 14.05 & 62.80 &  72.14  \\
        L2P~\cite{wang2022learning}   & 49.01   & 12.12  & 53.13 & 41.64 & 51.69 & 58.96 & 38.90 & 13.78 & 62.22 &  71.78  \\
        LWF~\cite{li2017learning}   &  26.41  & 36.94  & 52.40 & 30.02 & 23.99 & 27.30 & 14.65 & 3.43 & 35.13 &  24.32 \\
        EWC~\cite{kirkpatrick2017overcoming}   &  27.24  & 32.52  & 52.93 & 31.84 & 25.13 & 28.61 & 15.25 & 5.03 & 35.21 &  23.91 \\
        Reh.~\cite{bonicelli2022effectiveness}  &  26.49  & 33.17  & 52.02 & 31.29 & 24.44 & 28.03 & 14.80 & 4.14 & 34.14 &  23.03 \\
        Seq.~\cite{cao2024continual}  &  25.35  & 35.82  & 51.57 & 30.19 & 23.27 & 26.08 & 14.19 & 1.32 & 33.49 &  22.67 \\
        \bottomrule
\end{tabular}}
\label{coast_tab3}
\end{table*}

\subsubsection{ViLCo-Bench: Video Language Continual learning Benchmark}

\label{appendix_ViLCo-Bench}

Multimodal large models in the domain of video-language continual learning involve the continuous adaptation to information from both video and text inputs, enhancing the model's ability to handle new tasks while retaining previous knowledge. This is a relatively under-explored field, and establishing appropriate benchmarks is crucial to promoting communication and research in this area. To address this, Tianqi Tang et al.~\cite{tang2024vilco} proposed the first benchmark specifically designed for video-language continual learning in multimodal large models, called ViLCo-Bench. This benchmark aims to evaluate continual learning models across a range of video-text tasks.

ViLCo-Bench includes three unique video-language tasks: 1) Moment Queries (MQ). 2) Natural Language Queries (NLQ). 3) Visual Queries (VQ). These tasks require the model to understand video content and retrieve relevant segments of the video based on language queries.

Table~\ref{vilco_tab1} presents the results of different continual learning methods on the MQ task. The evaluation used Average Recall, including R@1 and R@5 (IoU=0.3 and IoU=0.5), to measure the model's performance at different Intersection over Union (IoU) thresholds.

\begin{table*}[!t]
\small
\centering
\caption{Results of Methods on Moment Query.~\cite{tang2024vilco}}
\scalebox{0.9}{
    \begin{tabular}{l|cc|c|ccc|ccc}
        \toprule 
        \multirow{2}{*}{\centering Method}
        & \multirow{2}{*}{\centering Num. Task}
        & \multirow{2}{*}{\centering Mem. Capacity}
        & \multirow{2}{*}{\centering BwF$\downarrow$}
        & \multicolumn{3}{c}{Avg R@1 ($\%$)$\uparrow$}
        & \multicolumn{3}{c}{Avg R@5 ($\%$)$\uparrow$} \\
        
        ~& & & & IoU=0.3   & IoU=0.5 & mean
        & IoU=0.3   & IoU=0.5 & mean  \\
        \midrule 

        Upper-Bound
        & None & None & None
        & $48.07_{\pm 0.09}$ & $38.71_{\pm 0.02}$ & 43.39
        & $67.30_{\pm 0.03}$ & $56.87_{\pm 0.005}$ & 62.09 \\
        Lower-Bound
        & None & None & None
        & $19.62_{\pm 0.25}$ & $10.87_{\pm 0.06}$ & 15.25
        & $31.61_{\pm 0.76}$ & $19.11_{\pm 0.41}$ & 25.36 \\
        \midrule
        % EWC~\cite{} 
        % & 10 & None & 20.9
        % & 11.74 & 10.15 & 10.95 
        % & 16.75 & 13.60 & 15.18 \\
        % Naive~\cite{} 
        % & 5 & None & 18.8
        % & 22.74 & 17.58 & 20.16
        % & 32.92 & 27.90 & 30.41 \\
        EWC~\cite{kirkpatrick2017overcoming}
        & 5 & None & $24.2_{\pm 0.03}$
        & $17.61_{\pm 0.57}$ & $12.51_{\pm 0.14}$ & 15.06
        & $28.13_{\pm 0.03}$ & $22.33_{\pm 0.51}$ & 25.23 \\
        MAS~\cite{aljundi2018memory}
        & 5 & None & $11.5_{\pm 0.01}$
        & $14.45_{\pm 0.01}$ & $9.88_{\pm 0.003}$ & 12.17
        & $22.50_{\pm 0.06}$ & $16.89_{\pm 0.07}$ & 19.70 \\
        iCaRL~\cite{rebuffi2017icarl}
        & 5 & 1010 & $4.6_{\pm 0.01}$
        & $32.01_{\pm 0.14}$ & $23.66_{\pm 0.30}$ & 27.84
        & $50.59_{\pm 0.12}$ & $39.68_{\pm 0.003}$ & 45.14 \\
        BiC~\cite{wu2019large}
        & 5 & 1010 & \bf $1.4_{\pm 0.001}$
        & $5.28_{\pm 0.42}$ & $3.39_{\pm 0.09}$ & 4.34
        & $6.90_{\pm 0.30}$ & $4.53_{\pm 0.003}$ & 5.72 \\
       VilCo~\cite{tang2024vilco}
        & 5 & 1010 & $2.9_{\pm 0.09}$
        & $\bf 33.58_{\pm 0.06}$ & $\bf 26.24_{\pm 0.04}$ & \bf 29.91
        & $\bf 53.75_{\pm 0.33}$ & $\bf 42.70_{\pm 0.006}$ & \bf 48.23 \\
        \bottomrule
    \end{tabular}
    }
\label{vilco_tab1}
\end{table*}

Table~\ref{vilco_tab2} presents the results of various continual learning methods on the NLQ task. The NLQ task is more complex than the MQ task, as language queries are not limited to human activities but involve open-vocabulary descriptions.

\begin{table*}[!t]
\small
\centering
\caption{Results of Methods on Natural Language query.~\cite{tang2024vilco}}
\scalebox{0.9}{
    \begin{tabular}{l|cc|c|ccc|ccc}
        \toprule 
        \multirow{2}{*}{\centering Method}
        & \multirow{2}{*}{\centering Num. Task}
        & \multirow{2}{*}{\centering Mem. Capacity}
        & \multirow{2}{*}{\centering BwF$\downarrow$}
        & \multicolumn{3}{c}{Avg R@1 ($\%$)$\uparrow$}
        & \multicolumn{3}{c}{Avg R@5 ($\%$)$\uparrow$} \\
        
        ~& & & & IoU=0.3   & IoU=0.5 & mean
        & IoU=0.3   & IoU=0.5 & mean  \\
        \midrule 

        Upper-Bound
        & None & None & None
        & 13.82 & 9.20 & 11.51
        & 33.59 & 23.18 & 28.39 \\
        \midrule
        Naive
        & 13 & None & 48.76
        & 6.05 & 3.61 & 4.83
        & 16.77 & 10.07 & 13.42 \\
        EWC~\cite{kirkpatrick2017overcoming}
        & 13 & None & 50.05
        & 6.34 & 4.05 & 5.20
        & 19.50 & 12.08 & 15.79 \\
        MAS~\cite{aljundi2018memory}
        & 13 & None & 35.92
        & 7.04 & 4.22 & 5.63
        & 21.56 & 12.63 & 17.10 \\
        % iCaRL~\cite{} 
        % & 13 & 1010 &
        % & & &
        % & & & \\
       ViLCo~\cite{tang2024vilco}\
        & 13 & 1010 & \textbf{10.60}
        & \textbf{9.49} & \textbf{6.21} & \textbf{7.85}
        & \textbf{25.52} & \textbf{16.36} & \textbf{20.94} \\
        \bottomrule
    \end{tabular}
}
\label{vilco_tab2}
\end{table*}

Table~\ref{vilco_tab3} presents the results of various continual learning methods on the VQ task. The VQ task requires the system to understand the visual content of the queried image. tAP (temporal Average Precision) is used as the performance metric, which measures the distance between predicted and true locations in continuous tasks.

\begin{table*}[!t]
\small
\centering
\caption{Results of Methods on Visual Query.~\cite{tang2024vilco}}
\scalebox{0.9}{
    \begin{tabular}{l|cc|c|cccc}
        \toprule 
        \centering Method
        & \centering Num. Task
        & \centering Mem. Capacity
        & \centering BwF$\downarrow$
        & Avg tAP$_{25}$ ($\%$)$\uparrow$
        & Avg stAP$_{25}$ ($\%$)$\uparrow$
        & Avg rec ($\%$)$\uparrow$
        & Avg Succ. ($\%$)$\uparrow$ \\
        \midrule 

        Upper-Bound
        & None & None & None
        & 31 & 22 & 47.05 & 55.89  \\
        \midrule
        % Naive
        % & 5 & None & 
        % &  &  &  & \\
        EWC~\cite{kirkpatrick2017overcoming}
        & 5 & None & 51.01
        & 11.48 & 7.81 & 16.79 & 22.05 \\
        MAS~\cite{aljundi2018memory}
        & 5 & None & 47.60
        & 12.13 & 9.16 & 17.80 & 22.51 \\
        % iCaRL~\cite{} 
        % & 13 & 1010 &
        % & & &
        % & & & \\
       ViLCo~\cite{tang2024vilco}\
        & 5 & 1010 & \textbf{23.77}
        & \textbf{17.85 }& \textbf{13.23} & \textbf{26.36} & \textbf{33.38} \\
        \bottomrule
    \end{tabular}
}
\label{vilco_tab3}
\end{table*}

\subsection{Framework Innovation}

Jiazuo Yu et al.~\cite{yu2024llms} introduced the Adapter-in-Adapter framework to enhance modality alignment and collaboration. They also proposed a flexible and scalable framework, PathWeave, which incorporates modality path switching and expansion capabilities. This allows MLLMs to continuously evolve on the modality used for X-modality reasoning, addressing the high computational burden when expanding to new modalities and reducing the dependency on large-scale joint pre-training.

Saurav Jha et al.~\cite{jha2024clap4clip} proposed the CLAP framework, which enhances the model's generalization ability and reduces forgetting through probabilistic fine-tuning. It is compatible with various prompt methods and strengthens the model's uncertainty estimation capabilities.

Longxiang Tang et al.~\cite{tang2025mind} proposed the DIKI framework, which efficiently preserves pre-trained knowledge through a residual mechanism and distribution-aware calibration. This approach addresses the problem of forgetting pre-trained knowledge in MLLMs during domain-category incremental learning, maintaining a balance between the model's adaptability to new tasks and the retention of old knowledge.

Xusheng Cao et al.~\cite{cao2024generative} proposed the GMM framework based on multimodal large models, which implements incremental learning through generated label text and feature matching. This approach reduces bias toward the current task and effectively minimizes forgetting.

Keon-Hee Park et al.~\cite{park2024pre} proposed the PriViLege framework, which effectively addresses catastrophic forgetting and overfitting in MLLMs through prompt functionality and knowledge distillation.

Fanhu Zeng et al.~\cite{zeng2024modalprompt} proposed the ModalPrompt framework, which implements continuous learning without data replay through bi-modal guided prototype prompts and knowledge transfer. This approach addresses the issue of forgetting old tasks when large multimodal models sequentially learn new tasks.

Emanuele Frascaroli et al.~\cite{frascaroli2024clip} proposed the CGIL framework, which combines prompt learning and latent generative replay. It uses VAEs to learn class-conditioned distributions and generate synthetic samples, effectively addressing the issue of catastrophic forgetting in multimodal large models during continual learning.

Yukun Li et al.~\cite{li2024coleclip} proposed the CoLeCLIP framework, which enhances the performance of multimodal large models in open-domain continual learning through joint learning of task prompts and cross-domain vocabularies. It achieves cross-domain vocabulary learning, maintaining a unified semantic space for multimodal large models, and reduces interference between tasks. The framework introduces task prompt learning, addressing domain differences and category associations, thereby improving the model's adaptability and discriminative ability for new tasks.

Biqing Qi et al.~\cite{qi2024interactive} proposed the ICL framework, which combines Vision Transformers (ViT) and MLLMs. By enabling interaction between a fast intuition model and a slow deep thinking model, the framework enhances the efficiency of continual learning in multimodal large language models.

Yuexiang Zhai et al.~\cite{zhai2023investigating} proposed the EMT framework to evaluate catastrophic forgetting in MLLMs. They found that moderate fine-tuning can improve continual learning performance, but excessive fine-tuning leads to a decline in performance and the emergence of hallucinations. This offers a new perspective for improving fine-tuning strategies in MLLMs.

Xiong Wang et al.~\cite{wang2024freeze} proposed the Freeze-Omni model, which implements a three-stage training strategy to enable speech input-output capabilities without unfreezing the LLM parameters. This approach addresses the issue of catastrophic forgetting when integrating the speech modality into multimodal LLMs, preserving the LLM's intelligence level and enabling low-latency speech-to-speech conversations.

Adyasha Maharana et al.~\cite{maharana2024adapt} proposed the Adapt-$infty$ framework, which optimizes model learning efficiency and reduces computational burden through dynamic data selection and a clustering-based permanent pruning strategy. This approach effectively mitigates catastrophic forgetting in multimodal large models.

Gen Luo et al.~\cite{luo2024mono} proposed Mono-InternVL, which integrates visual experts using a mixture-of-experts structure without altering the pre-trained language model. By introducing endogenous visual pretraining, it enables progressive learning of visual knowledge from noise to high-quality data through incremental learning, effectively preventing forgetting. This approach addresses the performance degradation and catastrophic forgetting issues that arise when expanding the visual and language capabilities of multimodal large language models.

Shanshan Zhong et al.~\cite{zhong2024moextend} proposed the MoExtend framework, which expands modality capabilities without adjusting the pre-trained model by integrating new experts. They designed a three-stage training process, including alignment, extension, and fine-tuning, to enable rapid modality adaptation. Additionally, they introduced an image localization score as a new scoring function to optimize multimodal sample selection. This approach addresses the issues of catastrophic forgetting and high training costs that arise when large language models are extended to multimodal tasks, particularly in the visual-language understanding domain.

Artemis Panagopoulou et al.~\cite{panagopoulou2023x} addressed the challenges faced by multimodal large language models in continual learning, particularly in self-supervised pretraining environments. They focused on how to effectively integrate and reason across knowledge from different modalities to overcome the performance limitations of traditional methods when handling multimodal data. They proposed the HiDe-Prompt framework, which is an scalable solution designed to align multiple modalities (such as images, 3D, audio, and video) with frozen large language models and enable cross-modal reasoning without joint optimization.

\subsection{Method Innovation}

Minh Le et al.~\cite{le2024mixture} revealed the connection between self-attention and mixture-of-experts, proposing the Non-linear Residual Gate (NoRGa) to enhance the continual learning performance of multimodal large language models.

Zangwei Zheng et al.~\cite{gaostabilizing} proposed the ZAF method, which preserves knowledge through zero-shot stability regularization. They introduced the EMA-based parameter-efficient EMA-LoRA architecture, achieving the decoupling of learning and forgetting.

Huancheng Chen et al.~\cite{chen2024dual} proposed DualLoRA, which utilizes orthogonal and residual low-rank adapters along with a dynamic memory mechanism to balance model stability and plasticity, thereby improving the efficiency and effectiveness of continual learning in multimodal large language models.

Weicai Yan et al.~\cite{yan2024low} proposed the Low-Rank Prompt Interaction (LPI) method, which enhances inter-modal and inter-task interactions through low-rank decomposition and contrastive learning. They introduced task semantic distance to guide prompt learning, addressing the insufficient interaction between modalities and tasks in continual learning of multimodal large language models (MLLMs), thereby reducing catastrophic forgetting.

Didi Zhu et al.~\cite{zhu2024model} proposed the Model Tailor method, which alleviates catastrophic forgetting during fine-tuning by retaining most of the pre-trained parameters and only replacing a small number of fine-tuned parameters. This approach helps to mitigate the forgetting problem while improving performance on new tasks.

Tianxiang Hao et al.~\cite{chen2024quantized} proposed a quantized prompt technique, which uses quantization errors as a form of regularization. They designed an efficient quantization-aware training algorithm that enhances the model's generalization ability while reducing its size. This approach addresses the issues of overfitting and catastrophic forgetting in MLLMs during downstream tasks, as well as the high storage and inference costs associated with large models.

Noranart Vesdapunt et al.~\cite{vesdapunt2025hvclip} proposed HVCLIP, which transforms CLIP features into a high-dimensional vector space. Through strategies such as forgetting reduction, discrepancy reduction, and feature enhancement, HVCLIP addresses the catastrophic forgetting issue encountered during fine-tuning of MLLM pre-trained models like CLIP in unsupervised domain adaptation. This approach helps mitigate the loss of pre-trained knowledge, enhancing the model's ability to retain critical information while adapting to new tasks or domains.

Meng Cao et al.~\cite{cao2024continual} proposed a parameter-efficient tuning method that does not require rehearsal. This approach constructs intrinsic and contextual incremental embeddings to encode task-specific features and inter-task dependencies. By doing so, the model can continuously adapt to new tasks while retaining prior knowledge. This significantly alleviates the catastrophic forgetting problem in MLLMs, enhancing their ability to preserve knowledge from previous tasks while accommodating new ones.

Shikhar Srivastava et al.~\cite{srivastava2024improving} proposed and evaluated five MLLM continual learning methods aimed at mitigating linguistic forgetting. Their findings revealed that the best-performing method significantly enhanced both language and vision task performance while maintaining multimodal accuracy. 

Jingyang Qiao et al.~\cite{qiao2024llaca} proposed the LLaCA method, which dynamically adjusts the EMA weights to reduce forgetting and introduces an approximation mechanism to lower computational costs, thereby addressing the issue of catastrophic forgetting in MLLMs when learning new tasks.

Clea Rebillard et al.~\cite{rebillard2024continually} proposed the Continual Visual Mapping (CVM) method, which reduces forgetting and improves generalization by mapping the representations of small visual models to the knowledge space of a fixed large language model.

Marco Mistretta et al.~\cite{mistretta2024re} proposed the RE-tune method, which freezes the backbone of the model and trains adapters, using text prompts to guide training. This approach enables privacy-preserving, computationally efficient, and anti-forgetting incremental learning. It optimizes pre-trained multimodal biomedical models for incremental learning scenarios in chest X-ray multi-label classification, addressing challenges related to computational resources, data privacy, and catastrophic forgetting.

Yuliang Cai et al.~\cite{cai2024clumo} proposed the CluMo method, which employs a two-stage training and modality fusion prompt strategy to combine visual and textual modalities, thereby enhancing the performance of multimodal large models in continual learning and improving their ability to retain old knowledge.

Yiduo Guo et al.~\cite{guo2024efficient} proposed three strategies to overcome the stability gap, including multi-round pretraining on small-scale high-quality datasets, selecting high-quality sub-corpora for pretraining, and employing a data-mixing strategy using data similar to pretraining data. These strategies effectively enhanced the performance and adaptability of multimodal large language models in new domains.

Jinghan He et al.~\cite{he2023continual} proposed a task similarity-guided regularization and model expansion method, which effectively enhances the continual learning capability of multimodal large models.

Junhao Zheng et al.~\cite{zheng2024beyond} proposed the Fwd-Prompt method, which utilizes gradient projection techniques and a multimodal prompt pool to achieve anti-forgetting and positive transfer, without requiring old samples and with minimal parameter updates. This approach improves the performance of multimodal large models in multimodal continual learning tasks.

Yuliang Cai et al.~\cite{cai2024dynamic} proposed dynamic model expansion and task attention layers to adapt to different tasks, while employing knowledge distillation and experience replay to mitigate catastrophic forgetting in multimodal large models.

\cite{d2023multimodal} proposed an incremental learning strategy for multimodal large language models, the CPE-CLIP method. By using learnable prompts and regularization strategies, it achieves parameter-efficient transfer learning for multimodal large language models, reducing the parameter size and training costs, while enhancing the performance of few-shot class incremental learning in multimodal large models.

Zilun Zhang et al.~\cite{zhang2024preserving} proposed the model-agnostic self-uncompression method, TG, which decompresses knowledge into the training corpus to reduce forgetting. They also designed the TG-SFT strategy for supervised fine-tuning of MLLMs, addressing the common issue of catastrophic forgetting encountered during post-training or supervised fine-tuning (SFT) on domain-specific data for multimodal large models.

Ke Wang et al.~\cite{wang2024lines} proposed the LiNeS technique, which performs parameter updates with layer-specific depth differentiation, preserving the generalization ability of pretraining while improving fine-tuning task performance. This approach addresses the issue of forgetting prior knowledge during the fine-tuning of multimodal pre-trained models.

Brian Lester et al.~\cite{lester2021power} proposed an end-to-end learning soft prompt method, which adapts to new tasks by adjusting input prompts rather than the entire model parameters. This approach enhances the performance and domain adaptability of multimodal large language models in continual learning.

Runqi Wang et al.~\cite{wang2023attriclip} proposed an non-incremental learning method based on CLIP, called AttriCLIP. This method adapts to new tasks using an attribute lexicon and textual prompts, without the need for additional memory data, thereby enhancing the generalization and continual learning capabilities of multimodal large models in multimodal tasks.

Shipeng Yan et al.~\cite{yan2022generative} introduced pseudo-text replay and multimodal knowledge distillation to enhance negative sample diversity, align predictions between old and new models, and improve the performance of multimodal large models in multimodal continual learning tasks.

Andrea Cossu et al.~\cite{cossu2024continual} explored how multimodal large language models can reduce catastrophic forgetting in continual learning environments through continuous pretraining, while maintaining adaptability to new knowledge. They demonstrated the advantages of self-supervised pretraining in preserving old knowledge and proposed effective pretraining strategies.

James Seale Smith et al.~\cite{smith2023continual} proposed the C-LoRA method, which effectively mitigates catastrophic forgetting by performing continual adaptive low-rank adjustments in the cross-attention layers of multimodal large models. This approach adapts to new concepts through a self-regulating mechanism while preserving knowledge of old concepts.

Tao He et al.~\cite{he2024towards} introduced a lifelong scene graph generation task and a knowledge-aware contextual prompt learning strategy, enabling the model to effectively retain old knowledge in incremental learning. This approach addresses the issue of updating and forgetting old and new knowledge in multimodal large models during scene graph generation tasks.

\end{document}